%% file: main.tex
\documentclass[10pt,twocolumn,letterpaper]{article}

\usepackage[pagenumbers]{cvpr} 

\input{preamble}

\definecolor{cvprblue}{rgb}{0.21,0.49,0.74}
\usepackage[pagebackref,breaklinks,colorlinks,allcolors=cvprblue]{hyperref}

\def\confName{CVPR}
\def\confYear{2027}

\title{InterRole: Late-Binding Interaction Evidence for Participant-Grounded Two-Person Motion Understanding}

\author{
Lin Wu$^{1,2}$\qquad Bowen Dang$^{3}$\qquad Jinxian Zhou$^{2}$\\
Xiangyu Huang$^{2}$\qquad Hengjin Zhu$^{2}$\qquad Jianglin Lan$^{1}$\\[0.5em]
$^{1}$University of Glasgow\qquad $^{2}$Carnegie Mellon University\\
$^{3}$University of Sheffield
}

\begin{document}
\maketitle
\input{sec/0_abstract}    
\input{sec/1_intro}
\input{sec/2_related_work}
\input{sec/3_method}
\input{sec/4_experiment}
\input{sec/5_conclusion}

{
    \small
    \bibliographystyle{ieeenat_fullname}
    \bibliography{main}
}

\input{sec/X_suppl}

\end{document}

%% file: preamble.tex
\usepackage{amsmath,amssymb,amsthm}
\usepackage{bm}
\usepackage{mathtools}

\usepackage{graphicx}
\graphicspath{{figures/}}
\usepackage{booktabs}
\usepackage{multirow}
\usepackage{array}
\usepackage{colortbl}
\usepackage{xcolor}
\usepackage{pifont}
\usepackage{pgfplots}
\pgfplotsset{compat=1.18}
\usetikzlibrary{arrows.meta,positioning,fit,calc,shapes.geometric,backgrounds}   

\definecolor{pone}{RGB}{42,120,214}     
\definecolor{ptwo}{RGB}{235,104,52}     
\definecolor{prel}{RGB}{27,150,105}     
\definecolor{pgray}{RGB}{110,110,105}
\tikzset{
  tok/.style={draw=white, line width=0.3pt, minimum width=0.17cm,
              minimum height=0.17cm, inner sep=0pt},
  blk/.style={draw=black!45, rounded corners=2pt, fill=white,
              align=center, font=\scriptsize, inner sep=2.5pt},
  grp/.style={draw=black!25, rounded corners=4pt, inner sep=4pt},
  flow/.style={-{Latex[length=1.6mm]}, black!60, line width=0.5pt},
  person/.pic={
    \fill[#1] (0,0.42) circle (0.075);
    \draw[#1, line width=1.3pt, line cap=round]
      (0,0.33)--(0,0.15) (0,0.15)--(-0.07,0) (0,0.15)--(0.08,0.01)
      (0,0.29)--(0.11,0.2) (0,0.29)--(-0.09,0.21);
  },
}
\newtheorem{definition}{Definition}
\newtheorem{proposition}{Proposition}

\theoremstyle{remark}

\theoremstyle{plain}

\newcommand{\R}{\mathbb{R}}
\newcommand{\Enc}{\mathcal{E}_\theta}
\newcommand{\Sw}{\Pi}
\newcommand{\bank}{\mathcal{B}}
\newcommand{\view}{v}
\newcommand{\Views}{\mathcal{V}}
\newcommand{\Ovr}{\mathsf{Ovr}}
\newcommand{\Pone}{\mathsf{P1}}
\newcommand{\Ptwo}{\mathsf{P2}}
\newcommand{\Rel}{\mathsf{Rel}}
\newcommand{\Tpi}{\mathcal{T}_{\Sw}}
\newcommand{\Zone}{\bm{Z}_1}
\newcommand{\Ztwo}{\bm{Z}_2}
\newcommand{\Zrel}{\bm{Z}_{12}}
\newcommand{\Zler}{\bm{Z}_{21}}
\newcommand{\Sone}{\bm{S}_1}
\newcommand{\Stwo}{\bm{S}_2}
\newcommand{\LN}{\operatorname{LN}}
\newcommand{\FFN}{\operatorname{FFN}}

\input{figures/qual_base_macros}

%% file: figures/qual_base_macros.tex
\providecolor{pone}{RGB}{42,120,214}     
\providecolor{ptwo}{RGB}{214,120,60}     
\definecolor{qokc}{RGB}{26,127,75}
\definecolor{qbadc}{RGB}{203,36,49}
\definecolor{qours}{RGB}{232,240,252}    
\definecolor{qlgray}{RGB}{120,120,120}
\providecommand{\Pone}{\textsc{P1}}\providecommand{\Ptwo}{\textsc{P2}}\providecommand{\Rel}{\textsc{Rel}}
\providecommand{\qA}{\textcolor{pone}{\mbox{Person\,1}}}
\providecommand{\qB}{\textcolor{ptwo}{\mbox{Person\,2}}}
\providecommand{\qel}{\textcolor{qlgray}{\ldots}}                  
\providecommand{\qyes}{\,\textcolor{qokc}{\ding{51}}}
\providecommand{\qno}{\,\textcolor{qbadc}{\ding{55}}}
\providecommand{\qg}[1]{\textcolor{qokc}{\underline{#1}}}          
\providecommand{\qb}[1]{\textcolor{qbadc}{\underline{#1}}}         
\providecommand{\qm}[1]{\textbf{#1}}
\providecommand{\qvA}{\textcolor{pone}{\textbf{\scriptsize P1}}}   
\providecommand{\qvB}{\textcolor{ptwo}{\textbf{\scriptsize P2}}}
\providecommand{\qvR}{\textcolor{qlgray}{\textbf{\scriptsize Rel}}}
\newcolumntype{R}[1]{>{\raggedright\arraybackslash}p{#1}}

%% file: sec/0_abstract.tex

\begin{abstract}
Describing a two-person interaction requires knowing not only what
happens but who does what to whom.
We find that current motion--language models largely lack this
ability: a strong two-person baseline gives both people the same
description in 67.8\% of interactions, and when the two motion streams
are exchanged, its participant and relation descriptions stay attached
to the input slots instead of following the people.
We formalize participant grounding as a swap-consistency law over four
semantic views, introduce reference-free counterfactual and probing
diagnostics, and attribute the failure to fusing the participants before
the requested view is known (provably fatal in the invariant limit).
We propose \emph{late binding}, instantiated in \textbf{InterRole}: a
Role-Resolved Interaction Learner keeps partner-conditioned participant
evidence and ordered pair evidence separately addressable, and a
Contextual Evidence Reader binds it to each requested view before a
LoRA-adapted language model generates the description.
InterRole reduces
identical participant descriptions to 19.7\%, makes relation
descriptions follow the swapped direction in 87.1\% of asymmetric
interactions (baselines: at most 26.5\%), improves generation on all
four views, consistently improves over the baseline zero-shot on
InterHuman (identical participant descriptions: 74.6\% to 27.9\%), and
its representation supports downstream applications such as role-aware
retrieval and dual-humanoid robot interaction.
Project page: \url{https://wulin97.github.io/interrole}.
\end{abstract}

%% file: sec/1_intro.tex

\section{Introduction}
\label{sec:intro}

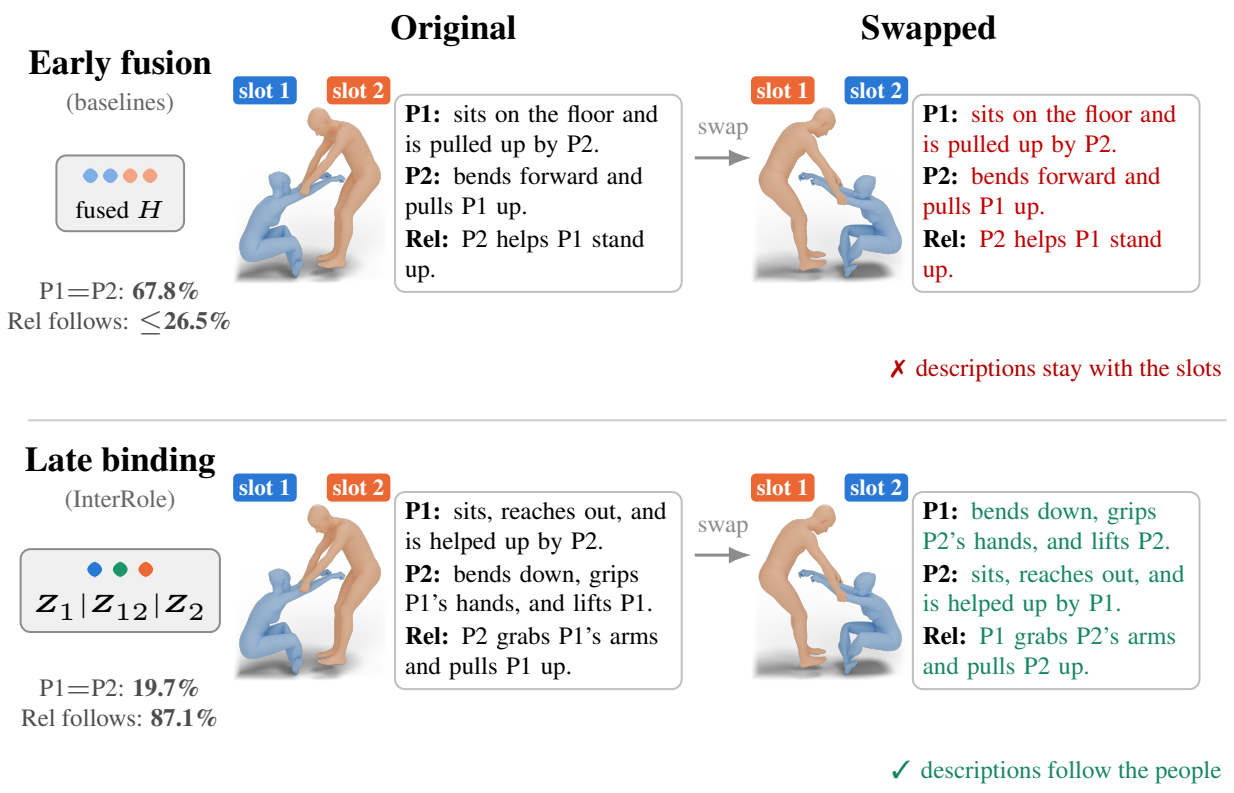
\begin{figure}[t]
\centering
\resizebox{\linewidth}{!}{\input{figures/teaser}}
\caption{
\textbf{Who did what?}
Models that fuse the two participants before the view is known (top) keep
their descriptions when the streams are exchanged: they follow the slots,
not the people.
InterRole (bottom) binds evidence to the view only at the end (\emph{late
binding}), so its descriptions follow the people.
Left: share of interactions where both people get the same description
(P1$=$P2) and where the relation follows the swap.
}
\label{fig:teaser}
\end{figure}

Understanding a two-person interaction means more than naming it.
``A hug'' or ``a push'' identifies the event, but most downstream uses
also need to know \emph{who did what}: which person initiated the
contact, how the other responded, and in which direction the relation
runs.
Two-person datasets now provide rich interaction
descriptions~\cite{interx,interhuman}, and coupling motion encoders
with large language models (LLM) has made fluent motion captioning
possible~\cite{motiongpt,motionllm,llamo,llammo}.
Whether these models attribute the right behaviour to the right
person, however, is rarely tested.

We test it with a simple counterfactual: describe each participant,
exchange the two motion streams, and describe again.
A model that grounds its descriptions in the participants should
\emph{exchange} its two descriptions; a model that has attached them to
the input slots will repeat itself.
For all baselines we evaluate, including a strong two-person
model~\cite{llammo}, the outputs before and after the swap are almost
as similar under the \emph{wrong} correspondence as under the correct
one, and the strongest baseline gives
literally the same description to both people in 67.8\% of test
interactions.
Linear probes on the frozen motion representations show the same
failure one level deeper: directed-role predictions flip under the swap
\emph{less} often than chance.
Existing models recognize the interaction, but not who is doing what.

The common design fuses the two participants into a shared
representation and only then tells the language model which view to
produce.
We attribute the failure to this order of operations: if the fused
representation is invariant to exchanging the participants, no later
instruction could recover which behaviour belongs to whom, and the model
could only give both people the same description (Prop.~\ref{prop:fusion}).
The baselines are nearly, not strictly, invariant
(Sec.~\ref{sec:factorial_ablation}).
We therefore propose \emph{late binding}: abstract the interaction
jointly, but keep the evidence re-bindable to each participant until
the requested view is known.
\textbf{InterRole} instantiates it with a \emph{Role-Resolved
Interaction Learner}, which builds an evidence bank of two
partner-conditioned participant streams and an ordered pair stream, and
a \emph{Contextual Evidence Reader}, which reads this bank differently
for each of four semantic views (the joint event, each participant, and
their relation) before conditioning a LoRA-adapted language model.
The structure makes exchanging the inputs change only nominal
identifiers while preserving which evidence plays the target and
partner roles (Prop.~\ref{prop:equivariance}); whether the model
exploits this is left to training.

InterRole substantially reduces the failure.
Its swap gain rises from at most 0.029 to 0.214, identical person
descriptions drop from 67.8\% to 19.7\%, and on asymmetric relations
its descriptions follow the swapped direction in 87.1\% of
interactions versus at most 26.5\% for the baselines, while generation
improves on all four views.
Giving the strongest baseline the same target/partner role embeddings,
or training it with swap augmentation, does not close the gap, whereas
applying late binding to it does lift its swap gain from 0.029
to 0.210, so the gain comes from late binding rather than role markers or
augmentation; InterRole's learned evidence and reading add semantics on top
(action probe BAcc 0.596 vs.\ 0.477).
Zero-shot on InterHuman, where reference-free swap diagnostics need no
participant-level annotation, InterRole again improves consistently over the
baseline.

\noindent Our contributions are:
\begin{itemize}
\item We identify a participant-grounding failure in two-person
motion--language models that caption metrics do not reveal, formalize
grounding as a swap-consistency law over four views, show why strictly
invariant early fusion cannot satisfy it, and provide counterfactual and probing
diagnostics.
\item We propose InterRole, guided by late binding: it keeps participant
evidence traceable and binds it to the view only at the end, so its
evidence carries the swap in a structured form.
\item InterRole substantially improves grounding and multi-view generation,
consistently improves over the baseline zero-shot on an unseen dataset, and
its representation supports downstream applications such as role-aware
retrieval and dual-humanoid robot interaction.
\end{itemize}

%% file: figures/teaser.tex
\begin{tikzpicture}[x=1cm,y=1cm, font=\scriptsize]
\newcommand{\bad}[1]{\textcolor{red!75!black}{#1}}
\newcommand{\good}[1]{\textcolor{prel!90!black}{#1}}
\tikzset{stag/.style={fill=#1, text=white, rounded corners=1pt, inner sep=1.2pt,
  font=\fontsize{4.6}{4.6}\selectfont\bfseries, anchor=south}}
\newcommand{\scene}[3]{%
  \ifnum#3=0
    \node[inner sep=0pt] at (#1,#2) {\includegraphics[height=1.27cm]{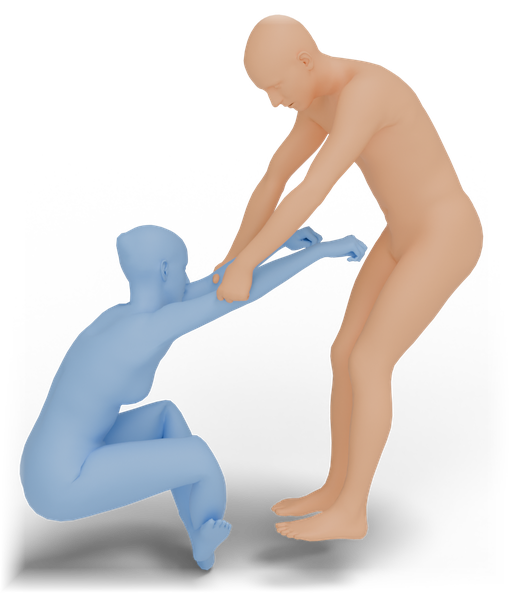}};
    \node[stag=pone] at (#1-0.33,#2+0.62) {slot 1};
    \node[stag=ptwo] at (#1+0.33,#2+0.62) {slot 2};
  \else
    \node[inner sep=0pt] at (#1,#2) {\includegraphics[height=1.27cm]{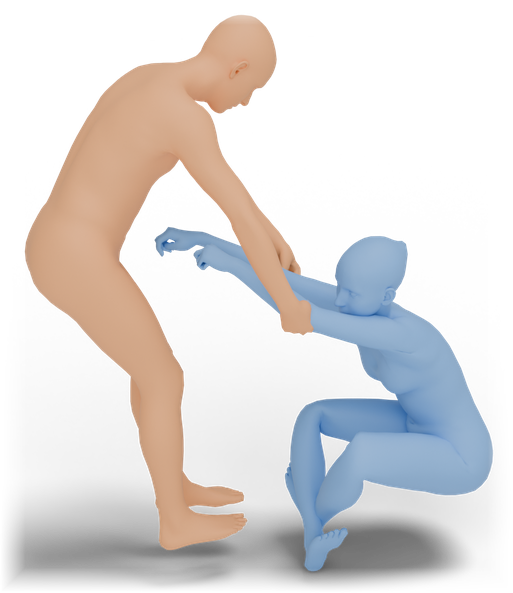}};
    \node[stag=ptwo] at (#1-0.33,#2+0.62) {slot 1};
    \node[stag=pone] at (#1+0.33,#2+0.62) {slot 2};
  \fi}
\newcommand{\outs}[5]{%
  \node[draw=black!25, rounded corners=2pt, fill=white, align=left, font=\tiny,
        anchor=west, inner sep=2pt, text width=1.88cm] at (#1,#2)
  {\textbf{P1:} #3\\[0.5pt]\textbf{P2:} #4\\[0.5pt]\textbf{Rel:} #5};}
\node[font=\scriptsize\bfseries] at (2.9,5.95) {Original};
\node[font=\scriptsize\bfseries] at (6.25,5.95) {Swapped};
\node[font=\scriptsize\bfseries] at (0.55,5.7) {Early fusion};
\node[font=\tiny, black!60] at (0.55,5.43) {(baselines)};
\node[blk, minimum width=0.9cm, fill=black!6] at (0.55,4.8) {%
  \tikz{\foreach \k/\c in {0/pone,1/pone,2/ptwo,3/ptwo}{\fill[\c!60] (\k*0.14,0) rectangle +(0.11,0.11);}}\\[-1pt]{\tiny fused $H$}};
\node[font=\tiny, align=center, anchor=north, black!70] at (0.55,4.3) {P1$=$P2: \textbf{67.8\%}\\Rel follows: \textbf{$\le$26.5\%}};
\scene{1.9}{4.8}{0}
\scene{5.55}{4.8}{1}
\outs{2.48}{4.8}{sits on the floor and is pulled up by P2.}{bends forward and pulls P1~up.}{P2 helps P1 stand up.}
\outs{6.13}{4.8}{\bad{sits on the floor and is pulled up by P2.}}{\bad{bends forward and pulls P1~up.}}{\bad{P2 helps P1 stand up.}}
\node[font=\tiny, red!70!black, align=center] at (7.15,3.55) {\ding{55}\, descriptions stay with the slots};
\draw[black!20] (-0.1,3.2)--(8.35,3.2);
\node[font=\scriptsize\bfseries] at (0.55,2.9) {Late binding};
\node[font=\tiny, black!60] at (0.55,2.63) {(InterRole)};
\node[blk, minimum width=0.9cm, fill=black!6] at (0.55,2.0) {%
  \tikz{\fill[pone] (0,0) rectangle +(0.11,0.11); \fill[prel] (0.18,0) rectangle +(0.11,0.11);
        \fill[ptwo] (0.36,0) rectangle +(0.11,0.11);}\\[-1pt]{\tiny $\bm Z_1|\bm Z_{12}|\bm Z_2$}};
\node[font=\tiny, align=center, anchor=north, black!70] at (0.55,1.5) {P1$=$P2: \textbf{19.7\%}\\Rel follows: \textbf{87.1\%}};
\scene{1.9}{2.0}{0}
\scene{5.55}{2.0}{1}
\outs{2.48}{2.0}{sits, reaches out, and is helped up by P2.}{bends down, grips P1's hands, and lifts P1.}{P2 grabs P1's arms and pulls P1~up.}
\outs{6.13}{2.0}{\good{bends down, grips P2's hands, and lifts P2.}}{\good{sits, reaches out, and is helped up by P1.}}{\good{P1 grabs P2's arms and pulls P2~up.}}
\node[font=\tiny, prel!90!black, align=center] at (7.15,0.72) {\ding{51}\, descriptions follow the people};
\draw[-{Latex[length=1.6mm]}, black!45, line width=0.6pt] (4.6,5.05)--node[above, font=\tiny]{swap} (5.0,5.05);
\draw[-{Latex[length=1.6mm]}, black!45, line width=0.6pt] (4.6,2.25)--node[above, font=\tiny]{swap} (5.0,2.25);
\end{tikzpicture}

%% file: sec/2_related_work.tex

\section{Related Work}
\label{sec:related}

\noindent\textbf{Motion--language understanding.}
Text-annotated motion corpora~\cite{humanml3d} enable motion
captioning~\cite{tm2t}, and coupling motion tokenizers or encoders with
large language models has produced general models that caption, answer
questions about, and reason over
motion~\cite{motiongpt,motionchain,motionllm,motionllm_wu,llamo,mgmotionllm}.
These models describe a single body, so \emph{whom} a description
refers to does not arise; extending them to two people by concatenation
or fusion inherits that single-subject view.

\noindent\textbf{Two-person interaction modelling.}
Inter-X and InterHuman~\cite{interx,interhuman} pair two-person
interactions with text, and the concurrent Inter-X++~\cite{interxpp}
adds causal-order labels.
Most work targets generation~\cite{interhuman,intermask,timotion,ttr}.
Language-model-based \emph{understanding} is more recent:
MIME~\cite{mime} learns a two-stream encoder with person-type
embeddings, VIM~\cite{vim} marks each person's tokens with person
markers, and SocialGen~\cite{socialgen} shuffles participant order as
augmentation.
The closest to our setting, LLaMMo~\cite{llammo}, encodes each person
and models intra- and inter-person dependencies before a language
model, and serves as our main baseline.
We study these mechanisms in the controlled form
(Secs.~\ref{sec:setup} and~\ref{sec:grounding_diagnosis}).
Such methods are evaluated by caption or retrieval quality; to our
knowledge, none tests whether participant-level outputs follow the
participants when they are exchanged.
Our diagnostics are model-agnostic, and InterRole addresses the failure
they reveal by deferring the binding of participant evidence to the
requested view.

\noindent\textbf{Binding, order sensitivity, and symmetry.}
Associating features with the entity they belong to is a long-standing
\emph{binding problem}~\cite{greff2020binding}; in vision--language
models it shows up as captions and retrievals that ignore who does what
to whom~\cite{winoground,aro}.
Identity-aware description in movies~\cite{park2020identity,micap}
relies on appearance cues that skeletal motion lacks; we study the
motion analogue, where the input slot is the only handle on ``who''.
Architectures with built-in permutation symmetry~\cite{deepsets,gcnn}
enforce invariance or equivariance, whereas we need both at once
(equivariance for the participant views, invariance up to relabelling
for the joint and relation views); we show that strictly invariant early
fusion precludes the former.
InterRole obtains participant equivariance by weight sharing and keeps
it attainable, rather than enforced, so that a single model serves all
four views.

%% file: sec/3_method.tex
%
%

\section{Method}
\label{sec:method}

\subsection{Problem Formulation and Framework Overview}
\label{sec:problem}

We represent a two-person interaction as an ordered pair of synchronized
motion streams $\mathcal{M}=(\bm{M}_1,\bm{M}_2)$,
$\bm{M}_i\in\R^{T\times d_{\mathrm{p}}}$.
The same motion supports a reading of the joint event, of each
individual's behaviour, and of their relation, so we define four
\emph{semantic views} $\Views=\{\Ovr,\Pone,\Ptwo,\Rel\}$: $\Ovr$ describes
the whole interaction, $\Pone$ and $\Ptwo$ the participant in each input
slot, and $\Rel$ the contact, coordination and directed
actor--recipient relation.
A single generator $\hat y_\view=F_\theta(\mathcal{M},\view)$ must serve
all four views.

\noindent\textbf{Participant grounding.}
Let $\Sw\mathcal{M}=(\bm{M}_2,\bm{M}_1)$ exchange the input streams.
It swaps the participant views, $\Sw:\Pone\leftrightarrow\Ptwo$, and fixes
$\Ovr$ and $\Rel$; let $\Tpi$ be the corresponding relabelling of
participant references in language (Person~1 $\leftrightarrow$ Person~2).

\begin{definition}[Participant-grounded generator]
\label{def:grounding}
$F_\theta$ is participant-grounded if, for every $\view\in\Views$,
\begin{equation}
F_\theta(\Sw\mathcal{M},\view)\ \approx\ \Tpi\,F_\theta(\mathcal{M},\Sw\view).
\label{eq:consistency}
\end{equation}
\end{definition}

\noindent
For $\Pone,\Ptwo$ this is \emph{equivariance}: after the swap, the $\Pone$
description must describe the person who previously occupied $\Ptwo$.
For $\Ovr,\Rel$ it is \emph{invariance up to relabelling}: ``Person~1
pushes Person~2'' must become ``Person~2 pushes Person~1'', so even a view
without a participant query must retain the \emph{direction} of the
interaction.
Eq.~\eqref{eq:consistency} is necessary but not sufficient (uninformative
outputs can satisfy it), so we also report quality (Swap-Tgt, Sec.~\ref{sec:setup}).

\noindent\textbf{Strictly invariant early fusion precludes grounding.}
A common design fuses both participants before the view is known,
$F_\theta(\mathcal{M},\view)=D(H(\bm{M}_1,\bm{M}_2),\view)$.

\begin{proposition}
\label{prop:fusion}
If $H(\bm{M}_1,\bm{M}_2)=H(\bm{M}_2,\bm{M}_1)$, then
$F_\theta(\Sw\mathcal{M},\Pone)=F_\theta(\mathcal{M},\Pone)$ for every
$\mathcal{M}$; hence Eq.~\eqref{eq:consistency} can hold only if
$F_\theta(\mathcal{M},\Pone)=\Tpi F_\theta(\mathcal{M},\Ptwo)$, i.e.\ only by
giving both participants the same description.
\end{proposition}

\noindent
Intuitively, $D$ receives the same input before and after the swap, so no
instruction applied after $H$ can recover a correspondence that $H$ has
discarded (proof in Supp.~Mat.~\ref{supp:proofs}).
Real models are only nearly invariant (Sec.~\ref{sec:factorial_ablation}),
so this is a limiting case that motivates \emph{late binding}: share
information between the participants early, but keep each piece of evidence
traceable to its source (\emph{preserve correspondence}) and read it as
target, partner or relation evidence only once the view is known
(\emph{defer binding}).


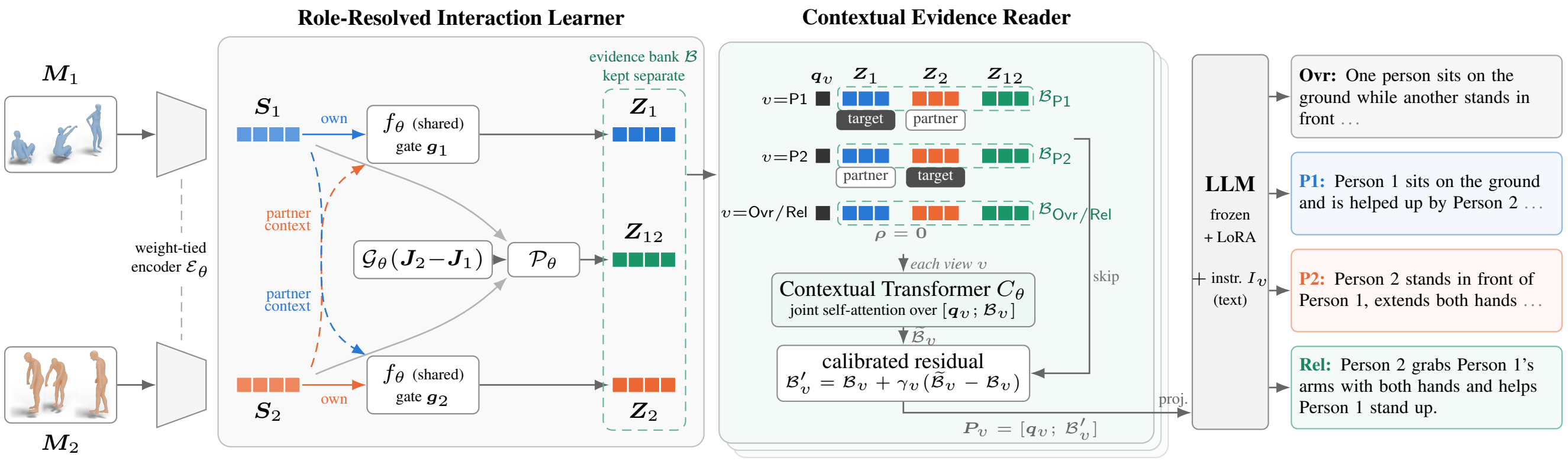
\begin{figure*}[t]
\centering
\input{figures/overview}
\caption{
Overview of InterRole.
A weight-tied encoder maps the two streams to person tokens; the
\emph{Role-Resolved Interaction Learner} builds a bank of partner-conditioned
participant streams $\Zone,\Ztwo$ and an ordered pair stream $\Zrel$, kept
as separate groups.
For each view, the \emph{Contextual Evidence Reader} tags the same bank
with view-dependent target/partner roles (rows; none for $\Ovr$, $\Rel$), reads
$[\bm q_\view;\mathcal B_\view]$ with a contextual Transformer and a
calibrated residual, and projects the prefix $\bm P_\view$ into a
LoRA-adapted language model (\emph{late binding}).
}
\label{fig:overview}
\end{figure*}

InterRole (Fig.~\ref{fig:overview}) realizes late binding with two
components: the \emph{Role-Resolved Interaction Learner}
(Sec.~\ref{sec:rril}) turns the two streams into an evidence bank
$\bank=(\Zone,\Ztwo,\Zrel)$ in which participant-specific and relational
evidence remain separately addressable, and the \emph{Contextual Evidence
Reader} (Sec.~\ref{sec:cer}) reads it differently for each view before it
conditions a LoRA-adapted language model (Sec.~\ref{sec:decode}).
Late binding thus factorizes into $\bank=\mathrm{Learn}(\mathcal M)$ and
$\bm P_\view=\mathrm{Read}(\bank,\view)$: the learner determines
\emph{what evidence exists}, and the reader determines \emph{how it is bound to the query}
(Sec.~\ref{sec:grounding_property} states what this guarantees).

\subsection{Role-Resolved Interaction Learner}
\label{sec:rril}

The learner keeps three evidence streams: participant-conditioned
evidence $\Zone,\Ztwo$ (\emph{how does each person act, given the
partner?}) and ordered pair evidence $\Zrel$ (\emph{what is the
directed relation between them?}).
Layer-level details are provided in Supp.~Mat.~\ref{supp:impl}.

\noindent\textbf{Participant-wise encoding.}
A weight-tied encoder maps each stream independently to $L$ tokens,
$\bm{S}_i=\Enc(\bm{M}_i)\in\R^{L\times d}$, so the two participants are
distinguished by structure rather than by asymmetric parameters.

\noindent\textbf{Participant-conditioned evidence.}
A gesture is defined by its partner: an extended arm reads as a wave, a
block, or a hand-off only once the other person's motion is known.
A single shared operator $f_\theta$ contextualizes each participant by
the other, $\Zone=f_\theta(\Sone,\Stwo)$ and
$\Ztwo=f_\theta(\Stwo,\Sone)$.
In each layer, stream $i$ retrieves partner context from its partner
$j$ by directed cross-attention, with its own tokens as queries, and
admits it through a feature-wise gate,
\begin{equation}
\begin{aligned}
\bm{C}_{i\leftarrow j}&=\mathrm{CrossAttn}_\theta\!\left(\bm{Q}{=}\bm{S}_i,\ \bm{K},\bm{V}{=}\bm{S}_j\right),\\
\bm{g}_i&=\sigma\!\left(\bm{W}_g[\bm{S}_i;\bm{C}_{i\leftarrow j}]\right),\\
\bm{U}_i&=\bm{g}_i\odot\bm{S}_i+(1{-}\bm{g}_i)\odot\bm{C}_{i\leftarrow j},\\
\bm{Z}_i&=\LN\!\left(\bm{U}_i+\FFN(\bm{U}_i)\right),
\end{aligned}
\label{eq:gate_main}
\end{equation}
and $f_\theta$ stacks such layers, with the same parameters and synchronous updates for both directions; $\bm{Z}_i$ is the output of the last layer.
$\Zone$ and $\Ztwo$ are never merged: partner information is shared, but
each stream stays attributable to one person \emph{in context}.

\noindent\textbf{Ordered pair evidence.}
Partner conditioning does not expose the relation itself---spatial
configuration, approach direction, contact geometry---as an object of
description.
We encode it in a dedicated stream built on a \emph{signed} geometric
relation: from joint positions $\bm{J}_i\in\R^{T\times J\times 3}$ we
take the per-frame, joint-wise ordered displacement
$\bm{D}_{1\to 2}=\bm{J}_2-\bm{J}_1$ (so $\bm{D}_{2\to 1}=-\bm{D}_{1\to 2}$),
pool it into $L$ relation tokens with a geometry encoder
$\mathcal{G}_\theta$, and combine them with the participant tokens in a
pair encoder $\mathcal{P}_\theta$,
\begin{equation}
\begin{aligned}
\bm{G}_{12}&=\mathcal{G}_\theta(\bm{D}_{1\to 2}),\\
\Zrel&=\mathcal{P}_\theta\bigl([\Sone;\Stwo;\bm{G}_{12}]\bigr)\in\R^{L\times d}.
\end{aligned}
\label{eq:pair_out}
\end{equation}
The signed displacement and ordered concatenation keep $\Zrel$ ordered
rather than symmetric (``1 pushes 2'' and ``2 pushes 1'' yield
different evidence).
Its role is the relational abstraction---contact, configuration,
coordination---that the participant streams lack
(Sec.~\ref{sec:factorial_ablation}).

\subsection{Contextual Evidence Reader}
\label{sec:cer}

Describing one participant depends on the partner's response and on
their relation, so the reader processes the whole bank jointly for
every view.

\noindent\textbf{Structural identifiers.}
Each participant stream receives a fixed slot embedding $\bm{e}_i$
marking it as the $\Pone$ or $\Ptwo$ stream.
For a participant-centred view $\view=\mathsf{P}k$ we additionally add
a \emph{target} embedding $\bm{r}_{\mathrm{tgt}}$ to the queried stream
$\bm{Z}_k$ and a \emph{partner} embedding $\bm{r}_{\mathrm{ptn}}$ to the
other; neutral views receive no role embedding, and the pair stream
never does.
Writing $\bm{\rho}_{\view,i}$ for the role embedding of stream $i$
under view $\view$ (zero for $\Ovr,\Rel$), the annotated bank is
\begin{equation}
\bank_\view
\!=\!
[
\bm{b}_1, \Zone{+}\bm{e}_1{+}\bm{\rho}_{\view,1}, 
\bm{b}_2, \Ztwo{+}\bm{e}_2{+}\bm{\rho}_{\view,2}, 
\bm{b}_{12}, \Zrel],
\label{eq:reader_input}
\end{equation}
where $\bm{b}_1,\bm{b}_2,\bm{b}_{12}$ are three learned boundary tokens
that open the corresponding stream.
These identifiers are view-dependent annotations added after the
learner, which itself is view-agnostic; the target/partner embeddings
tie the two participant views to a shared notion of ``the person being
described'', independent of which slot that person occupies.

\noindent\textbf{Joint reading and calibrated update.}
A learned view token $\bm{q}_\view=\operatorname{Emb}(\view)$, not part
of the bank, is prepended, $\bm{R}_\view=[\bm{q}_\view;\bank_\view]$,
and an $N$-layer pre-norm \emph{contextual Transformer} $C_\theta$
applies joint self-attention to all tokens; let
$\widetilde{\bank}_\view$ be its outputs at the evidence positions.
Rather than replacing the evidence, the reader applies a
view-dependent calibrated residual,
\begin{equation}
\bank'_\view
=
\bank_\view
+
\gamma_{\view}\bigl(\widetilde{\bank}_\view-\bank_\view\bigr),
\label{eq:calibration}
\end{equation}
where $\gamma_\view\in\R$ is an unconstrained learnable scalar per
view, initialized to $0.01$, so training starts close to the unread
evidence and learns how much contextual refinement each view needs.
Setting $\gamma_\view\equiv 0$ recovers the \emph{static} reader, which
we use as the Static baseline in Sec.~\ref{sec:factorial_ablation}.
The motion prefix keeps the original view token,
$\bm{P}_{\view}=\bigl[\bm{q}_{\view};\ \bank'_\view\bigr]$.

\subsection{Decoding and Objective}
\label{sec:decode}

A learnable projector maps $\bm{P}_\view$ into the embedding space of a
pretrained instruction-tuned language model, where it is prepended to
the textual view instruction $I_\view$.
The language model stays frozen except for low-rank (LoRA)
adapters~\cite{lora}, trained jointly with the motion modules;
$\theta$ collects all trainable parameters.
All four views share one autoregressive objective,
\begin{equation}
\mathcal{L}
=
-\!\!\sum_{(\mathcal{M},\view,y_\view)}\ \sum_{t}
\log p_\theta\!\bigl(y_{\view,t}\mid y_{\view,<t},\bm{P}_\view,I_\view\bigr).
\label{eq:loss}
\end{equation}
InterRole is trained without person-swap augmentation; input swapping is
used only at evaluation (Sec.~\ref{sec:grounding_diagnosis}).
Any grounding therefore arises from the structure and multi-view
supervision; Sec.~\ref{sec:swap_aug} examines what adding such
augmentation changes.

\subsection{Swap Structure of the Evidence}
\label{sec:grounding_property}

Slot embeddings, boundary tokens and position indices belong to
\emph{sequence slots}, not to persons: under the swap, a participant's
evidence moves into the other slot and acquires that slot's identifiers.

\begin{proposition}[The swap acts only on nominal identifiers]
\label{prop:equivariance}
If every layer of $f_\theta$ shares its parameters across the two
directions and updates both from the same pre-update states, then: 
(a) $f_\theta$ commutes with the swap $\Sw$, which maps $(\Zone,\Ztwo)$ to
$(\Ztwo,\Zone)$; and
(b) for every $\view\in\Views$, the reader inputs for $\Sw\mathcal{M}$
under $\view$ and for $\mathcal{M}$ under $\Sw\view$ assign the \emph{same}
participant evidence to the target and partner roles. They differ only in
(i) the slot identity of each participant stream (slot embedding, boundary
token, position indices), (ii) the view token ($\bm{q}_\view$ vs.\
$\bm{q}_{\Sw\view}$, identical for $\Ovr,\Rel$), and (iii) the orientation
of the pair stream ($\Zler$ vs.\ $\Zrel$, computed with $\bm{D}_{2\to1}$
and $\bm{D}_{1\to2}$).
Whenever $\Zone\neq\Ztwo$, the bank is not participant-invariant in the sense of
Prop.~\ref{prop:fusion}.
\end{proposition}

\noindent
The proof is provided in Supp.~Mat.~\ref{supp:proofs}.
Prop.~\ref{prop:equivariance} makes Eq.~\eqref{eq:consistency}
\emph{attainable}, in contrast to Prop.~\ref{prop:fusion}, but does not
enforce it: no training term asks the reader--decoder to treat (i)--(iii)
as the relabelling $\Tpi$.

%% file: figures/overview.tex
\begin{tikzpicture}[x=1cm,y=1cm]
\newcommand{\tokrow}[4]{
  \foreach \k in {1,...,#3}{\node[tok, fill=#4] at ({#1+0.18*(\k-1)},#2) {};}}
\node[blk, minimum width=1.25cm, minimum height=0.85cm] (m1) at (0.7,3.55) {};
\node[inner sep=0pt] at (m1.center) {\includegraphics[height=0.8cm,width=1.2cm,keepaspectratio]{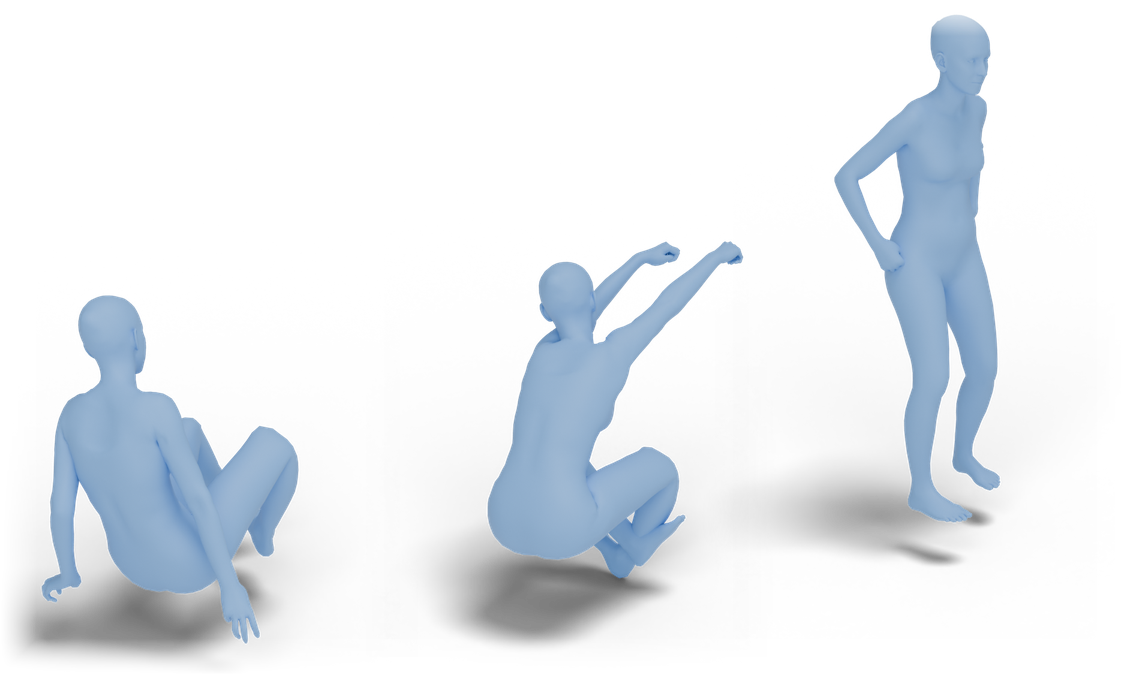}};
\node[font=\scriptsize, anchor=south] at (0.7,3.99) {$\bm M_1$};
\node[blk, minimum width=1.25cm, minimum height=0.85cm] (m2) at (0.7,0.75) {};
\node[inner sep=0pt] at (m2.center) {\includegraphics[height=0.8cm,width=1.2cm,keepaspectratio]{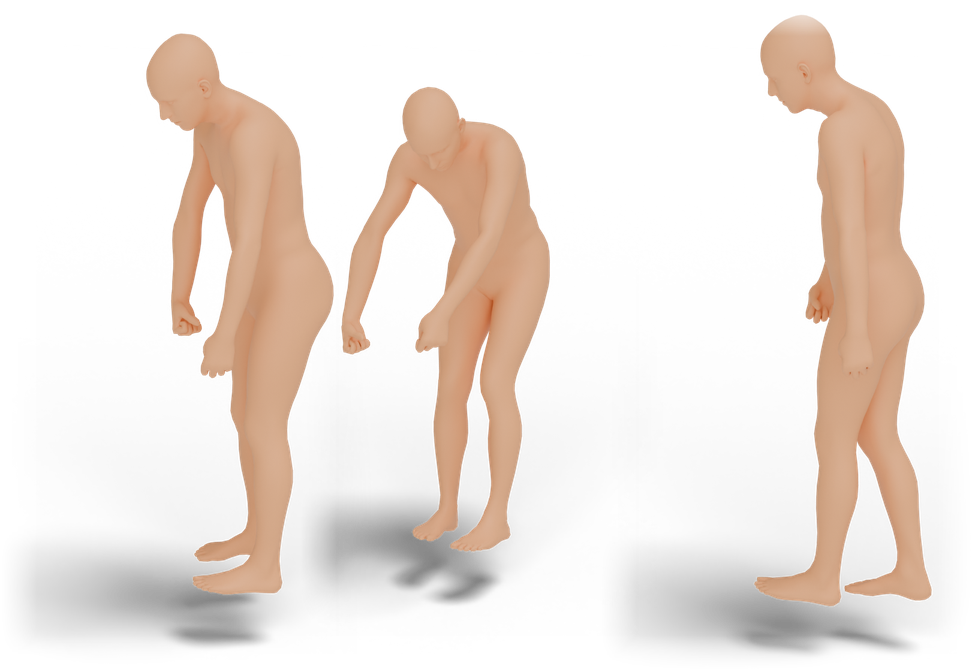}};
\node[font=\scriptsize, anchor=north] at (0.7,0.32) {$\bm M_2$};
\node[trapezium, trapezium angle=70, draw=black!45, fill=black!4,
      rotate=-90, minimum width=0.9cm, minimum height=0.55cm] (e1) at (2.05,3.55) {};
\node[trapezium, trapezium angle=70, draw=black!45, fill=black!4,
      rotate=-90, minimum width=0.9cm, minimum height=0.55cm] (e2) at (2.05,0.75) {};
\node[font=\tiny, align=center, inner sep=1pt] at (1.93,2.15) {weight-tied\\encoder $\mathcal{E}_\theta$};
\draw[black!35, densely dashed] (2.05,3.05)--(2.05,2.45) (2.05,1.85)--(2.05,1.25);
\draw[flow] (m1)--(e1); \draw[flow] (m2)--(e2);
\tokrow{2.75}{3.55}{4}{pone!75}  \node[font=\scriptsize] at (3.02,3.85) {$\bm S_1$};
\tokrow{2.75}{0.75}{4}{ptwo!80}  \node[font=\scriptsize] at (3.02,0.45) {$\bm S_2$};
\node[blk, minimum width=1.25cm] (f1) at (4.75,3.55) {$f_\theta$ {\tiny(shared)}\\[-1pt]{\tiny gate $\bm g_1$}};
\node[blk, minimum width=1.25cm] (f2) at (4.75,0.75) {$f_\theta$ {\tiny(shared)}\\[-1pt]{\tiny gate $\bm g_2$}};
\draw[flow, pone] (3.4,3.55)--node[above, font=\tiny]{own}(f1.west);
\draw[flow, ptwo] (3.4,0.75)--node[below, font=\tiny]{own}(f2.west);
\draw[flow, ptwo, densely dashed] (3.5,0.95) to[out=70,in=210] (f1.south west);
\draw[flow, pone, densely dashed] (3.5,3.35) to[out=-70,in=150] (f2.north west);
\node[font=\tiny, ptwo, anchor=east, align=right] at (3.62,2.6) {partner\\[-2pt]context};
\node[font=\tiny, pone, anchor=east, align=right] at (3.62,1.7) {partner\\[-2pt]context};
\node[blk] (g) at (4.75,2.15) {$\mathcal{G}_\theta(\bm J_2{-}\bm J_1)$};
\node[blk, minimum width=0.8cm] (pp) at (6.1,2.15) {$\mathcal{P}_\theta$};
\draw[flow] (g)--(pp);
\draw[flow, black!30] (3.55,3.43) to[out=-23,in=137] (pp.north west);
\draw[flow, black!30] (3.55,0.87) to[out=23,in=-137] (pp.south west);
\tokrow{6.95}{3.55}{4}{pone}   \node[font=\scriptsize] at (7.22,3.85) {$\bm Z_1$};
\tokrow{6.95}{2.15}{4}{prel}
\node[font=\scriptsize] at (7.22,2.45) {$\bm Z_{12}$};
\tokrow{6.95}{0.75}{4}{ptwo}   \node[font=\scriptsize] at (7.22,0.45) {$\bm Z_2$};
\draw[flow] (f1.east)--(6.83,3.55);
\draw[flow] (pp.east)--(6.83,2.15);
\draw[flow] (f2.east)--(6.83,0.75);
\node[draw=prel!70, densely dashed, rounded corners=3pt, fit={(6.8,0.28) (7.65,4.02)}, inner sep=1pt] (bank) {};
\node[font=\tiny, prel!80!black, align=center, anchor=south, inner sep=1pt] at (7.22,4.07) {evidence bank $\mathcal{B}$\\kept separate};
\begin{scope}[on background layer]
\node[grp, fill=black!2.5, fit={(2.6,0.2) (7.75,4.5)}] (learner) {};
\end{scope}
\node[font=\scriptsize\bfseries, anchor=south] at (5.175,4.65) {Role-Resolved Interaction Learner};
\tikzset{
  tgt/.style={rounded corners=1.5pt, fill=black!70, text=white, font=\tiny,
              inner sep=1.2pt, minimum width=0.66cm},
  ptn/.style={rounded corners=1.5pt, draw=black!45, fill=white, text=black!75, font=\tiny,
              inner sep=1.2pt, minimum width=0.66cm},
  nor/.style={font=\tiny, black!45},
}
\draw[flow] (7.70,3.10)--(8.04,3.10);
\begin{scope}[on background layer]
\node[fill=white, draw=black!15, rounded corners=2pt, fit={(8.28,2.22) (12.02,4.36)}, inner sep=0pt] (bankv) {};
\end{scope}
\node[font=\tiny, anchor=south, inner sep=1pt] at (9.22,4.08) {$\bm q_\view$};
\node[font=\tiny, anchor=south, inner sep=1pt] at (9.70,4.08) {$\bm Z_1$};
\node[font=\tiny, anchor=south, inner sep=1pt] at (10.48,4.08) {$\bm Z_2$};
\node[font=\tiny, anchor=south, inner sep=1pt] at (11.26,4.08) {$\bm Z_{12}$};
\newcommand{\viewrow}[6]{%
  \node[font=\tiny, anchor=east, inner sep=0pt] at (9.06,#1) {$\view{=}#2$};
  \node[tok, fill=black!80] at (9.22,#1) {};
  \tokrow{9.52}{#1}{3}{pone}\tokrow{10.30}{#1}{3}{ptwo}\tokrow{11.08}{#1}{3}{prel}
  \node[draw=prel!70, densely dashed, rounded corners=1.5pt, inner sep=1.2pt,
        fit={(9.43,#1-0.09) (11.53,#1+0.09)}] (bvrow) {};
  \node[font=\tiny, prel!80!black, anchor=west, inner sep=1pt] at (bvrow.east) {$\mathcal{B}_{#2}$};
  \node[#3] at (9.70,#1-0.24) {#4};
  \node[#5] at (10.48,#1-0.24) {#6};}
\viewrow{3.95}{\Pone}{tgt}{target}{ptn}{partner}
\viewrow{3.31}{\Ptwo}{ptn}{partner}{tgt}{target}
\viewrow{2.67}{\Ovr/\Rel}{nor}{}{nor}{}
\node[font=\tiny, black!55] at (10.09,2.43) {$\bm\rho=\bm 0$};
\node[blk, minimum width=2.7cm, fill=prel!8, font=\tiny] (ctx) at (10.12,1.70)
  {{\scriptsize Contextual Transformer $C_\theta$}\\joint self-attention over $[\bm q_\view;\mathcal{B}_\view]$};
\draw[flow] (10.12,0 |- bankv.south)--node[right, font=\tiny\itshape, black!55, inner sep=2pt]{each view $\view$}(ctx.north);
\node[blk, minimum width=2.7cm, font=\tiny] (res) at (10.12,0.88)
  {{\scriptsize calibrated residual}\\$\mathcal{B}'_\view=\mathcal{B}_\view+\gamma_\view\bigl(\widetilde{\mathcal{B}}_\view-\mathcal{B}_\view\bigr)$};
\draw[flow] (ctx)--node[right, font=\tiny, inner sep=2pt]{$\widetilde{\mathcal{B}}_\view$}(res);
\draw[flow] (bankv.east |- 0,3.50)-|(12.20,1.20)|-(res.east);
\node[font=\tiny, black!55, anchor=west, inner sep=1pt] at (12.20,1.95) {skip};
\begin{scope}[on background layer]
\node[grp, draw=black!12, fill=prel!2, fit={(8.36,0.08) (12.94,4.28)}] {};
\node[grp, draw=black!18, fill=prel!3, fit={(8.28,0.16) (12.86,4.36)}] {};
\node[grp, fill=prel!5, fit={(8.2,0.24) (12.78,4.44)}] (reader) {};
\end{scope}
\node[font=\scriptsize\bfseries, anchor=south] at (10.49,4.65) {Contextual Evidence Reader};
\tikzset{outbox/.style={rounded corners=2pt, align=left, anchor=north west,
  font=\fontsize{5.8}{6.8}\selectfont, inner sep=2.5pt, text width=2.85cm,
  minimum height=0.92cm, execute at begin node={\hyphenpenalty=10000}}}
\node[outbox, draw=black!35, fill=black!3] (o1) at (14.45,4.44) {%
  \textbf{Ovr:} One person sits on the ground while another stands in front \textcolor{black!50}{\ldots}};
\node[outbox, draw=pone!55, fill=pone!5] (o2) at ($(o1.south west)+(0,-0.15)$) {%
  \textcolor{pone}{\textbf{P1:}} Person~1 sits on the ground and is helped up by Person~2 \textcolor{black!50}{\ldots}};
\node[outbox, draw=ptwo!55, fill=ptwo!5] (o3) at ($(o2.south west)+(0,-0.15)$) {%
  \textcolor{ptwo}{\textbf{P2:}} Person~2 stands in front of Person~1, extends both hands \textcolor{black!50}{\ldots}};
\node[outbox, draw=prel!60, fill=prel!5] (o4) at ($(o3.south west)+(0,-0.15)$) {%
  \textcolor{prel!90!black}{\textbf{Rel:}} Person~2 grabs Person~1's arms with both hands and helps Person~1 stand up.};
\node[blk, fill=black!5, inner sep=0pt, fit={(13.35,0 |- o4.south) (14.2,0 |- o1.north)}] (llm) {};
\node[align=center, font=\scriptsize, inner sep=0pt] at (llm.center)
  {\textbf{LLM}\\[1pt]{\tiny frozen}\\[-1pt]{\tiny + LoRA}\\[5pt]{\tiny $+$ instr.\ $I_\view$}\\[-1pt]{\tiny (text)}};
\foreach \o in {o1,o2,o3,o4}{\draw[flow] (llm.east |- \o.west)--(\o.west);}
\draw[flow] (res.south) |- node[pos=0.72, below, font=\tiny, inner sep=1.5pt]{$\bm P_\view=[\bm q_\view;\,\mathcal{B}'_\view]$} (llm.west |- 0,0.44);
\node[font=\tiny, black!60, anchor=south, inner sep=1.5pt] at (13.14,0.44) {proj.};
\end{tikzpicture}

%% file: sec/4_experiment.tex
\section{Experiments}
\label{sec:experiments}

\subsection{Experimental Setup}
\label{sec:setup}

\noindent\textbf{Data and semantic views.}
We use Inter-X~\cite{interx} with fixed splits of 9.1k / 0.6k / 1.7k for
train / validation / test sequences shared by all methods.
An LLM-based pipeline derives the four views of Sec.~\ref{sec:problem}
from each source caption (automatic supervision, manually checked on
randomly sampled test sequences; Supp.~Mat.~\ref{supp:data}), and
relations are tagged asymmetric, reciprocal, symmetric, or role-neutral
for stratified analysis.
A sample lacking a target for a view is excluded from that view only
(about 1.7k test sequences for Overall/Relation, 1.5k for P1/P2).
For zero-shot transfer (Sec.~\ref{sec:cross_dataset}), we also use the
1{,}176 test sequences of InterHuman~\cite{interhuman}, which has only
Overall captions.

\noindent\textbf{Compared methods.}
All methods share the splits, targets, language model, LoRA setup,
schedule and decoding, with equal-size motion Transformers; they differ
only in how participant and view information is routed.
\emph{Instruction Conditioning} specifies the view by a textual
instruction over a shared motion representation, and \emph{Task-Query
Conditioning} adds a learned view token (the simplest two-stream encoder
with a task token, close to MIME~\cite{mime}); both separate the two
people only by concatenation order.
\emph{LLaMMo}~\cite{llammo} (our reimplementation, as no official code is
available) models intra- and inter-person dependencies with
temporal/social attention and adds a learned slot-identity embedding to
each person.
\emph{InterRole} adds slot identity and, for the $\Pone/\Ptwo$ views,
target/partner embeddings (Sec.~\ref{sec:cer}).
Controls that instantiate prior mechanisms in this same pipeline (person
markers, late binding, swap augmentation) are introduced in
Sec.~\ref{sec:factorial_ablation}, so that differences reflect how
participant information is routed; the main comparison uses no swap augmentation.

\noindent\textbf{Metrics.}
Generation is evaluated per view against all references with
BERTScore-F1~\cite{bertscore} (primary; RoBERTa-large, baseline-rescaled),
BLEU-4~\cite{bleu} and CIDEr~\cite{cider}.
\emph{Canonical Reference Agreement} scores each held-out canonical
candidate against the others (reference consistency).
For grounding, every test interaction $x$ is also fed as the swapped input
$\tau x=(x_2,x_1)$ (Sec.~\ref{sec:problem}); participant-centred references
exchange and directed-role labels reverse.
Let $s$ be BERTScore-F1; the original outputs are first relabelled by
$\Tpi$, since a description may mention both participants.
$S_{\mathrm{correct}}$ compares original and swapped outputs about the same
person (in the other slot), $S_{\mathrm{wrong}}$ about the other person
(same slot), and $G_{\mathrm{swap}}=S_{\mathrm{correct}}-S_{\mathrm{wrong}}$
measures participant selectivity; \emph{Swap-Tgt} scores the swapped
outputs against the exchanged references and measures absolute quality
(exact definitions in Supp.~Mat.~\ref{supp:data}).
\emph{Frozen linear probes}~\cite{alain2016probes} classify the frozen
motion representation: the action probe should be invariant under
swapping and the directed-role probe should reverse.
We report balanced accuracy (BAcc) on original and swapped inputs and pair
correctness (Pair-C), which requires both members of an
original--swapped pair to be correct; invariance (Inv.) and equivariance
(Equiv.) can be satisfied by consistently wrong predictions, so they are
read with Pair-C.

\noindent\textbf{Implementation.}
We use Qwen2.5-1.5B-Instruct~\cite{qwen25} and Gemma3-1B-IT~\cite{gemma3}
with frozen base weights and LoRA~\cite{lora} adapters (rank 16), a 4-layer,
768-wide motion Transformer on 30-fps SMPL-X features, and a 10-epoch budget
with the checkpoint of lowest validation loss (details, including trainable
parameter counts, are in Supp.~Mat.~\ref{supp:impl}).

\subsection{Quantitative and Qualitative Comparison}
\label{sec:main_results}

\begin{table*}[t]
\centering
\small
\renewcommand{\arraystretch}{1.15}
\caption{
Multi-view generation on the test set.
Canonical Reference Agreement$^\dagger$: consistency among held-out
references, not a trained model or an upper bound.
All metrics higher is better; best trained result per backbone in \textbf{bold}.
}
\label{tab:main_multiview}
\resizebox{0.95\textwidth}{!}{%
\begin{tabular}{ll|ccc|ccc|ccc|ccc}
\toprule
\multirow{2}{*}{Backbone} & \multirow{2}{*}{Method}
& \multicolumn{3}{c|}{\textsc{Overall}}
& \multicolumn{3}{c|}{\textsc{P1}}
& \multicolumn{3}{c|}{\textsc{P2}}
& \multicolumn{3}{c}{\textsc{Relation}} \\
\cmidrule(lr){3-5}\cmidrule(lr){6-8}\cmidrule(lr){9-11}\cmidrule(lr){12-14}
& & BERT$\uparrow$ & BLEU-4$\uparrow$ & CIDEr$\uparrow$ & BERT$\uparrow$ & BLEU-4$\uparrow$ & CIDEr$\uparrow$
  & BERT$\uparrow$ & BLEU-4$\uparrow$ & CIDEr$\uparrow$ & BERT$\uparrow$ & BLEU-4$\uparrow$ & CIDEr$\uparrow$ \\
\midrule
\multicolumn{2}{l|}{\color{gray}Canonical Ref.\ Agreement$^\dagger$}
& \color{gray}0.580 & \color{gray}0.270 & \color{gray}0.456
& \color{gray}0.697 & \color{gray}0.400 & \color{gray}1.293
& \color{gray}0.695 & \color{gray}0.396 & \color{gray}1.313
& \color{gray}0.697 & \color{gray}0.435 & \color{gray}1.362 \\
\midrule
\multirow{5}{*}{Qwen2.5-1.5B}
& Instruction Cond.
& 0.499 & 0.241 & 0.158 & 0.544 & 0.217 & 0.281 & 0.549 & 0.214 & 0.285 & 0.521 & 0.196 & 0.269 \\
& Task-Query Cond.
& 0.500 & 0.238 & 0.157 & 0.536 & 0.216 & 0.285 & 0.541 & 0.213 & 0.284 & 0.520 & 0.192 & 0.259 \\
& LLaMMo
& 0.513 & 0.250 & 0.177 & 0.547 & 0.226 & 0.328 & 0.552 & 0.226 & 0.329 & 0.538 & 0.212 & 0.316 \\
& \textbf{InterRole}
& \textbf{0.548} & \textbf{0.299} & \textbf{0.252}
& \textbf{0.597} & \textbf{0.296} & \textbf{0.504}
& \textbf{0.605} & \textbf{0.312} & \textbf{0.531}
& \textbf{0.595} & \textbf{0.301} & \textbf{0.534} \\
\midrule
\multirow{5}{*}{Gemma3-1B}
& Instruction Cond.
& 0.505 & 0.228 & 0.126 & 0.513 & 0.159 & 0.177 & 0.508 & 0.157 & 0.188 & 0.466 & 0.148 & 0.152 \\
& Task-Query Cond.
& 0.501 & 0.225 & 0.122 & 0.506 & 0.157 & 0.168 & 0.504 & 0.155 & 0.160 & 0.460 & 0.140 & 0.146 \\
& LLaMMo
& 0.506 & 0.235 & 0.144 & 0.514 & 0.164 & 0.186 & 0.510 & 0.160 & 0.193 & 0.483 & 0.163 & 0.188 \\
& \textbf{InterRole}
& \textbf{0.528} & \textbf{0.263} & \textbf{0.197}
& \textbf{0.537} & \textbf{0.191} & \textbf{0.241}
& \textbf{0.537} & \textbf{0.192} & \textbf{0.253}
& \textbf{0.520} & \textbf{0.210} & \textbf{0.295} \\
\bottomrule
\end{tabular}%
}
\end{table*}

\noindent\textbf{Multi-view numerical evaluation.}
Table~\ref{tab:main_multiview} shows that InterRole improves every view
and metric on both backbones.
On Qwen2.5-1.5B, the gain over the strongest baseline (LLaMMo) is
largest on the views that require separating or relating the two
people: CIDEr rises from .328 to .504 on $\Pone$, from .329 to .531 on
$\Ptwo$, and from .316 to .534 on $\Rel$, compared with .177 to .252 on
$\Ovr$; on Gemma3-1B the gain is again largest on Relation.
InterRole thus improves participant-specific and relational
descriptions without sacrificing the joint-event view (BERTScore gap to
Canonical Reference Agreement on $\Ovr$: .032).

\noindent\textbf{Qualitative evaluation.}
Fig.~\ref{fig:qualitative} shows two illustrative cases, bowing and
shoulder massage.
On the original input (a), the P1/P2 and Relation excerpts distinguish
the bowing person from the recipient and the masseur from the seated
recipient.
After swapping the two motion streams (b), InterRole's attributions
follow the physical participants, while LLaMMo repeats its original
outputs.

\input{figures/fig_qualitative}

\subsection{Further Analysis}
\label{sec:grounding_diagnosis}

\begin{table*}[t]
\centering
\small
\setlength{\tabcolsep}{3.0pt}
\renewcommand{\arraystretch}{1.12}
\caption{
Participant-grounding diagnosis under participant swapping.
\textbf{Language Counterfactual}: BERTScore-F1, $G_{\mathrm{swap}}=S_{\mathrm{correct}}-S_{\mathrm{wrong}}$.
\textbf{Frozen Linear Probes}: action predictions should stay invariant, role predictions should reverse; Pair-C requires both members of an original--swapped pair to be correct.
Higher is better except $S_{\mathrm{wrong}}$; best per backbone in \textbf{bold}
}
\label{tab:grounding_diagnosis}
\resizebox{0.95\textwidth}{!}{%
\begin{tabular}{ll|cccc|cccc|cccc}
\toprule
\multirow{3}{*}{Backbone} & \multirow{3}{*}{Method}
& \multicolumn{4}{c|}{Language Counterfactual}
& \multicolumn{8}{c}{Frozen Linear Probes} \\
\cmidrule(lr){7-14}
& & & & & & \multicolumn{4}{c|}{Action} & \multicolumn{4}{c}{Directed Role} \\
\cmidrule(lr){3-6}\cmidrule(lr){7-10}\cmidrule(lr){11-14}
& & Swap-Tgt$\uparrow$ & $S_{\mathrm{correct}}\!\uparrow$
  & $S_{\mathrm{wrong}}\!\downarrow$ & $G_{\mathrm{swap}}\!\uparrow$
  & BAcc$\uparrow$ & Swap BAcc$\uparrow$ & Pair-C$\uparrow$ & Inv.$\uparrow$
  & BAcc$\uparrow$ & Swap BAcc$\uparrow$ & Pair-C$\uparrow$ & Equiv.$\uparrow$ \\
\midrule
\multirow{5}{*}{Qwen2.5-1.5B}
& Instruction Cond.
& 0.539 & \textbf{0.843} & 0.817 & 0.027
& 0.361 & 0.336 & 0.295 & \textbf{0.743}
& 0.542 & 0.504 & 0.109 & 0.172 \\
& Task-Query Cond.
& 0.531 & 0.832 & 0.807 & 0.025
& 0.360 & 0.333 & 0.293 & 0.722
& 0.542 & 0.475 & 0.045 & 0.073 \\
& LLaMMo
& 0.544 & 0.822 & 0.793 & 0.029
& 0.423 & 0.402 & 0.340 & 0.700
& 0.561 & 0.503 & 0.118 & 0.174 \\
& \textbf{InterRole}
& \textbf{0.594} & 0.816 & \textbf{0.603} & \textbf{0.214}
& \textbf{0.596} & \textbf{0.586} & \textbf{0.489} & 0.694
& \textbf{0.723} & \textbf{0.695} & \textbf{0.569} & \textbf{0.722} \\
\midrule
\multirow{5}{*}{Gemma3-1B}
& Instruction Cond.
& 0.504 & 0.840 & 0.821 & 0.020
& 0.311 & 0.289 & 0.247 & \textbf{0.721}
& 0.590 & 0.518 & 0.176 & 0.246 \\
& Task-Query Cond.
& 0.499 & 0.816 & 0.812 & 0.005
& 0.277 & 0.270 & 0.220 & 0.675
& 0.590 & 0.538 & 0.217 & 0.308 \\
& LLaMMo
& 0.508 & \textbf{0.841} & 0.819 & 0.022
& 0.374 & 0.354 & 0.302 & 0.709
& 0.532 & 0.503 & 0.077 & 0.118 \\
& \textbf{InterRole}
& \textbf{0.531} & 0.826 & \textbf{0.774} & \textbf{0.052}
& \textbf{0.529} & \textbf{0.514} & \textbf{0.433} & 0.681
& \textbf{0.711} & \textbf{0.666} & \textbf{0.561} & \textbf{0.743} \\
\bottomrule
\end{tabular}%
}
\vspace{-6pt}
\end{table*}

\begin{figure}[t]
\centering
\begin{tikzpicture}
\begin{axis}[
  width=\linewidth, height=4.5cm,
  ybar stacked, bar width=7pt,
  ymin=0, ymax=100, xmin=-0.8, xmax=18.8,
  ytick={0,25,50,75,100}, ylabel={Share of interactions (\%)},
  xtick={0,1,2,3,5,6,7,8,10,11,12,13,15,16,17,18},
  xticklabels={Instr.,Task-Q.,LLaMMo,\textbf{Ours},Instr.,Task-Q.,LLaMMo,\textbf{Ours},Instr.,Task-Q.,LLaMMo,\textbf{Ours},Instr.,Task-Q.,LLaMMo,\textbf{Ours}},
  xticklabel style={font=\tiny, rotate=90, anchor=east, inner sep=1pt},
  yticklabel style={font=\scriptsize},
  label style={font=\small},
  major x tick style={draw=none},
  ymajorgrids, grid style={black!8},
  axis line style={black!40}, tick style={black!40},
  clip=false,
  legend style={font=\scriptsize, at={(0.5,1.03)}, anchor=south,
    legend columns=3, draw=none, fill=none, column sep=6pt},
  legend image code/.code={\fill[#1] (0cm,-0.08cm) rectangle (0.22cm,0.08cm);},
]
\addplot[fill={rgb,255:red,42;green,120;blue,214}, draw=white, line width=0.5pt,
  error bars/.cd, y dir=both, y explicit,
  error bar style={black!80, line width=0.5pt}, error mark options={rotate=90, mark size=1.2pt, black!80}]
  coordinates {(0,24.8) += (0,2.5) -= (0,2.5) (1,26.0) += (0,2.6) -= (0,2.5) (2,26.5) += (0,2.8) -= (0,2.6) (3,87.1) += (0,1.9) -= (0,2.0) (5,24.1) += (0,5.2) -= (0,4.9) (6,23.7) += (0,5.2) -= (0,4.9) (7,23.3) += (0,5.3) -= (0,4.9) (8,58.3) += (0,5.6) -= (0,5.7) (10,19.6) += (0,4.7) -= (0,5.3) (11,23.0) += (0,5.7) -= (0,5.6) (12,24.8) += (0,5.6) -= (0,5.2) (13,33.9) += (0,6.1) -= (0,6.1) (15,14.5) += (0,8.4) -= (0,7.3) (16,21.7) += (0,8.4) -= (0,8.4) (17,21.7) += (0,9.6) -= (0,8.4) (18,57.8) += (0,10.9) -= (0,10.8)};
\addplot[fill={rgb,255:red,195;green,194;blue,183}, draw=white, line width=0.5pt]
  coordinates {(0,4.7) (1,4.4) (2,4.3) (3,5.7) (5,11.7) (6,12.4) (7,15.0) (8,32.0) (10,21.3) (11,21.7) (12,23.9) (13,60.4) (15,30.1) (16,28.9) (17,33.7) (18,38.6)};
\addplot[fill={rgb,255:red,235;green,104;blue,52}, fill opacity=0.85, draw=white, line width=0.5pt]
  coordinates {(0,70.5) (1,69.6) (2,69.2) (3,7.2) (5,64.2) (6,63.9) (7,61.7) (8,9.7) (10,59.1) (11,55.3) (12,51.3) (13,5.7) (15,55.4) (16,49.4) (17,44.6) (18,3.6)};
\legend{Follows swap, Tie, Keeps direction}
\node[font=\scriptsize, anchor=north] at (axis cs:1.5,0) [yshift=-31pt] {Asymmetric};
\node[font=\tiny, text=black!55, anchor=north] at (axis cs:1.5,0) [yshift=-39pt] {$N{=}1{,}108$};
\node[font=\scriptsize, anchor=north] at (axis cs:6.5,0) [yshift=-31pt] {Reciprocal};
\node[font=\tiny, text=black!55, anchor=north] at (axis cs:6.5,0) [yshift=-39pt] {$N{=}266$};
\node[font=\scriptsize, anchor=north] at (axis cs:11.5,0) [yshift=-31pt] {Symmetric};
\node[font=\tiny, text=black!55, anchor=north] at (axis cs:11.5,0) [yshift=-39pt] {$N{=}230$};
\node[font=\scriptsize, anchor=north] at (axis cs:16.5,0) [yshift=-31pt] {Role-neutral};
\node[font=\tiny, text=black!55, anchor=north] at (axis cs:16.5,0) [yshift=-39pt] {$N{=}83$};
\end{axis}
\end{tikzpicture}
\caption{
Relation-view behaviour under participant swapping (Qwen2.5-1.5B).
An interaction \emph{follows} the swap if the swapped description is closer to the relabelled original, \emph{ties} if it reads the same after relabelling, and \emph{keeps} the original direction otherwise; error bars: 95\% bootstrap CIs.
}
\label{fig:relation_direction}
\vspace{-10pt}
\end{figure}
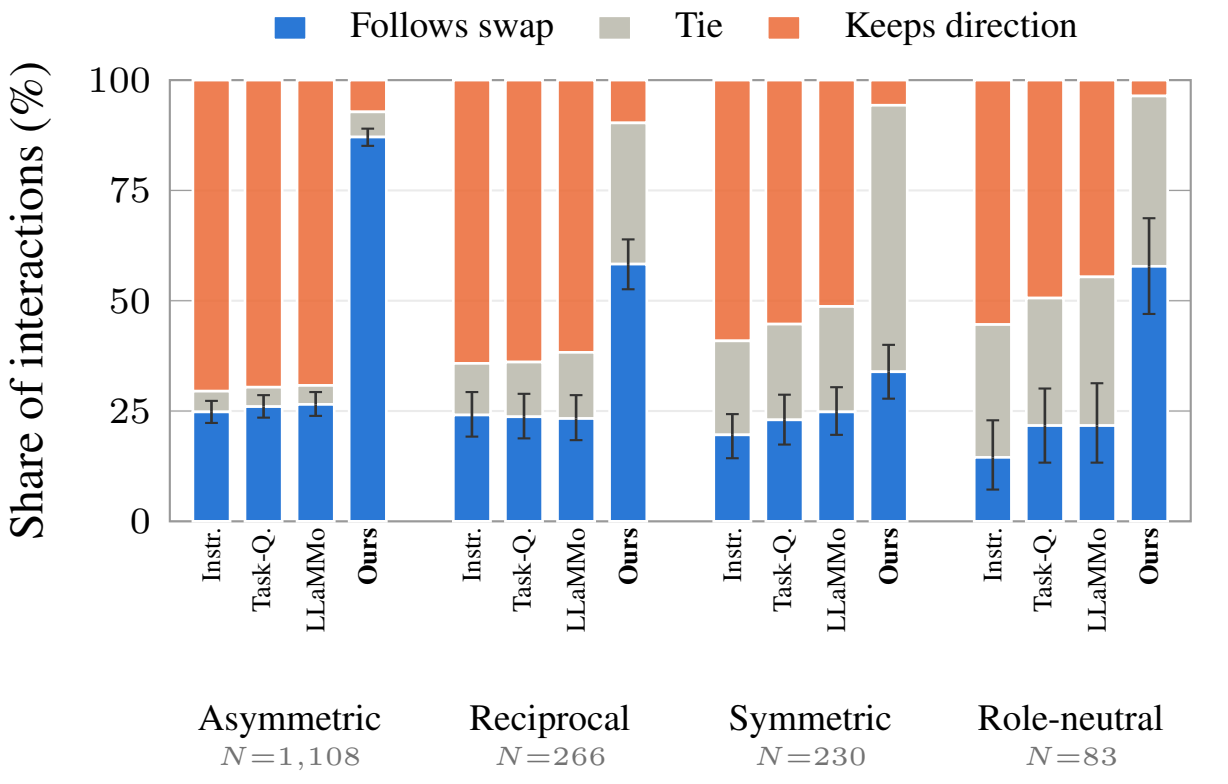

\noindent\textbf{InterRole follows the people.}
We swap the two streams and test three levels.
\emph{Language} (Table~\ref{tab:grounding_diagnosis}): all baselines keep
$S_{\mathrm{correct}}$ high but $S_{\mathrm{wrong}}$ almost as high, so their
swap gains are at most .029; InterRole lowers $S_{\mathrm{wrong}}$
($G_{\mathrm{swap}}=.214$) without losing quality (Swap-Tgt .594 vs.\ at
most .544), whereas LLaMMo gives both people the same description for 67.8\%
of interactions (InterRole: 19.7\%; Fig.~\ref{fig:qualitative}b).
\emph{Direction} (Fig.~\ref{fig:relation_direction}): on asymmetric
relations InterRole's swapped description is closer to the relabelled
original in 87.1\% of interactions, versus at most 26.5\% for the baselines,
which keep the original actor--recipient direction (a rule that reads the
actor off each description gives the same picture, Supp.~Mat.~\ref{supp:data}); on symmetric relations
InterRole mostly produces descriptions that read the same after relabelling
(ties, 60.4\%), as desired.
\emph{Representation}: the baselines' directed-role probes are slot-bound,
with equivariance and Pair-C below chance, whereas InterRole reaches .722 and
.569; Gemma3-1B shows the same pattern, with InterRole ahead on Pair-C (.561)
and swap gain (.052).

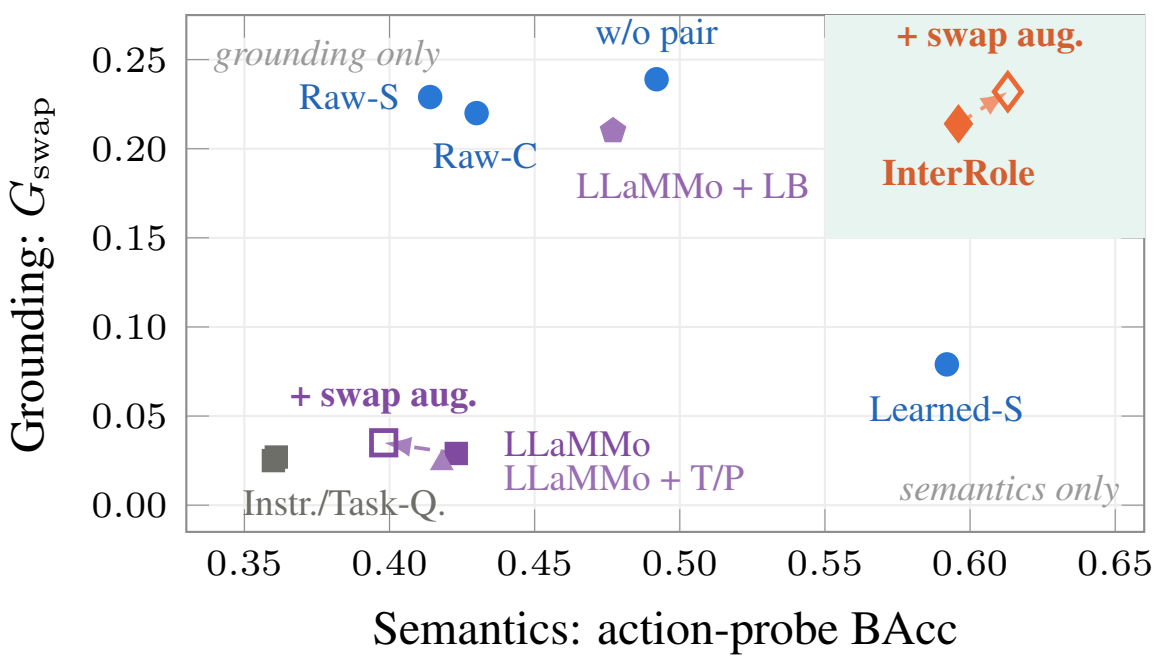
\begin{figure}[t]
\centering
\definecolor{pllammo}{RGB}{128,74,160}
\begin{tikzpicture}
\begin{axis}[
  width=0.95\linewidth, height=5.0cm, scale only axis=false,
  xlabel={Semantics: action-probe BAcc},
  ylabel={Grounding: $G_{\mathrm{swap}}$},
  xmin=0.33, xmax=0.66, ymin=-0.015, ymax=0.275,
  xtick={0.35,0.40,0.45,0.50,0.55,0.60,0.65},
  ytick={0,0.05,0.10,0.15,0.20,0.25},
  xticklabel={\pgfmathprintnumber[fixed,fixed zerofill,precision=2]{\tick}},
  yticklabel={\pgfmathprintnumber[fixed,fixed zerofill,precision=2]{\tick}},
  tick label style={font=\scriptsize}, label style={font=\small},
  axis line style={black!45}, tick style={black!45},
  ymajorgrids, xmajorgrids, grid style={black!7},
  clip=false,
]
\fill[prel!10] (axis cs:0.55,0.15) rectangle (axis cs:0.66,0.275);
\node[font=\scriptsize\itshape, text=black!40, anchor=north west] at (axis cs:0.333,0.272) {grounding only};
\node[font=\scriptsize\itshape, text=black!40, anchor=south east] at (axis cs:0.658,-0.012) {semantics only};
\addplot[only marks, mark=square*, mark size=2pt, pgray]
  coordinates {(0.361,0.027) (0.360,0.025)};
\node[font=\scriptsize, pgray, anchor=north west] at (axis cs:0.343,0.019) {Instr./Task-Q.};
\addplot[only marks, mark=square*, mark size=2.0pt, pllammo] coordinates {(0.423,0.029)};
\addplot[only marks, mark=triangle*, mark size=2.2pt, pllammo!70] coordinates {(0.418,0.0235)};
\node[font=\scriptsize, pllammo, anchor=west] at (axis cs:0.433,0.0345) {LLaMMo};
\node[font=\scriptsize, pllammo!80, anchor=west] at (axis cs:0.433,0.0155) {LLaMMo + T/P};
\addplot[only marks, mark=pentagon*, mark size=2.4pt, pllammo!75] coordinates {(0.477,0.210)};
\node[font=\scriptsize, pllammo!85, anchor=north west] at (axis cs:0.458,0.197) {LLaMMo + LB};
\addplot[only marks, mark=*, mark size=2.1pt, pone]
  coordinates {(0.414,0.229) (0.430,0.220) (0.592,0.079) (0.492,0.239)};
\node[font=\scriptsize, pone!85!black, anchor=east] at (axis cs:0.410,0.229) {Raw-S};
\node[font=\scriptsize, pone!85!black, anchor=north] at (axis cs:0.433,0.214) {Raw-C};
\node[font=\scriptsize, pone!85!black, anchor=north] at (axis cs:0.592,0.072) {Learned-S};
\node[font=\scriptsize, pone!85!black, anchor=south] at (axis cs:0.492,0.244) {w/o pair};
\addplot[only marks, mark=diamond*, mark size=3.4pt, ptwo]
  coordinates {(0.596,0.214)};
\node[font=\scriptsize\bfseries, ptwo!90!black, anchor=north] at (axis cs:0.596,0.204) {InterRole};
\draw[pllammo!70, densely dashed, line width=0.7pt, -{Latex[length=1.8mm]}]
  (axis cs:0.423,0.029) -- (axis cs:0.398,0.035);
\addplot[only marks, mark=square, mark size=2.3pt, pllammo, mark options={solid, line width=0.9pt, fill=white}]
  coordinates {(0.398,0.035)};
\node[font=\scriptsize\bfseries, pllammo, anchor=south] at (axis cs:0.398,0.041) {+ swap aug.};
\draw[ptwo!70, densely dashed, line width=0.7pt, -{Latex[length=1.8mm]}]
  (axis cs:0.596,0.214) -- (axis cs:0.613,0.232);
\addplot[only marks, mark=diamond, mark size=3.4pt, ptwo, mark options={solid, line width=0.9pt, fill=white}]
  coordinates {(0.613,0.232)};
\node[font=\scriptsize\bfseries, ptwo!90!black, anchor=south] at (axis cs:0.607,0.243) {+ swap aug.};
\end{axis}
\end{tikzpicture}
\caption{
Semantics versus grounding (Qwen2.5-1.5B).
Baselines (grey, LLaMMo purple; triangle: + T/P; pentagon: + LB, late binding without prefix pooling, with InterRole's 52-token prefix) have neither or only grounding; InterRole variants (blue) lack semantics without learned evidence (Raw-S/C) or the pair stream, or lose grounding without contextual reading (Learned-S); only the default InterRole reaches both (shaded).
Dashed arrows: swap augmentation (hollow markers).
}
\label{fig:pareto}
\vspace{-8pt}
\end{figure}

\noindent\textbf{Why it works.}
\label{sec:factorial_ablation}
We ablate InterRole on Qwen2.5-1.5B (Table~\ref{tab:factorial_ablation},
Fig.~\ref{fig:pareto}): \emph{Raw} evidence uses the independently encoded
streams $(\Sone,\Stwo)$ without partner conditioning or pair stream, the
\emph{Static} reader sets $\gamma_\view\equiv0$ in
Eq.~\eqref{eq:calibration}, and one variant removes the pair stream.
Two controls instantiate prior mechanisms on LLaMMo: \emph{LLaMMo + T/P} adds
InterRole's target/partner embeddings (person markers, as in
VIM~\cite{vim}) to its two person streams before pooling, and
\emph{LLaMMo + LB} applies late binding by feeding its participant and global
tokens to the language model without pooling, at InterRole's prefix length
and a similar parameter count.
The baselines are nearly, though not strictly, invariant: a swap moves their
frozen representation by only 0.15--0.16 of the median inter-sample distance
(cosine .97), consistent with their slot-bound role probes
(Table~\ref{tab:grounding_diagnosis}).
Swap gain separates the two descriptions without judging them, so we read
it together with Swap-Tgt, which scores the swapped descriptions against the
exchanged references.

\noindent\textbf{Fusing before binding is the problem.}
Every InterRole variant follows the relation direction in 78.6--88.1\% of
asymmetric interactions, whereas LLaMMo, which tags each person but fuses
before pooling, stays at 26.5\%.
Person markers do not help (+ T/P: $G_{\mathrm{swap}}$ .024, follow 29.2\%):
tags do not protect the correspondence once the streams are fused.
Late binding does: applied to LLaMMo, it lifts $G_{\mathrm{swap}}$ from .029
to .210 and the follow rate to 84.1\%, so the gain is not due to bandwidth or
capacity; InterRole's learned evidence and contextual reading add semantics
on top (action BAcc .596 vs.\ .477, CIDEr +.11--.14).

\noindent\textbf{Grounding and semantics are obtained together.}
Raw--Static has a high swap gain (.229) but the weakest semantics of the
variants and the lowest Swap-Tgt.
Learned evidence brings semantics but erodes binding ($G_{\mathrm{swap}}$
.079), as partner conditioning makes the two streams more alike (pooled
cosine .458 vs.\ .526) and the decoder falls back on slot identity;
contextual reading restores it while keeping the semantics, and only the
combination reaches both.
The pair stream adds relational semantics: removing it lowers Relation CIDEr,
action BAcc, and Swap-Tgt (Table~\ref{tab:factorial_ablation}).

\begin{table}[t]
\centering
\small
\setlength{\tabcolsep}{3pt}
\renewcommand{\arraystretch}{1.1}
\caption{
Ablation on Qwen2.5-1.5B: evidence (Raw vs.\ Learned), reader (Static vs.\ Contextual), and pair stream.
All rows below the baseline are InterRole variants; Learned + Contextual is the default.
\emph{Grounding}: swap quality (Swap-Tgt), swap gain, directed-role Pair-C.
Higher is better; best per column in \textbf{bold}.
}
\label{tab:factorial_ablation}
\resizebox{\linewidth}{!}{%
\begin{tabular}{l|ccc|ccc}
\toprule
& \multicolumn{3}{c|}{Semantics} & \multicolumn{3}{c}{Grounding} \\
Variant & Person & Rel. & Act. BAcc & Swap-Tgt & $G_{\mathrm{swap}}$ & Role Pair-C \\
\midrule
\color{gray}LLaMMo (baseline)
& \color{gray}0.550 & \color{gray}0.538 & \color{gray}0.423 & \color{gray}0.544 & \color{gray}0.029 & \color{gray}0.118 \\
\midrule
\multicolumn{7}{l}{\emph{InterRole variants}} \\
Raw + Static           & 0.569 & 0.553 & 0.414 & 0.564 & 0.229 & 0.530 \\
Raw + Contextual       & 0.581 & 0.567 & 0.430 & 0.574 & 0.220 & 0.572 \\
Learned + Static       & 0.586 & 0.590 & 0.592 & 0.578 & 0.079 & 0.561 \\
\rowcolor{qours}
Learned + Contextual (default)
& \textbf{0.601} & \textbf{0.595} & \textbf{0.596} & \textbf{0.594} & 0.214 & 0.569 \\
Learned + Contextual, w/o pair
& 0.579 & 0.571 & 0.492 & 0.573 & \textbf{0.239} & \textbf{0.619} \\
\bottomrule
\end{tabular}%
}
\vspace{-6pt}
\end{table}

\noindent\textbf{Supervising the swap does not substitute for structure.}\label{sec:swap_aug}
We also train LLaMMo and InterRole with participant-swap augmentation, which
exchanges the two streams together with the participant-centred references,
following the participant-order shuffling of SocialGen~\cite{socialgen}
(dashed arrows in Fig.~\ref{fig:pareto}; 20 epochs, as it enlarges the
training set, and 10 epochs gives the same picture, Supp.~Mat.~\ref{supp:role_marker}).
For LLaMMo, $G_{\mathrm{swap}}$ moves only from .029 to .035, one sixth of
InterRole \emph{without} augmentation; for InterRole it is modestly
complementary (.232 vs.\ .214), and we keep it unaugmented to isolate its
inductive bias.

\input{figures/fig_retrieval_main}

\subsection{Cross-Dataset Transfer}
\label{sec:cross_dataset}

Swap diagnostics need no references, so they can test grounding where
participant-level annotations do not exist.
We train on Inter-X only, with a 20-joint representation shared by both
datasets, and apply the models zero-shot to the 1{,}176 InterHuman test
sequences (Table~\ref{tab:cross_dataset}; representation details in
Supp.~Mat.~\ref{supp:cross}).
Only Overall has references, so participant and relation scores are
reference-free consistency diagnostics.
Both the failure and the remedy transfer: the baseline gives both people
the same description for 74.6\% of sequences (InterRole 27.9\%), its swap
gain stays at .034 (InterRole .081), and only InterRole's relation
descriptions move toward the relabelled originals; InterRole's Overall
captions also match the references at least as well (.301 vs.\ .291).

\begin{table}[tb]
\centering
\small
\setlength{\tabcolsep}{3.0pt}
\caption{Zero-shot transfer from Inter-X to InterHuman (1{,}176 test sequences); swap metrics as in Table~\ref{tab:grounding_diagnosis}.
Identical: fraction of sequences whose two person descriptions coincide; Rel.\ dir.\ gain: reference-free relation-reversal gain (Sec.~\ref{sec:grounding_diagnosis}); Overall BERT: BERTScore-F1 of the generated Overall description against the InterHuman reference captions.
}
\label{tab:cross_dataset}
\resizebox{\linewidth}{!}{%
\begin{tabular}{l|cccc|c|c}
\toprule
Method & $S_{\mathrm{corr}}\!\uparrow$ & $S_{\mathrm{wrong}}\!\downarrow$
& $G_{\mathrm{swap}}\!\uparrow$ & Identical$\downarrow$ & Rel.\ dir.\ gain$\uparrow$
& Overall BERT$\uparrow$ \\
\midrule
LLaMMo & 0.582 & 0.549 & 0.034 & 74.6\% & $-0.010$ & 0.291 \\
InterRole    & \textbf{0.613} & \textbf{0.532} & \textbf{0.081} & \textbf{27.9\%} & $\mathbf{+0.020}$ & \textbf{0.301} \\
\bottomrule
\end{tabular}%
}
\vspace{-4pt}
\end{table}

\subsection{Application: Retrieval and Execution}
\label{sec:retrieval}

A grounded representation should support retrieval by who does what.
We freeze each model and train only two linear projections (motion prefix
and reference text) with a symmetric InfoNCE loss and no swapped motion
(details in Supp.~Mat.~\ref{supp:retrieval}).
On asymmetric interactions, we rank each $\Pone$, $\Ptwo$ or $\Rel$ query
against 16 interactions and their swapped copies (same actions, swapped
performers; swap-hard R@1), and test preference for its own motion over
its swapped copy (2AFC, chance 50\%).

\begin{table}[t]
\centering
\small
\setlength{\tabcolsep}{3.2pt}
\caption{
Who-did-what retrieval (Qwen2.5-1.5B, asymmetric interactions).
Swap-hard R@1 (\%): 32 candidates, i.e.\ 16 interactions plus their swapped copies (chance 3.1); 2AFC (\%): preference for the original over its swapped copy (chance 50).
}
\label{tab:retrieval}
\scalebox{0.90}{%
\begin{tabular}{l|ccc|ccc}
\toprule
& \multicolumn{3}{c|}{Swap-hard R@1} & \multicolumn{3}{c}{2AFC} \\
Method & $\Pone$ & $\Ptwo$ & $\Rel$ & $\Pone$ & $\Ptwo$ & $\Rel$ \\
\midrule
Instruction Cond. & 24.8 & 27.3 & 28.3 & 56.8 & 60.9 & 59.7 \\
Task-Query Cond.  & 22.1 & 23.7 & 23.2 & 56.8 & 56.6 & 55.9 \\
LLaMMo            & 27.1 & 27.9 & 30.5 & 59.7 & 58.3 & 60.2 \\
\textbf{InterRole} & \textbf{39.7} & \textbf{36.0} & \textbf{48.4}
& \textbf{68.1} & \textbf{63.7} & \textbf{75.9} \\
\bottomrule
\end{tabular}}
\vspace{-6pt}
\end{table}

InterRole's swap-hard R@1 is 1.3--1.6$\times$ that of the strongest
baseline (Table~\ref{tab:retrieval}).
The 2AFC shows why: the baselines stay near chance (55.9--60.9\%),
retrieving the right action but not the right performer, whereas
InterRole picks the correct orientation for 75.9\% of relation and
68.1\% of $\Pone$ queries; on $\Ptwo$ its 2AFC lead is small.
Without swapped distractors, InterRole also retrieves best on all four
views under the protocols of~\cite{tmr} (Supp.~Mat.~\ref{supp:retrieval}).
Retrieval tests the same property from the retrieval side: the query names who does what, and only a representation that keeps the participants distinct can rank the right performer above the swapped copy.
As a further use, when the retrieved motion is executed by two simulated humanoid robots driven by a multi-agent (MAPPO) tracking policy~\cite{chen2026rhythm} (Fig.~\ref{fig:retrieval_main}), the baseline's Top-1 makes the wrong robot act (a push with the roles reversed), whereas InterRole's makes Person~1's robot act on Person~2's, as the query asks: retrieval decides not only what the robots do but who does it.


%% file: figures/fig_qualitative.tex
\usetikzlibrary{calc}

\input{figures/qual_base_macros}
\tikzset{qtag/.style={anchor=south, inner sep=1.2pt, rounded corners=1pt,
  font=\bfseries\fontsize{5.2}{5.2}\selectfont, text=white, draw=white, line width=0.25pt}}
\newcommand{\qimg}[5]{\makebox[1.06in]{%
  \begin{tikzpicture}[baseline=(im.center)]
    \node[inner sep=0pt] (im) {\includegraphics[width=1.06in,height=0.98in,keepaspectratio]{#1}};
    \node[qtag, fill=pone] at ($(im.north west)+#2*($(im.north east)-(im.north west)$)+#3*($(im.south west)-(im.north west)$)$) {P1};
    \node[qtag, fill=ptwo] at ($(im.north west)+#4*($(im.north east)-(im.north west)$)+#5*($(im.south west)-(im.north west)$)$) {P2};
  \end{tikzpicture}}}

\newcommand{\qualcase}[4]{%
  \noindent
  \begin{minipage}[c]{0.13in}\centering
    \rotatebox[origin=c]{90}{\textbf{\small #1}}
  \end{minipage}\hfill
  \begin{minipage}[c]{2.16in}\centering
    #2\hspace{2pt}#3
  \end{minipage}\hfill
  \begin{minipage}[c]{4.48in}
    \renewcommand{\arraystretch}{1.0}\setlength{\tabcolsep}{2pt}
    \begin{tabular}{@{}p{0.40in}p{0.15in}R{1.88in}R{1.88in}@{}}
      #4
    \end{tabular}
  \end{minipage}\par
}
\newcommand{\qrule}{\par\vspace{1pt}{\color{black!22}\hrule height 0.4pt}\vspace{2pt}}

\begin{figure*}[!t]
\centering
\fontsize{6.6}{7.7}\selectfont
\noindent
\begin{minipage}[c]{0.13in}\ \end{minipage}\hfill
\begin{minipage}[c]{2.16in}\centering
  \makebox[1.06in]{\textbf{(a)}~Original}\hspace{2pt}\makebox[1.06in]{\textbf{(b)}~Swapped}
\end{minipage}\hfill
\begin{minipage}[c]{4.48in}
  \setlength{\tabcolsep}{2pt}
  \begin{tabular}{@{}p{0.40in}p{0.15in}R{1.88in}R{1.88in}@{}}
    & & \textbf{(a)}~Original input & \textbf{(b)}~Participants swapped \\
  \end{tabular}
\end{minipage}\par
\qrule

\qualcase{Bend}{\qimg{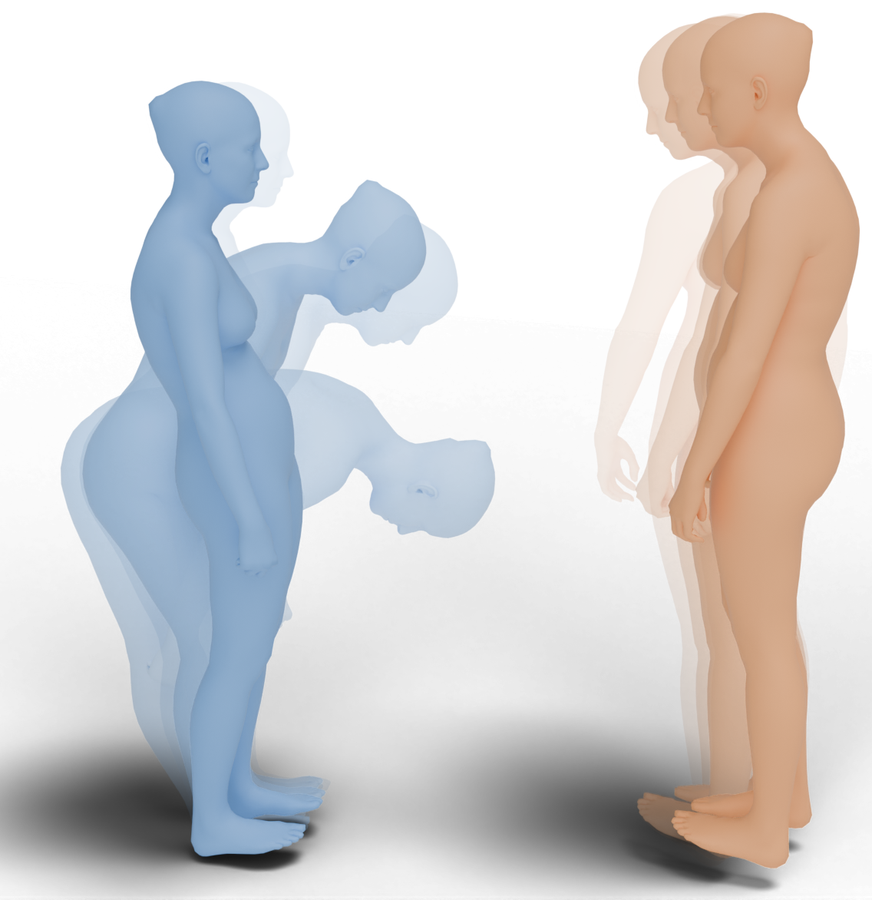}{0.230}{0.187}{0.876}{0.112}}{\qimg{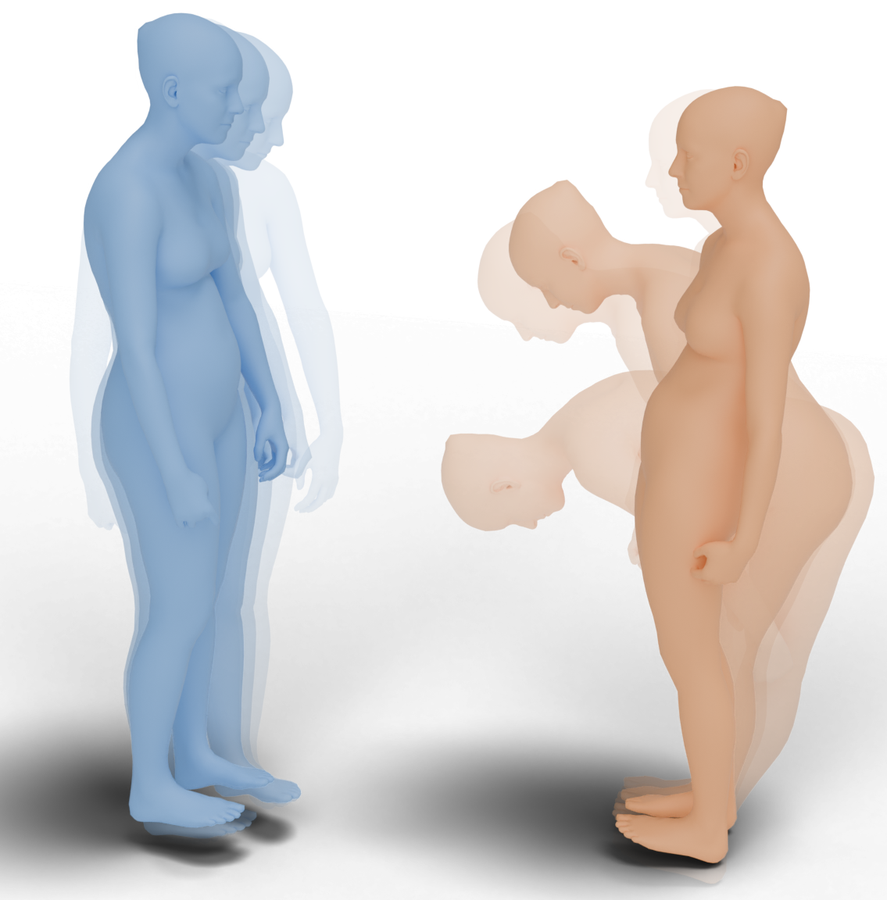}{0.204}{0.112}{0.829}{0.193}}{%
  \qm{Ref.}      & \qvR & \qA\ bows to \qB\ while facing each other. & \qB\ bows to \qA\ while facing each other. \\[1.5pt]
  \qm{LLaMMo}    & \qvA & \qg{\qel\ bows to \qB}.\qyes & \qb{\qel\ bows to \qB}.\qno \\
                 & \qvB & \qb{\qel\ bows to \qA}.\qno & \qg{\qel\ bows to \qA}.\qyes \\
                 & \qvR & \qb{\qB\ bows to \qA} \qel\qno & \qg{\qB\ bows to \qA} \qel\qyes \\[1.5pt]
  \rowcolor{qours}
  \qm{InterRole} & \qvA & \qel\ \qg{bends over to bow}.\qyes & \qel\ \qg{is bowed to by \qB}.\qyes \\
  \rowcolor{qours}
                 & \qvB & \qel\ \qg{is bowed to by \qA}.\qyes & \qel\ \qg{bends over to bow}.\qyes \\
  \rowcolor{qours}
                 & \qvR & \qg{\qA\ bows to \qB}, and \qB\ nods in acknowledgment.\qyes & \qg{\qB\ bows to \qA}, and \qA\ nods in acknowledgment.\qyes \\
}
\qrule

\qualcase{Massage}{\qimg{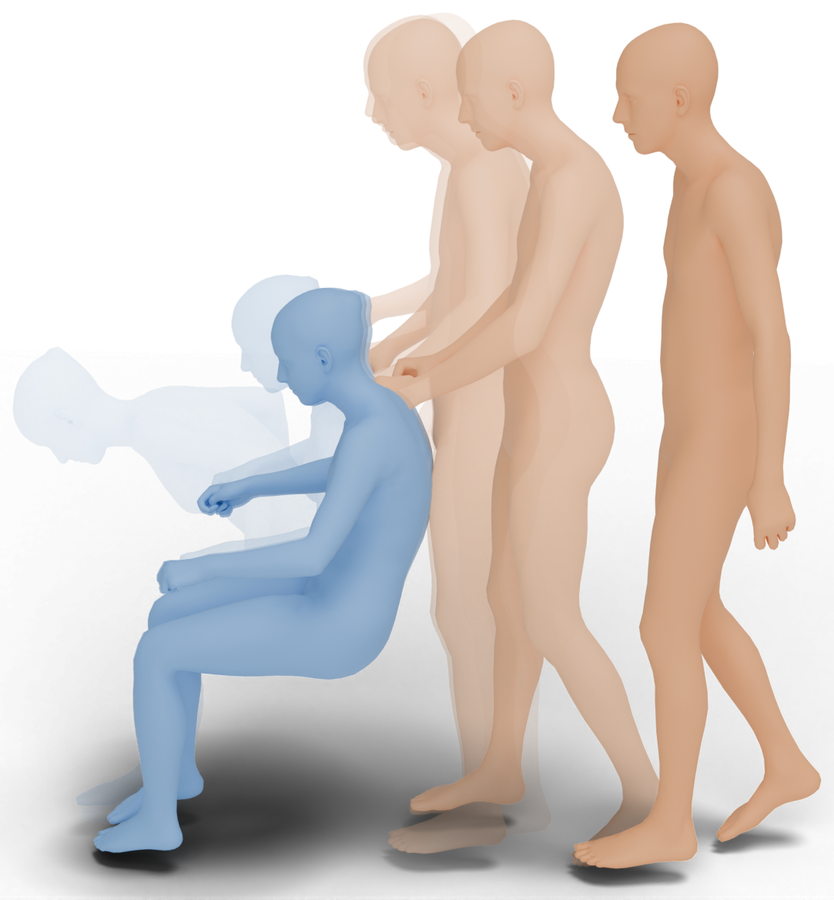}{0.393}{0.417}{0.810}{0.123}}{\qimg{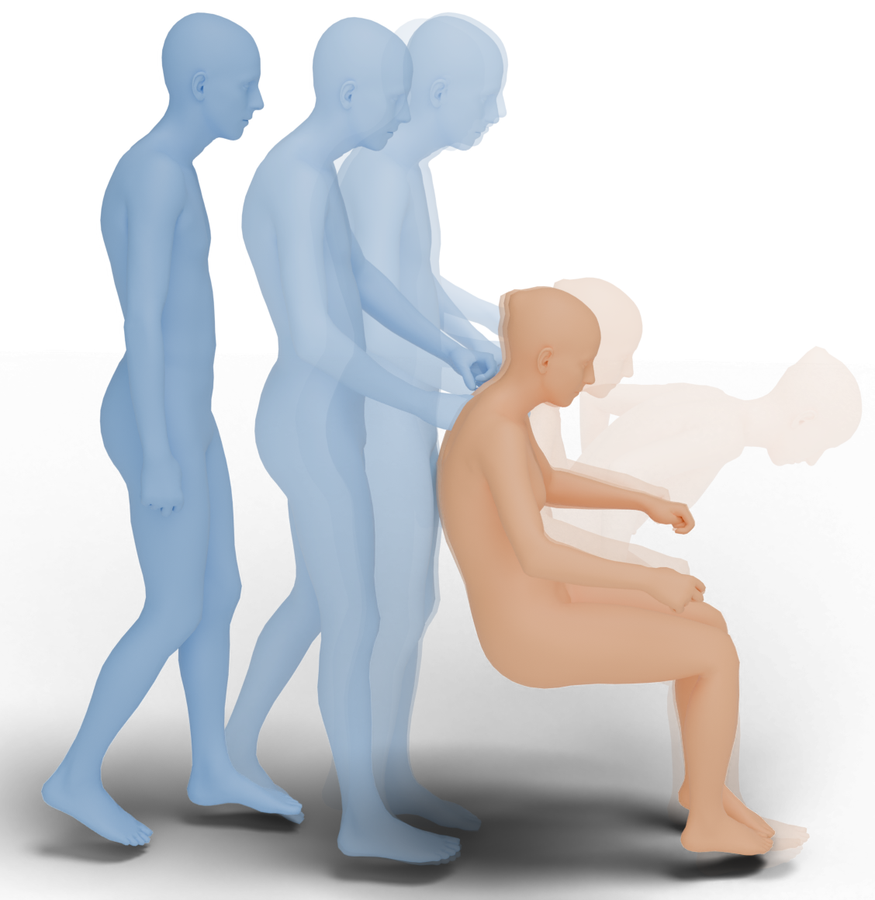}{0.232}{0.112}{0.620}{0.417}}{%
  \qm{Ref.}      & \qvR & \qB\ stands behind \qA, places his/her hands \qel & \qA\ stands behind \qB, places his/her hands \qel \\[1.5pt]
  \qm{LLaMMo}    & \qvA & \qg{\qel\ sits on a chair}.\qyes & \qb{\qel\ sits on a chair}.\qno \\
                 & \qvB & \qb{\qel\ sits on a chair}.\qno & \qg{\qel\ sits on a chair}.\qyes \\
                 & \qvR & \qb{\qB\ sits on \qA's lap} \qel\qno & \qb{\qB\ sits on \qA's lap} \qel\qno \\[1.5pt]
  \rowcolor{qours}
  \qm{InterRole} & \qvA & \qel\ \qg{sits on a chair}.\qyes & \qel\ \qg{stands behind \qB} \qel\qyes \\
  \rowcolor{qours}
                 & \qvB & \qel\ \qg{stands behind \qA} and \qg{massages} \qel\qyes & \qel\ \qg{sits on a chair}.\qyes \\
  \rowcolor{qours}
                 & \qvR & \qg{\qB\ walks behind \qA} and \qg{massages} \qA's shoulders \qel\qyes & \qg{\qA\ stands behind \qB} and \qg{massages} \qB's shoulders \qel\qyes \\
}

\caption{
Qualitative results on Inter-X (Qwen2.5-1.5B).
\emph{Left:} slot~1 (\textcolor{pone}{blue}) and slot~2 (\textcolor{ptwo}{orange}); fainter silhouettes are earlier frames; swapped inputs are viewed from the opposite side.
\emph{Right:} reference ($\Rel$ only), LLaMMo and InterRole outputs for the original (\textbf{a}) and swapped (\textbf{b}) input, verbatim except elisions (\qel).
\textcolor{qokc}{\ding{51}}/\textcolor{qbadc}{\ding{55}}: action (for $\Rel$, its direction) attributed to the correct person.
}
\label{fig:qualitative}
\end{figure*}

%% file: figures/fig_retrieval_main.tex
\input{figures/retrieval_robot_macros}
\begin{figure*}[!t]
\centering
\fontsize{6.8}{8.0}\selectfont
\noindent\begin{minipage}[c]{0.16in}\ \end{minipage}\hfill\begin{minipage}[c]{1.22in}\textbf{Text query}\end{minipage}\hfill\makebox[2.64in][l]{\textbf{InterRole (ours): Top-1 motion}}\hfill\makebox[2.52in][l]{\textbf{LLaMMo: Top-1 motion}}\par\rrule
\rcaseR{Push}{\qA{} pushes \qB{}'s left arm, causing \qB{} to take a few steps to the right.}{{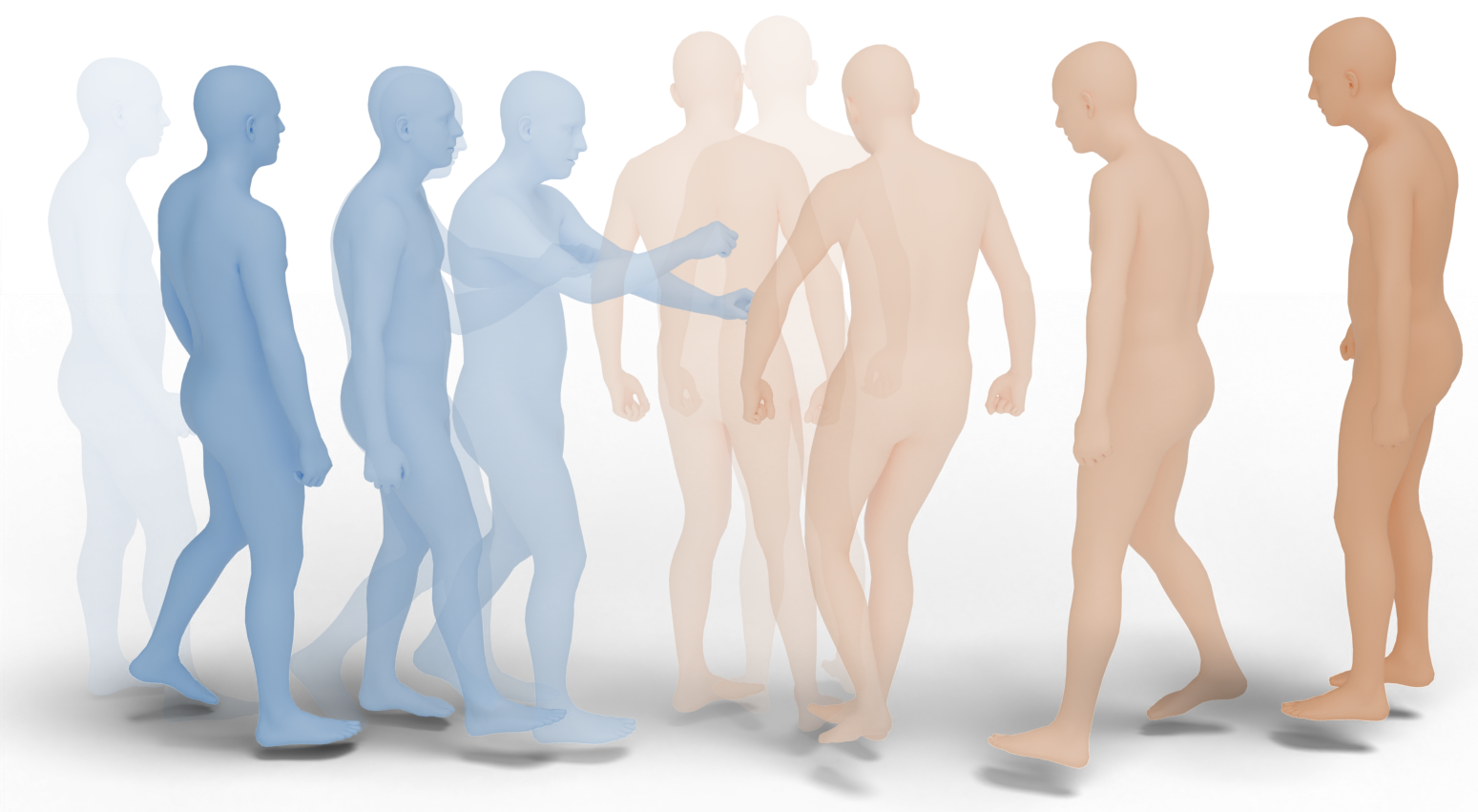}{0.158}{0.121}{0.919}{0.061}}{%
{\color{qlgray}\scriptsize Push}\\[1pt]``\qA{} pushes \qB{}'s hand from the side, causing \qB{} to be pushed to the side.''\\[2pt]\ryes{correct roles}}{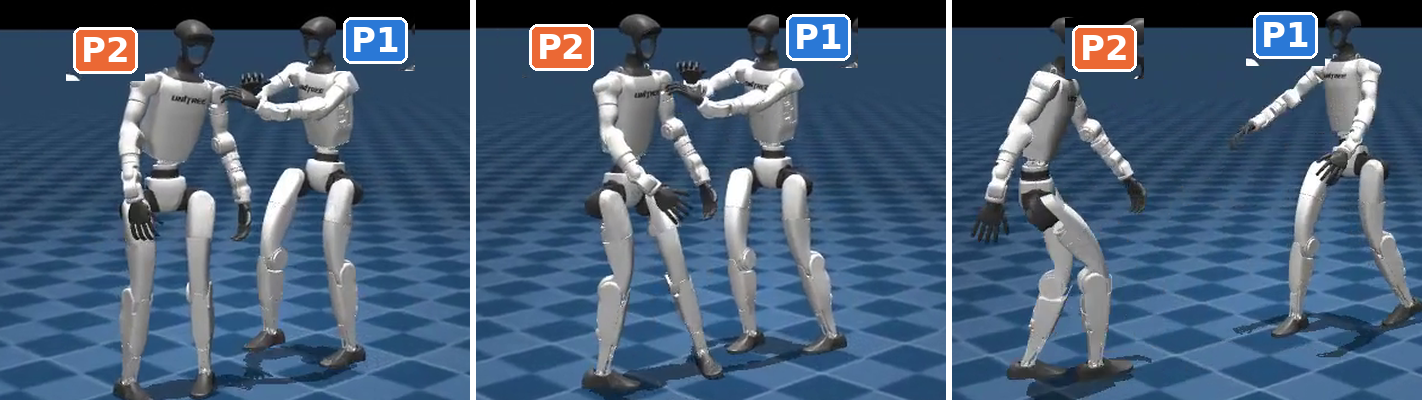}{{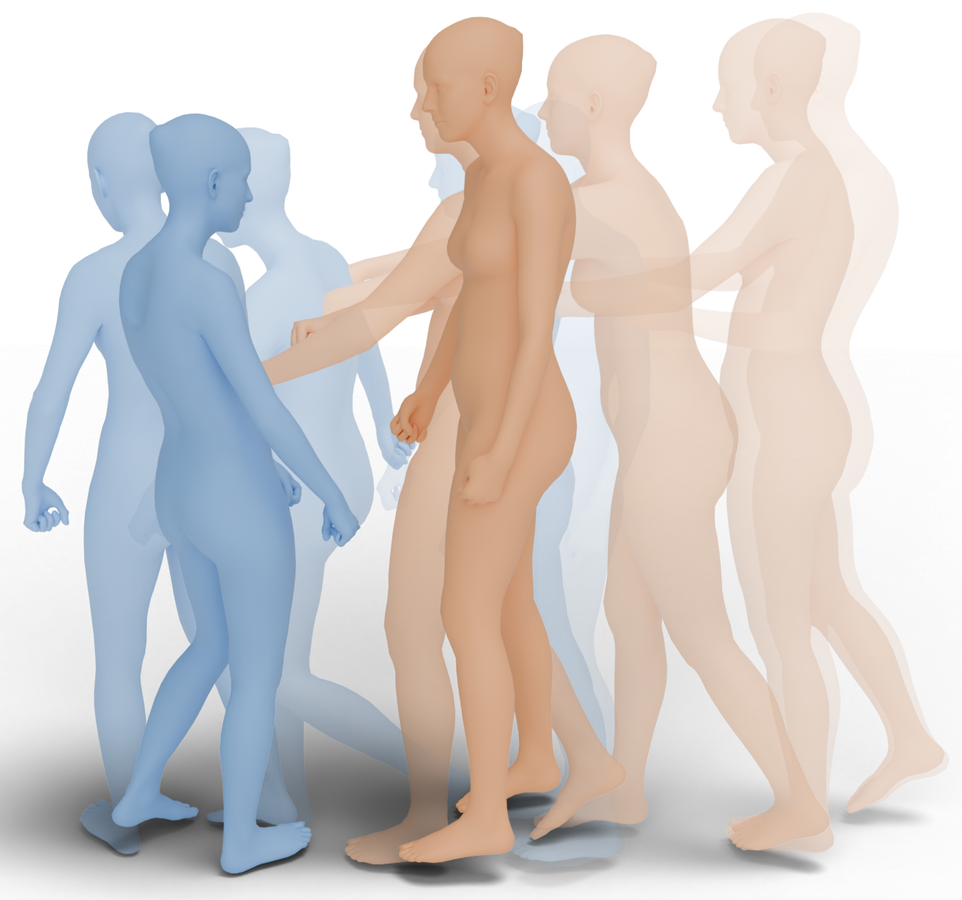}{0.203}{0.166}{0.498}{0.059}}{%
{\color{qlgray}\scriptsize Push}\\[1pt]``\qB{} pushes \qA{} towards \qA{}'s back, and \qA{} walks forward.''\\[2pt]\rno{roles reversed}}{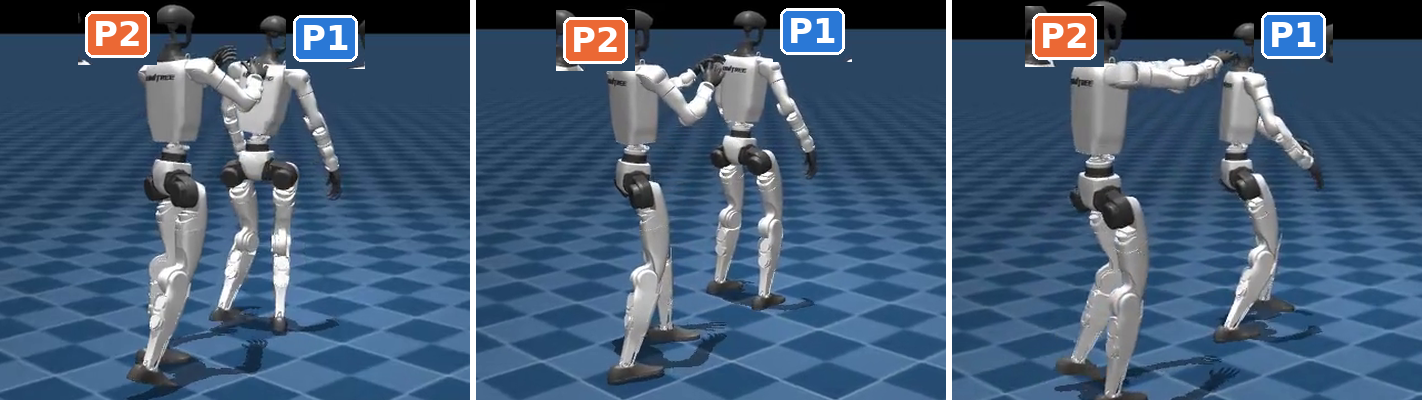}
\caption{Top-1 text-to-motion retrieval on Inter-X for a Relation query (gallery of originals and swapped copies; \textcolor{pone}{blue} = \qA{}, \textcolor{ptwo}{orange} = \qB{}). Each cell shows the retrieved motion with its action label and annotation, and, below it, its execution by two simulated humanoid robots driven by a multi-agent (MAPPO) tracking policy. InterRole's Top-1 matches the query, whereas LLaMMo's shows the roles reversed. More examples in Supp.~Mat.~\ref{supp:qualitative}.}
\label{fig:retrieval_main}
\vspace{-8pt}
\end{figure*}

%% file: figures/retrieval_robot_macros.tex
\providecolor{pone}{RGB}{42,120,214}
\providecolor{ptwo}{RGB}{214,120,60}
\providecolor{qokc}{RGB}{26,127,75}
\providecolor{qbadc}{RGB}{203,36,49}
\providecolor{qours}{RGB}{232,240,252}
\providecolor{qlgray}{RGB}{110,110,110}
\providecommand{\qA}{\textcolor{pone}{\mbox{Person\,1}}}
\providecommand{\qB}{\textcolor{ptwo}{\mbox{Person\,2}}}
\providecommand{\ryes}[1]{\textcolor{qokc}{\ding{51}~#1}}
\providecommand{\rno}[1]{\textcolor{qbadc}{\ding{55}~#1}}
\providecommand{\rimg}[1]{\includegraphics[width=1.42in,height=1.00in,keepaspectratio]{#1}}
\providecommand{\rcase}[6]{%
  \noindent
  \begin{minipage}[c]{0.16in}\centering\rotatebox[origin=c]{90}{\textbf{\small #1}}\end{minipage}\hfill
  \begin{minipage}[c]{1.22in}\raggedright #2\end{minipage}\hfill
  \colorbox{qours}{\begin{minipage}[c]{2.52in}
    \begin{minipage}[c]{1.44in}\centering\rimg{#3}\end{minipage}\hfill\begin{minipage}[c]{1.04in}\raggedright #4\end{minipage}\end{minipage}}\hfill
  \begin{minipage}[c]{2.52in}
    \begin{minipage}[c]{1.44in}\centering\rimg{#5}\end{minipage}\hfill\begin{minipage}[c]{1.04in}\raggedright #6\end{minipage}\end{minipage}\par
}
\providecommand{\rrule}{\par\vspace{2pt}{\color{black!22}\hrule height 0.4pt}\vspace{3pt}}
\providecommand{\rrobotdir}{retrieval/robot}
\providecommand{\rrobot}[1]{\IfFileExists{#1}%
  {\includegraphics[width=2.48in,height=0.74in,keepaspectratio]{#1}}%
  {\fcolorbox{black!25}{black!4}{\parbox[c][0.72in][c]{2.40in}{\centering\scriptsize\color{black!50}two-robot execution (to be added)}}}}
\tikzset{rtag/.style={anchor=south, inner sep=1.2pt, rounded corners=1pt,
  font=\bfseries\fontsize{5.2}{5.2}\selectfont, text=white, draw=white, line width=0.25pt}}
\providecommand{\rimgm}[5]{%
  \begin{tikzpicture}[baseline=(im.center)]
    \node[inner sep=0pt] (im) {\includegraphics[width=1.30in,height=0.74in,keepaspectratio]{#1}};
    \node[rtag, fill=pone] at ($(im.north west)+#2*($(im.north east)-(im.north west)$)+#3*($(im.south west)-(im.north west)$)$) {P1};
    \node[rtag, fill=ptwo] at ($(im.north west)+#4*($(im.north east)-(im.north west)$)+#5*($(im.south west)-(im.north west)$)$) {P2};
  \end{tikzpicture}}
\providecommand{\rcaseR}[8]{%
  \noindent
  \begin{minipage}[c]{0.16in}\centering\rotatebox[origin=c]{90}{\textbf{\small #1}}\end{minipage}\hfill
  \begin{minipage}[c]{1.22in}\raggedright #2\end{minipage}\hfill
  \colorbox{qours}{\begin{minipage}[c]{2.52in}\centering
    \parbox[c][0.78in][c]{2.52in}{\begin{minipage}[c]{1.30in}\centering\rimgm#3\end{minipage}\hfill\begin{minipage}[c]{1.18in}\raggedright #4\end{minipage}}\par\vspace{2pt}
    \rrobot{#5}\end{minipage}}\hfill
  \colorbox{white}{\begin{minipage}[c]{2.52in}\centering
    \parbox[c][0.78in][c]{2.52in}{\begin{minipage}[c]{1.30in}\centering\rimgm#6\end{minipage}\hfill\begin{minipage}[c]{1.18in}\raggedright #7\end{minipage}}\par\vspace{2pt}
    \rrobot{#8}\end{minipage}}\par
}

%% file: sec/5_conclusion.tex

\section{Conclusion}
\label{sec:conclusion}

This paper studies participant grounding in two-person motion understanding and
shows that the existing motion--language models describe input slots rather
than people, a failure that caption metrics miss but participant-swap
diagnostics expose.
We attributed it to fusing the participants before the requested view is
known and proposed InterRole, which keeps participant evidence separate
and binds it to each view only at the end.
InterRole makes participant and relation descriptions follow the people,
improves generation on all views, and consistently improves over the
baseline zero-shot on an unseen dataset.
The principle is not tied to our architecture: applied to LLaMMo alone, late binding lifts $G_{\mathrm{swap}}$ from .029 to .210.
We hope these diagnostics and the late-binding principle will help future
interaction models know not only what happens, but who does what.
Limitations are discussed in Supp.~Mat.~\ref{supp:limitations}.

\clearpage
\section*{Acknowledgments}
We thank Prof. Changliu Liu and the Intelligent Control Lab at Carnegie Mellon University for their support and access to computational resources.
Lin Wu acknowledges mobility funding from the University of Glasgow Graduate School that supported his research visit to the Robotics Institute at Carnegie Mellon University.

\noindent\textbf{LLM Usage Statement.}
LLMs are pipeline components (Qwen2.5 and Gemma3 decoders; Qwen3-8B for view
annotation, Supp.~Mat.~\ref{supp:data}).
An LLM assistant also helped with editing and with figure and analysis scripts;
the authors checked all content and are responsible for it.

%% file: sec/X_suppl.tex

\ifdefined\suppstandalone\else\clearpage\fi
\appendix
\renewcommand{\thesection}{S\arabic{section}}
\renewcommand{\thefigure}{S\arabic{figure}}\setcounter{figure}{0}
\renewcommand{\thetable}{S\arabic{table}}\setcounter{table}{0}
\renewcommand{\theequation}{S\arabic{equation}}\setcounter{equation}{0}
\ifdefined\suppstandalone\else\section*{Supplementary Material}\fi

\newcommand{\suppentry}[2]{\noindent\hyperref[#1]{\textbf{\ref*{#1}}\hspace{0.8em}\nameref*{#1}}%
  {\small\ \textcolor{black!55}{#2}}\dotfill\pageref*{#1}\par\vspace{1pt}}
\begin{center}\textbf{Contents}\end{center}
\vspace{-4pt}
\suppentry{supp:proofs}{Props.~\ref{prop:fusion} and~\ref{prop:equivariance} of the main paper}
\suppentry{supp:data}{canonicalization, swap protocol, probes}
\suppentry{supp:impl}{architecture, optimization}
\suppentry{supp:role_marker}{LLaMMo + T/P, + LB, swap augmentation, slot prior, reordering}
\suppentry{supp:ablations}{gates, similarity, invariance}
\suppentry{supp:cross}{InterHuman, 20 joints}
\suppentry{supp:retrieval}{probe, who-did-what}
\suppentry{supp:limitations}{}
\suppentry{supp:qualitative}{Inter-X, InterHuman, retrieval}
\bigskip

\section{Proofs}
\label{supp:proofs}

\paragraph{Proof of Prop.~\ref{prop:fusion} (main paper).}
$F_\theta(\Sw\mathcal{M},\Pone)=D(H(\bm{M}_2,\bm{M}_1),\Pone)
=D(H(\bm{M}_1,\bm{M}_2),\Pone)=F_\theta(\mathcal{M},\Pone)$, whereas
Eq.~\eqref{eq:consistency} of the main paper requires $F_\theta(\Sw\mathcal{M},\Pone)$ to
match $\Tpi F_\theta(\mathcal{M},\Ptwo)$.
The two coincide only if $F_\theta(\mathcal{M},\Pone)=\Tpi
F_\theta(\mathcal{M},\Ptwo)$, i.e.\ only if both participants receive the
same description up to relabelling.
\hfill$\square$

\paragraph{Proof of Prop.~\ref{prop:equivariance}(a) (main paper).}
Write one layer of $f_\theta$ as the map
$(\bm{a},\bm{b})\mapsto(\phi(\bm{a},\bm{b}),\phi(\bm{b},\bm{a}))$, where
$\phi$ is Eq.~\eqref{eq:gate_main} of the main paper with the layer's
parameters, applied to the target $\bm{a}$ and partner $\bm{b}$.
Shared parameters mean that the same $\phi$ is used for both
directions; synchronous updates mean that both calls read the
pre-update pair $(\bm{a},\bm{b})$.
Let $(\bm{a}_\ell,\bm{b}_\ell)$ be the states after $\ell$ layers for
input $(\Sone,\Stwo)$ and $(\bm{a}'_\ell,\bm{b}'_\ell)$ those for input
$(\Stwo,\Sone)$.
For $\ell=0$, $(\bm{a}'_0,\bm{b}'_0)=(\bm{b}_0,\bm{a}_0)$.
If $(\bm{a}'_\ell,\bm{b}'_\ell)=(\bm{b}_\ell,\bm{a}_\ell)$, then
$\bm{a}'_{\ell+1}=\phi(\bm{b}_\ell,\bm{a}_\ell)=\bm{b}_{\ell+1}$ and
$\bm{b}'_{\ell+1}=\phi(\bm{a}_\ell,\bm{b}_\ell)=\bm{a}_{\ell+1}$.
By induction the outputs are exchanged after every layer, so
$\Sw$ maps $(\Zone,\Ztwo)$ to $(\Ztwo,\Zone)$.
The argument holds for deterministic layers; we verify it in evaluation
mode (dropout disabled), where swapping the inputs exchanges the outputs
with bitwise-identical results (float32) for 1- and 3-layer learners.
\hfill$\square$

\paragraph{Proof of Prop.~\ref{prop:equivariance}(b) (main paper).}
Let primes denote quantities computed from $\Sw\mathcal{M}$.
(1)~The encoder is applied to each stream with the same weights, so
$\bm{S}'_1=\Stwo$ and $\bm{S}'_2=\Sone$.
(2)~By part~(a), $\bm{Z}'_1=\Ztwo$ and $\bm{Z}'_2=\Zone$.
(3)~$\bm{D}'_{1\to2}=\bm{J}_1-\bm{J}_2=\bm{D}_{2\to1}$, hence
$\bm{Z}'_{12}=\mathcal{P}_\theta([\Stwo;\Sone;\mathcal{G}_\theta(\bm{D}_{2\to1})])=\Zler$, the pair stream with swapped orientation.
(4)~The role embedding $\bm{\rho}_{\view,i}$ depends only on the view and
the slot index: for $\view=\mathsf{P}k$, slot $k$ receives
$\bm{r}_{\mathrm{tgt}}$ and slot $3{-}k$ receives
$\bm{r}_{\mathrm{ptn}}$; neutral views receive none.

Participant-centred views.
Let $\view=\mathsf{P}k$, so $\Sw\view=\mathsf{P}(3{-}k)$.
In $\bank_\view(\Sw\mathcal{M})$, slot $k$ carries
$\bm{Z}'_k+\bm{e}_k+\bm{r}_{\mathrm{tgt}}=\bm{Z}_{3-k}+\bm{e}_k+\bm{r}_{\mathrm{tgt}}$
and slot $3{-}k$ carries $\bm{Z}_k+\bm{e}_{3-k}+\bm{r}_{\mathrm{ptn}}$.
In $\bank_{\Sw\view}(\mathcal{M})$, the target slot is $3{-}k$, which
carries $\bm{Z}_{3-k}+\bm{e}_{3-k}+\bm{r}_{\mathrm{tgt}}$, and slot $k$
carries $\bm{Z}_k+\bm{e}_k+\bm{r}_{\mathrm{ptn}}$.
Both inputs therefore attach $\bm{r}_{\mathrm{tgt}}$ to $\bm{Z}_{3-k}$ and
$\bm{r}_{\mathrm{ptn}}$ to $\bm{Z}_k$; they differ only in the slot
embedding, boundary token and position indices of each participant
stream, which are attached to slots rather than persons (item~(i)), the view token $\bm{q}_{\mathsf{P}k}$ vs.\
$\bm{q}_{\mathsf{P}(3-k)}$ (item~(ii)), and $\Zler$ vs.\ $\Zrel$
(item~(iii)).

Neutral views.
For $\view\in\{\Ovr,\Rel\}$, $\Sw\view=\view$, the view tokens coincide
and no role embedding is present; the participant streams exchange
slots and the pair stream changes orientation, i.e.\ only items~(i) and
(iii) differ.

Non-invariance.
Whenever $\Zone\neq\Ztwo$ (the participants behave differently), slot~1
holds different evidence in $\bank_\view(\mathcal{M})$ and
$\bank_\view(\Sw\mathcal{M})$, so the bank is not swap-invariant in the
sense of Prop.~\ref{prop:fusion} of the main paper.
\hfill$\square$

\section{Data and Evaluation Protocol}
\label{supp:data}

\paragraph{Semantic-view canonicalization.}
An LLM-based pipeline (Qwen3-8B) decomposes each source caption into caption-local
A/B behaviours, a relation, and a likely initiator, then deterministically
maps the local roles to motion tracks using the initiator and the
dataset's interaction-order metadata.
The resulting four views are automatically derived supervision (checked on a sample below), and multiple valid candidates
are retained per view.
A sample lacking a valid target for a view is excluded only from that
view; sequences without any semantic target are discarded.
After filtering one test motion shorter than 15 frames and ten without
semantic views, methods without a joint-geometry requirement evaluate
1{,}697 Overall/Relation, 1{,}488 P1 and 1{,}484 P2 sequences; the learned
pair branch additionally excludes ten sequences with invalid or
unavailable synchronized skeletons, leaving 1{,}687 / 1{,}480 / 1{,}476 /
1{,}687.
These are validity masks on the same split, not different test splits, and
relation-type comparisons use the common 1{,}687 sequences.

\emph{Annotation prompt.}
The prompt asks the LLM to process each reference caption independently and return, as JSON, the two caption-local participants (A is the first distinct participant introduced in that caption, B the other; A is not assumed to be the initiator), each participant's observable behaviour, the relation between them, the likely initiator with a confidence, and an interaction type (asymmetric, reciprocal, symmetric or role-neutral).
All content must come from the current caption: no details are carried over between captions and no causal wording is added.
A participant who is only touched, pushed or approached, with no stated response, receives \texttt{null} rather than a passive description, while an explicitly stated response is written into that participant's own field.
The initiator is the participant performing the primary directed action; mention order, grammatical subject, and who moves first are not evidence, a responsive action is not initiation, and when no reliable initiator exists the initiator fields are \texttt{null}.
The prompt contains six worked examples (push, hug, handshake, responsive handshake, pat, role-neutral poses) and a 16-item checklist.
One of its examples is (abridged to the fields shown):
\begin{quote}\footnotesize\raggedright
\textbf{Input:} The first person stands still while the second person pushes the first person's shoulders, causing the first person to step backward.\\
\textbf{A:} ``Participant A stands still and steps backward after being pushed.''\ \textbf{B:} ``Participant B pushes Participant A's shoulders.''\\
\textbf{Relation:} ``Participant B pushes Participant A's shoulders, causing Participant A to step backward.''\\
\textbf{Initiator:} B (high).\ \textbf{Type:} asymmetric.
\end{quote}
\noindent
Over the Inter-X captions, the resulting annotations contain
34{,}065 non-empty relation descriptions, of which 27{,}258 ($\approx$80\%)
mention both participants explicitly; each relation is tagged as
asymmetric, reciprocal, symmetric, or role-neutral.

\paragraph{Human verification.}
We randomly sampled 100 test sequences (one reference caption each, fixed seed), and two annotators independently checked the canonical views against the source caption and a colour-coded skeleton animation of the two people, blind to each other's labels.
A sample passes if the $\Pone$ and $\Ptwo$ descriptions are attributed to the correct person, the relation has the correct direction, and no action is invented or dropped.
The annotators pass 91 and 90 of the 100 samples (95\% Wilson CIs [83.8, 95.2] and [82.6, 94.5]), agree on 97 (Cohen's $\kappa=0.83$), and 89 samples pass under both.
In most failures the description follows the caption but is attached to the other person's track, which is determined by the dataset's interaction-order annotation; the others are source captions that do not match the motion.

\paragraph{Participant-swap protocol.}
For a test interaction $x=(x_1,x_2)$ we also form the swapped input
$\tau x=(x_2,x_1)$, which contains the same two people with exchanged
slots; no model sees swapped inputs during training.
For each input we generate the $\Pone$ and $\Ptwo$ descriptions,
$\hat y_1(\cdot)$ and $\hat y_2(\cdot)$, from the same trained model.
After the swap, the $\Pone$ description must describe the person who
previously occupied slot $\Ptwo$ and vice versa.
Because a description may mention both participants, we first relabel
the original outputs by $\Tpi$ (exchanging ``Person~1'' and
``Person~2''), so that a correct model produces identical content up to
slot names.
Fig.~\ref{fig:swap_counterfactual} summarizes the comparisons.
$S_{\mathrm{correct}}$ compares each relabelled original output with
the swapped output \emph{about the same person} (the other slot), and
$S_{\mathrm{wrong}}$ with the output \emph{of the same slot}, which
describes the other person; their difference
$G_{\mathrm{swap}}=S_{\mathrm{correct}}-S_{\mathrm{wrong}}$ measures
whether descriptions follow people rather than slots.
Swap-Tgt instead scores the swapped outputs against the exchanged
canonical references and therefore measures absolute quality.
All similarities $s$ are BERTScore-F1 (RoBERTa-large, baseline-rescaled,
no idf weighting), computed per test interaction in both swap
directions and averaged; the direction-specific values are reported in
the complete swap tables.
Interactions for which a participant reference is missing are excluded
from the corresponding direction.
The $\Rel$ view is not tied to a slot: it must follow the direction of
the relation, which we measure with the follow rate of
Sec.~\ref{sec:grounding_diagnosis} of the main paper.
The follow rate does not depend on the judge: replacing the BERTScore
comparison by a rule that reads the actor off each Relation text (including
passive voice) gives 83.4\% for InterRole and 22.7--25.3\% for the baselines
on asymmetric interactions, and the two judges agree on 98.5--99.9\% of the
decidable cases.
$G_{\mathrm{swap}}$, the identical-description rate and the follow rate
compare only the model's own outputs, so at evaluation time they do not
depend on the canonical role assignment.
Frozen probes use the same swapped inputs: the action probe should be
invariant, the directed-role probe should reverse, and Pair-C requires
both members of an original--swapped pair to be correct.

\begin{figure*}[t]
\centering
\input{figures/fig_swap_protocol}
\caption{Participant-swap evaluation. The two motion streams exchange
their slots while keeping their identity; colour marks the person, not the
slot. Green solid links compare each (relabelled) original output with the
swapped output about the same person ($S_{\mathrm{correct}}$), red dashed
links compare it with the output of the same slot, which describes the
other person ($S_{\mathrm{wrong}}$). Their difference is the swap gain
$G_{\mathrm{swap}}$. Swap-Tgt compares the swapped outputs with the
exchanged human references. Table: how to read each score. A slot-bound
model has $S_{\mathrm{wrong}}\approx S_{\mathrm{correct}}$, so $G_{\mathrm{swap}}\approx0$;
a grounded model keeps $S_{\mathrm{correct}}$ and Swap-Tgt high and lowers
$S_{\mathrm{wrong}}$. $S_{\mathrm{correct}}$ alone does not discriminate
(both models are high), which is why it is read with $S_{\mathrm{wrong}}$.}
\label{fig:swap_counterfactual}
\end{figure*}

\paragraph{Frozen linear probes.}
Both probes read the same frozen motion representation of each method and
the same swapped inputs, but their labels behave differently under the
swap, so the consistency metric differs (Tab.~\ref{tab:probe_protocol}).
The fine-grained \emph{action} label does not change when the two streams
are exchanged, so the action probe should be \emph{invariant}: Inv.\ is the
fraction of interactions whose predicted action is unchanged by the swap.
The directed \emph{active--passive} label reverses, so the role probe should
be \emph{equivariant}: Equiv.\ is the fraction whose predicted order flips
consistently with the swap.
Neither measures correctness: a probe that always predicts the same wrong
class is perfectly invariant, and a slot-bound role probe, which gives the
same answer before and after the swap, has Equiv.\ near zero.
We therefore always read them together with balanced accuracy on original
(BAcc) and swapped inputs (Swap BAcc) and with Pair-C, which requires both
members of an original--swapped pair to be correct and is the strictest
measure.
For the binary role label, chance is $.50$ for Equiv.\ and $.25$ for Pair-C.

\begin{table}[t]
\centering
\footnotesize
\setlength{\tabcolsep}{4pt}
\caption{Frozen-probe protocol. Under the participant swap the action label
stays the same and the directed role label reverses, hence Inv.\ for the
former and Equiv.\ for the latter. For all metrics a good model scores high.}
\label{tab:probe_protocol}
\resizebox{\columnwidth}{!}{%
\begin{tabular}{@{}lccc@{}}
\toprule
Probe & Label under swap & Consistency & Also reported\\
\midrule
Action & unchanged & Inv.\ (prediction unchanged) & BAcc, Swap BAcc, Pair-C\\
Directed role & reversed & Equiv.\ (prediction flips) & BAcc, Swap BAcc, Pair-C\\
\bottomrule
\end{tabular}}
\end{table}

\section{Implementation Details}
\label{supp:impl}
\paragraph{Motion features and architecture.}
Main experiments use 30-fps, 336-dimensional per-person SMPL-X features:
55 rotations in 6D form plus root translation and frame displacement,
without feature standardization.
We retain 15--300 frames, randomly cropping longer sequences in training
and centre-cropping at test time. Pair geometry uses 61 selected skeleton
joints, synchronized by downsampling the 120-fps skeletons by four.
The motion Transformer has 4 layers, width 768, 8 heads, feed-forward
width 2{,}048, dropout 0.1, and 16 learned pooling queries.
InterRole shares this encoder between the two people. Its person learner
has one shared partner-conditioning block; the pair branch has a
one-layer geometry encoder and a two-layer pair Transformer.
The uncompressed contextual reader has 2 layers and one scalar gate per
view initialized to $\gamma_\view^{(0)}=0.01$.
All evidence attention modules use width 768 and 8 heads; a two-layer
MLP with hidden width 2{,}048 projects the prefix to the LLM embedding
dimension.

\paragraph{Language model and optimization.}
We use Qwen2.5-1.5B-Instruct~\cite{qwen25} and Gemma3-1B-IT~\cite{gemma3},
freezing base weights and training LoRA~\cite{lora} with rank 16,
$\alpha=32$ and dropout 0.05 on the query/key/value/output attention
projections and gate/up/down MLP projections.
All main runs use a 10-epoch budget.
Trainable parameters (Qwen2.5-1.5B; motion modules, projector and LoRA): 48.1M for Instruction and Task-Query, 81.9M for LLaMMo, and 86.4M for InterRole.
AdamW uses $10^{-4}$ for motion, evidence, embedding and projection
modules and $2\!\times\!10^{-4}$ for LoRA, weight decay 0.01,
500 warm-up steps, cosine decay, and gradient clipping at norm 1.
Batch size is 8 with 4 gradient-accumulation steps (effective batch 32).
Training samples a view uniformly among its available targets and a
non-empty reference for that view. Targets are limited to 96 tokens;
test decoding is greedy with at most 128 new tokens.

\paragraph{Baselines.}
LLaMMo is a lightweight adaptation with a shared Conv1D-ResNet human
encoder, instruction-aware soft frame selection, temporal/social
attention, and motion--instruction cross-attention; its 16 pooled
interaction tokens are gated residual additions to the 16 global
Transformer tokens, and only the fixed task instruction, never the target
caption, conditions this interaction branch.

Checkpoints are selected by validation loss.
Participant-swap augmentation is disabled for every model in the main
paper; it is used only in the study of Sec.~\ref{sec:swap_aug} of the main paper.

\section{Controls and Alternatives}
\label{supp:role_marker}
\paragraph{LLaMMo + T/P.}
The control, a variant of LLaMMo (\emph{LLaMMo + T/P}), keeps the LLaMMo
architecture and changes a single step.
Its social-temporal module applies temporal and social attention but
returns two separate person streams; the control adds the view-relative
target/partner embeddings of InterRole to these streams (for the
$\Pone/\Ptwo$ views only, as in InterRole).
The marked streams then follow the unchanged baseline path: they are
concatenated in a fixed order, contextualized jointly, and pooled by 16
learned queries into shared interaction tokens without per-person
boundaries, which are added to the global tokens through a gated
residual.
The control therefore tests role markers placed after social fusion but
before shared pooling, without turning the baseline into InterRole.
All training and decoding settings match the original LLaMMo run
(Qwen2.5-1.5B).

\paragraph{LLaMMo + LB.}
This variant applies late binding to LLaMMo: it removes the pooling of the interaction branch into 16 shared query tokens and feeds the P1, P2 and global tokens (16 each) to the language model separately.
It matches InterRole's prefix length (52 tokens) and has a comparable trainable parameter count (79.5M vs.\ 86.4M), so it separates the effect of late binding from prefix bandwidth and capacity.
All other settings match the original LLaMMo run.
Their metrics are in the main paper (Fig.~\ref{fig:pareto}).

As a sanity check on Qwen2.5-1.5B, the $\Rel$ view receives no marker and should behave
as in the original baseline.
The results confirm this on the same 1{,}687 test samples and BERTScore
protocol as in the main paper (asymmetric relations,
$N=1{,}108$): the follow rate is 26.5\% [23.9, 29.3] for LLaMMo and
29.2\% [26.5, 31.9] for the control (InterRole: 87.1\% [85.1, 89.0]),
ties are 4.3\%, 4.4\% and 5.7\%, and the direction gain stays negative
for both ($-.021$ and $-.019$; InterRole $+.040$).

\paragraph{Swap augmentation.}
\label{supp:swap_aug}
All augmented runs select the checkpoint by
validation loss (epoch 12 for both 20-epoch runs).
Table~\ref{tab:supp_swap_aug_full} gives the complete results.
For the LLaMMo baseline, augmentation leaves Swap-Tgt (.544 vs.\ .543)
and both probes essentially unchanged at 10 epochs; at 20 epochs its
action and directed-role BAcc even drop (.398 and .517).
For InterRole, augmentation at 20 epochs raises Swap-Tgt from .594 to
.597 and action BAcc from .596 to .613, whereas directed-role BAcc
(.716 vs.\ .723) and Relation CIDEr (.529 vs.\ .534) do not exceed the
unaugmented model.
At 10 epochs, augmentation raises $S_{\mathrm{correct}}$ and
$S_{\mathrm{wrong}}$ together on both architectures while
participant-view CIDEr drops.

\begin{table*}[t]
\centering
\small
\setlength{\tabcolsep}{4.5pt}
\caption{
Architectural grounding vs.\ participant-swap augmentation
(Qwen2.5-1.5B). Ep.\ is the training budget; checkpoints are selected
by validation loss. Generation columns report CIDEr; counterfactual
columns BERTScore-F1; Person is the mean of $\Pone$ and $\Ptwo$.
}
\label{tab:supp_swap_aug_full}
\begin{tabular}{lcc|ccc|cccc|cc}
\toprule
Method & Aug. & Ep.
& Ovr & Person & Rel
& Swap-Tgt & $S_{\mathrm{corr}}$ & $S_{\mathrm{wrong}}\!\downarrow$ & $G_{\mathrm{swap}}$
& Act.\ BAcc & Role BAcc \\
\midrule
LLaMMo & & 10 & 0.177 & 0.329 & 0.316 & 0.544 & 0.822 & 0.793 & 0.029 & 0.423 & 0.561 \\
LLaMMo & \checkmark & 10 & 0.173 & 0.302 & 0.332 & 0.543 & 0.896 & 0.861 & 0.034 & 0.425 & 0.564 \\
LLaMMo & \checkmark & 20 & 0.193 & 0.311 & 0.332 & 0.545 & \textbf{0.898} & 0.863 & 0.035 & 0.398 & 0.517 \\
\midrule
InterRole    & & 10 & 0.252 & 0.518 & 0.534 & 0.594 & 0.816 & \textbf{0.603} & 0.214 & 0.596 & \textbf{0.723} \\
InterRole    & \checkmark & 10 & \textbf{0.266} & 0.484 & \textbf{0.539} & 0.593 & 0.853 & 0.630 & 0.223 & 0.608 & 0.715 \\
InterRole    & \checkmark & 20 & 0.260 & \textbf{0.527} & 0.529 & \textbf{0.597} & 0.846 & 0.614 & \textbf{0.232} & \textbf{0.613} & 0.716 \\
\bottomrule
\end{tabular}
\end{table*}

Augmentation also makes the two person descriptions of an interaction more similar (1{,}467 test samples shared by all four runs): their BERTScore similarity $s(\hat y_1(x),\hat y_2(x))$ rises from .859 to .880 for LLaMMo ($+.020$, 95\% CI [.010, .030]) and from .618 to .634 for InterRole ($+.016$ [.006, .026]), and the fraction of exactly identical descriptions after aligning person identifiers from 67.8\% to 74.2\% and from 19.7\% to 24.8\%.

\paragraph{Slot-order prior.}
The swap test would penalize a model that exploits a regularity of the training data, such as ``the person in slot~1 is usually the actor''.
There is none: for asymmetric interactions, the actor is the person in slot~1 in 49.4\% (3{,}371 of 6{,}828) of the training sequences and in slot~2 in 50.6\%, and the raw interaction-order annotation is equally balanced over all 9{,}061 training sequences (51.1\%).
Slot order therefore carries no role prior, and a model that writes the same description for both people (67.8\% for LLaMMo even on unswapped inputs) is not exploiting a prior; it simply fails to distinguish the participants.

\paragraph{Query-time reordering.}
Another way to route the view before fusion is to place the queried participant in slot~1 for each $\Pone/\Ptwo$ query.
This makes the view known before the streams are fused and makes the participant views swap-consistent by construction, but it re-encodes and re-fuses the input for each queried participant and is undefined for the Overall and Relation views, which have no target participant, so the direction of the relation stays tied to the input slots.
InterRole computes its evidence bank once; only the reader is view-specific, and it serves all four views from one input, leaving the swap behaviour to the structure of its evidence.

\section{Additional Analyses of InterRole}
\label{supp:ablations}

\paragraph{Calibration gate.}
The reader updates the evidence with a learnable gate (Eq.~\eqref{eq:calibration} of the main paper).
Table~\ref{tab:supp_gate_granularity} compares one scalar per view, $\gamma_\view$, with one scalar per view and stream, $\gamma_{\view,s}$.
On Qwen2.5-1.5B, $\gamma_\view$ is higher on generation (all four views), Swap-Tgt, $G_{\mathrm{swap}}$, action BAcc and role BAcc, and lower only on role Pair-C, with fewer gate parameters; we therefore use $\gamma_\view$.

\begin{table}[t]
\centering
\footnotesize
\setlength{\tabcolsep}{6pt}
\caption{Calibration gate: one scalar per view ($\gamma_\view$) or per view and stream ($\gamma_{\view,s}$), Qwen2.5-1.5B (BERTScore-F1 for generation and counterfactual rows; best in bold).}
\label{tab:supp_gate_granularity}
\begin{tabular}{l|cc}
\toprule
& $\gamma_\view$ & $\gamma_{\view,s}$ \\
\midrule
Overall & \textbf{0.548} & 0.543 \\
P1 & \textbf{0.597} & 0.592 \\
P2 & \textbf{0.605} & 0.600 \\
Rel & \textbf{0.595} & 0.589 \\
\midrule
Swap-Tgt & \textbf{0.594} & 0.588 \\
$G_{\mathrm{swap}}$ & \textbf{0.214} & 0.193 \\
\midrule
Action BAcc & \textbf{0.596} & 0.588 \\
Role BAcc & \textbf{0.723} & 0.713 \\
Role Pair-C & 0.569 & \textbf{0.583} \\
\bottomrule
\end{tabular}
\end{table}

\paragraph{Participant representations.}
We measure how alike the two person streams are and how a swap changes the frozen representation.
The cosine similarity between the token-pooled person streams (before slot and role embeddings and the reader; about 1.7k test motions) has mean (median) .455 (.560) for $(\Sone,\Stwo)$ of the Raw checkpoint, and .458 (.585) before and .526 (.646) after the learner for $(\Sone,\Stwo)$ and $(\Zone,\Ztwo)$ of the Learned checkpoint: partner conditioning raises the similarity by $+.068$ (95\% CI $[.063,.074]$; higher for 79.1\% of samples).

To see how close the baselines are to the invariant limit of Prop.~\ref{prop:fusion} of the main paper, Table~\ref{tab:supp_invariance} compares the frozen probe representation $h(x)$ with that of the swapped input $h(\tau x)$ (standardized with training statistics): mean cosine similarity, and the distance $\lVert h(x)-h(\tau x)\rVert$ relative to the median distance to other test interactions.
A swap moves the baselines' representations by only 0.15--0.16 of a typical inter-sample distance (cosine .97), i.e.\ they are nearly, though not strictly, invariant.
InterRole's representation, in contrast, is not swap-invariant (cosine .513).
Whether the change under a swap is the correct one is measured by the role probe and Pair-C of the main paper.

\begin{table}[t]
\centering
\footnotesize
\setlength{\tabcolsep}{5pt}
\caption{Effect of the participant swap on the frozen probe representation (Qwen2.5-1.5B, test set).
Cos.: cosine similarity of $h(x)$ and $h(\tau x)$; Dist.: $\lVert h(x)-h(\tau x)\rVert$ divided by the median distance to other samples.}
\label{tab:supp_invariance}
\begin{tabular}{l|cc}
\toprule
Method & Cos. & Dist. \\
\midrule
Instruction Cond. & 0.972 & 0.147 \\
Task-Query Cond. & 0.971 & 0.150 \\
LLaMMo & 0.967 & 0.163 \\
LLaMMo + T/P & 0.971 & 0.153 \\
InterRole & 0.513 & 0.636 \\
\bottomrule
\end{tabular}
\end{table}

\section{Cross-Dataset Details}
\label{supp:cross}

Both datasets are mapped to 20 canonical joints with positions and
velocities (120 dimensions per person, normalized with Inter-X training
statistics only).
Table~\ref{tab:supp_interhuman} extends Table~\ref{tab:cross_dataset} of the main paper
with confidence intervals and per-direction results (1{,}176 test
sequences; no empty predictions).

\begin{table}[t]
\centering
\footnotesize
\setlength{\tabcolsep}{3pt}
\caption{Zero-shot InterHuman diagnostics (1{,}176 test sequences, no empty predictions) with 95\% bootstrap CIs.
Identical (semantic): BERTScore threshold 0.9.}
\label{tab:supp_interhuman}
\resizebox{\columnwidth}{!}{%
\begin{tabular}{l|cc}
\toprule
& LLaMMo & InterRole \\
\midrule
Overall BERT (vs.\ references) & 0.291 [0.286, 0.297] & \textbf{0.301} [0.295, 0.307] \\
$S_{\mathrm{correct}}\uparrow$ & 0.582 [0.572, 0.592] & \textbf{0.613} [0.603, 0.622] \\
\quad P1$\to$P2 / P2$\to$P1 & 0.590 / 0.575 & 0.611 / 0.614 \\
$S_{\mathrm{wrong}}\downarrow$ & 0.549 [0.539, 0.557] & \textbf{0.532} [0.523, 0.540] \\
\quad P1 / P2 wrong pair & 0.548 / 0.549 & 0.531 / 0.532 \\
$G_{\mathrm{swap}}\uparrow$ & 0.034 [0.032, 0.035] & \textbf{0.081} [0.074, 0.088] \\
Identical, exact (\%)$\downarrow$ & 74.6 [72.1, 76.9] & \textbf{27.9} [25.3, 30.4] \\
Identical, semantic (\%)$\downarrow$ & 78.9 & \textbf{31.5} \\
Mean $s(\hat y_1(x),\hat y_2(x))\downarrow$ & 0.932 & \textbf{0.714} \\
\midrule
Relation: reversed / not reversed & 0.527 / 0.537 & 0.601 / 0.581 \\
Relation directional gain$\uparrow$ & $-0.010$ [$-0.012$, $-0.008$] & $\mathbf{+0.020}$ [0.018, 0.022] \\
\bottomrule
\end{tabular}}
\end{table}
\section{Frozen-Representation Retrieval Probe}
\label{supp:retrieval}

To test whether the representations are useful beyond captioning, we
probe them with text--motion retrieval (Sec.~\ref{sec:retrieval} of the
main paper).
This is a frozen-representation probe, not zero-shot retrieval: all
captioning models stay frozen, and only two linear projections are
trained.

\begin{table*}[t]
\centering
\footnotesize
\setlength{\tabcolsep}{2.6pt}
\caption{Frozen-representation retrieval probe (text$\to$motion,
Qwen2.5-1.5B, Inter-X test, 1{,}697 interactions). R@1 (\%) under TMR
protocols (a) full gallery, (b) full gallery with text-similarity
threshold, (d) batches of 32, and median rank under (a). Random R@1:
0.06 (a), 3.1 (d); random MedR (a): 849. Best in \textbf{bold}.}
\label{tab:supp_retrieval}
\begin{tabular}{l|cccc|cccc|cccc|cccc}
\toprule
& \multicolumn{4}{c|}{$\Ovr$} & \multicolumn{4}{c|}{$\Pone$}
& \multicolumn{4}{c|}{$\Ptwo$} & \multicolumn{4}{c}{$\Rel$} \\
Method & (a) & (b) & (d) & MedR & (a) & (b) & (d) & MedR
& (a) & (b) & (d) & MedR & (a) & (b) & (d) & MedR \\
\midrule
Instruction Cond. & 3.4 & 8.0 & 45.4 & 42 & 2.0 & 6.9 & 35.3 & 75
  & 2.4 & 8.1 & 36.5 & 76 & 2.6 & 10.4 & 39.9 & 59 \\
Task-Query Cond.  & 2.2 & 6.1 & 38.3 & 62 & 1.4 & 5.3 & 31.6 & 91
  & 1.7 & 6.7 & 31.7 & 92 & 2.3 & 10.5 & 34.7 & 77 \\
LLaMMo            & 3.9 & 9.1 & 48.5 & 38 & 1.8 & 7.5 & 38.5 & 63
  & 2.3 & 9.0 & 39.7 & 62 & 2.5 & 13.0 & 42.6 & 48 \\
\textbf{InterRole} & \textbf{6.7} & \textbf{14.3} & \textbf{60.8} & \textbf{20}
  & \textbf{3.7} & \textbf{12.0} & \textbf{48.3} & \textbf{38}
  & \textbf{3.6} & \textbf{14.3} & \textbf{49.3} & \textbf{35}
  & \textbf{6.6} & \textbf{20.9} & \textbf{57.9} & \textbf{23} \\
\bottomrule
\end{tabular}
\end{table*}

\begin{figure*}[t]
\centering
\input{figures/fig_supp_2afc}
\caption{Who-did-what retrieval on asymmetric interactions
($N=1{,}097/1{,}094/1{,}108$ for $\Pone/\Ptwo/\Rel$).
\emph{Left:} accuracy of choosing the original motion over its
participant-swapped version (2AFC), with 95\% bootstrap CIs; the dashed
line is chance (50\%).
\emph{Right:} role retention, the R@1 in 16-interaction batches with
their 16 swapped versions added (Hard) divided by the R@1 without them
(Std); a representation that ignores who does what retains about half.
For InterRole, Std/Hard R@1 is 61.0/39.7 ($\Pone$), 61.0/36.0 ($\Ptwo$)
and 68.5/48.4 ($\Rel$); for the baselines, Std R@1 is 41.8--51.3 and
Hard R@1 22.1--30.5.}
\label{fig:supp_retrieval_2afc}
\end{figure*}
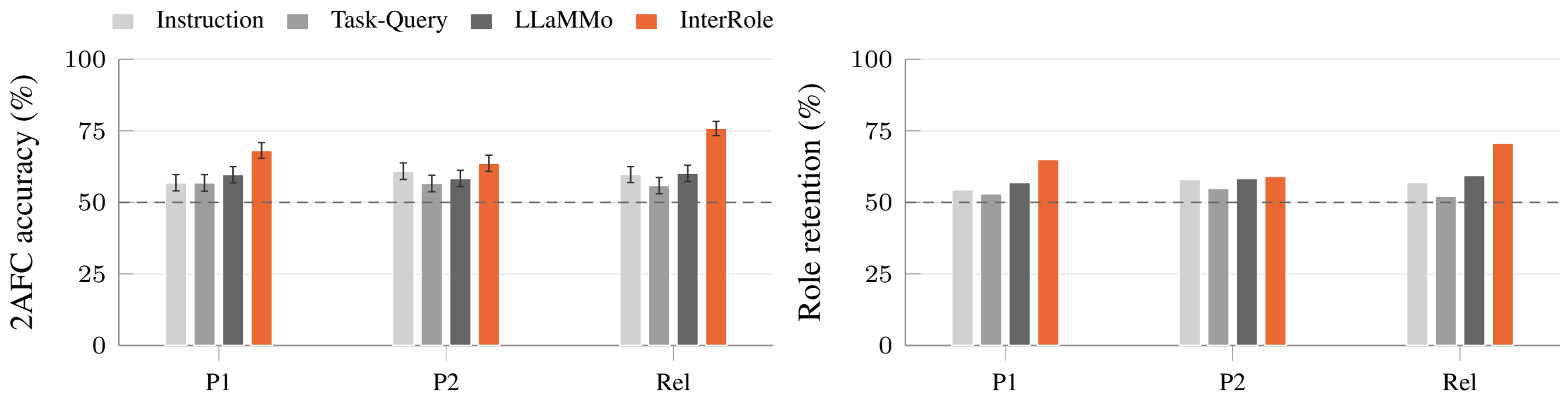

\paragraph{Representations.}
\emph{Motion.} For view $\view$ we take each method's projected motion
prefix $\bm P_\view(\mathcal M)\in\R^{N_\view\times d}$ ($d=1{,}536$),
i.e.\ the tokens produced by the method's own projector before they
enter the language model; each method keeps its own prefix structure
(view-independent for Instruction Conditioning, with a view token for
Task-Query Conditioning, the instruction-conditioned fused prefix for
LLaMMo, and the view-conditioned evidence prefix for InterRole).
The standard retrieval probe uses its global mean,
$\bm r_M=\operatorname{Mean}(\bm P_\view)\in\R^{1{,}536}$.
The who-did-what probe keeps coarse token order: one-dimensional
adaptive average pooling along the token axis yields eight ordered bins
$I_0,\dots,I_7$, with
$I_k=\{\lfloor kN_\view/8\rfloor,\dots,\lceil (k{+}1)N_\view/8\rceil-1\}$,
and
\begin{equation}
\bm r_M=\bigl[\operatorname{Mean}(\bm P_\view[I_0]);\dots;
\operatorname{Mean}(\bm P_\view[I_7])\bigr]\in\R^{12{,}288}.
\end{equation}
The bins are positional rather than participant-specific; view and
boundary tokens are pooled together with evidence tokens.
\emph{Text.} The query string is the fixed view instruction followed by
``Answer:'' and the reference description (at most 128 tokens).
It is encoded by the method's own frozen Qwen2.5-1.5B-Instruct
checkpoint, including its LoRA adapters, and $\bm r_T\in\R^{1{,}536}$
is the final-layer hidden state at the last non-padding token.

\paragraph{Probe.}
Both representations are standardized with per-dimension training-set
statistics and mapped by independent bias-free linear layers to 256
dimensions,
\begin{equation}
\bm z_M=\frac{\bm W_M\operatorname{Std}(\bm r_M)}
{\lVert\bm W_M\operatorname{Std}(\bm r_M)\rVert},\qquad
\bm z_T=\frac{\bm W_T\operatorname{Std}(\bm r_T)}
{\lVert\bm W_T\operatorname{Std}(\bm r_T)\rVert},
\end{equation}
and candidates are ranked by cosine similarity.
The projections are trained with symmetric InfoNCE (temperature 0.07),
and the checkpoint is selected by mean validation R@10.
Swapped motions are never used during probe training.
For the who-did-what probe, one probe per method is shared across the
$\Pone$, $\Ptwo$ and $\Rel$ views.
All methods use identical test identifiers, gallery order, projection
heads and training hyperparameters.
Each view is evaluated separately: a text query of view $\view$
retrieves among the view-$\view$ motion representations of all 1{,}697
test interactions.
We follow the protocols of TMR~\cite{tmr}: (a) the full gallery;
(b) the full gallery, counting a retrieved motion as correct when its
reference text is near-identical to the query (MPNet~\cite{mpnet}
similarity $\ge 0.95$ after rescaling to $[0,1]$); and (d) batches of 32
candidates.

Table~\ref{tab:supp_retrieval} shows that InterRole retrieves best on
every view and protocol.
The margin over the strongest baseline is largest on the relation view
(R@1 $+7.9$ under (b) and $+15.3$ under (d)), the view whose content
depends most on how the two participants relate.
Because this gallery contains no swapped motions, these numbers
measure the semantic content of each view rather than participant
binding.

\paragraph{Who did what.}
To probe binding with the same retrieval heads, we pose a
two-alternative choice on asymmetric interactions: given the
$\Pone$, $\Ptwo$ or $\Rel$ reference text of an interaction, the
candidates are the interaction's own motion and its participant-swapped
version, which contain the same actions and differ only in who performs
them (random: 50\%).
We also report \emph{role retention}: the R@1 obtained in batches of 16
interactions to which their 16 swapped versions are added, divided by
the R@1 in the same 16-interaction batches without them.
A representation that ignores who does what confuses each motion with
its swap and retains about half of its R@1; one that binds evidence to
participants retains more.
All methods use the same queries, the eight-bin ordered readout above,
and identical probe training.

Fig.~\ref{fig:supp_retrieval_2afc} shows that the baselines stay close
to chance in the two-alternative choice (55.9--60.9\%) and retain only
52--59\% of their R@1, consistent with their slot-bound representations
(Tab.~\ref{tab:grounding_diagnosis} of the main paper).
InterRole selects the correctly oriented motion for 75.9\% of
asymmetric relation queries, 15.7 points above the best baseline with
non-overlapping confidence intervals, and retains 70.7\% of its relation
R@1 (baselines $\le$ 59.4\%); with swapped distractors present, its
relation R@1 (48.4\%) nearly matches the baselines' R@1 \emph{without}
them ($\le$ 51.3\%).
On $\Pone$ it reaches 68.1\% (best baseline 59.7\%) with 65.0\%
retention.
On $\Ptwo$ its lead is small (63.7\% vs.\ 60.9\%, overlapping intervals;
retention 59.1\% vs.\ 58.3\%).

\section{Limitations}
\label{supp:limitations}
The four semantic views are derived from source captions by an
LLM-based pipeline and assigned to the two people using the dataset's interaction-order annotation, so annotation noise remains, mostly descriptions attached to the wrong person (Sec.~\ref{supp:data}); this noise affects all compared methods equally.
Our diagnostics rely on embedding-based text similarity, and the
language models we use are small; the gain in swap consistency is
smaller on the smaller Gemma3-1B decoder, although InterRole still
improves all four views and swap consistency over every baseline there, and
its role probes are as strong as on Qwen2.5-1.5B.
We study two-person interactions; groups introduce permutations beyond
a single swap.

\section{Additional Qualitative Results}
\label{supp:qualitative}
\paragraph{Inter-X.}
Fig.~\ref{fig:qualitative_supp} shows further Inter-X cases in the format of
Fig.~\ref{fig:qualitative} of the main paper, including the Overall description.

\input{figures/fig_qualitative_supp}

\paragraph{InterHuman (zero-shot).}
Fig.~\ref{fig:crossdataset_supp} shows four InterHuman test sequences on which both models, trained on Inter-X only, are applied zero-shot.

\input{figures/fig_crossdataset_supp}

\paragraph{Retrieval.}
Fig.~\ref{fig:retrieval_top5} shows the Top-5 motions retrieved for a $\Rel$ query from the full gallery of originals and their participant-swapped copies, for three of the queries, and Fig.~\ref{fig:retrieval_binary} compares the two alternatives of the who-did-what choice on selected examples.

\input{figures/fig_retrieval_top5}

\paragraph{Robot execution.}
Fig.~\ref{fig:robot_supp} shows the two-robot execution of the retrieved motions for the reciprocal Hug query and for a Pull query (the Push example is in the main paper).

\input{figures/fig_robot_supp}
\input{figures/fig_retrieval_binary}

\clearpage

%% file: figures/fig_swap_protocol.tex
\resizebox{0.9\textwidth}{!}{%
\begin{tikzpicture}[
    >=Latex, font=\small,
    box/.style={draw=#1!75, fill=#1!9, rounded corners=3pt, minimum width=4.5cm, minimum height=0.95cm,
                align=center, inner sep=3pt, line width=0.6pt},
    panel/.style={draw=black!22, fill=black!2.5, rounded corners=5pt, inner sep=7pt},
    pill/.style={fill=white, inner sep=1.6pt, font=\scriptsize, rounded corners=1pt},
    head/.style={font=\bfseries\small, anchor=south},
    ok/.style={draw=prel!90!black, line width=1.3pt, <->},
    bad/.style={draw=qbadc!90!black, line width=0.9pt, densely dashed, <->},
    note/.style={font=\scriptsize, text=black!60, align=center},
    fbox/.style={draw=black!25, rounded corners=3pt, inner xsep=7pt, inner ysep=5pt, align=center}
]
\node[head] at (0,4.55)    {Original input $x=(x_1,x_2)$};
\node[head] at (6.4,4.55)  {Swapped input $\tau x=(x_2,x_1)$};
\node[head] at (12.8,4.55) {Human references};
\node[box=pone] (x1)  at (0,3.65)   {$x_1$\\[-1pt]\scriptsize slot \textsc{P1}};
\node[box=ptwo] (x2)  at (0,2.45)   {$x_2$\\[-1pt]\scriptsize slot \textsc{P2}};
\node[box=pone] (tx1) at (6.4,2.45) {$x_1$\\[-1pt]\scriptsize now in slot \textsc{P2}};
\node[box=ptwo] (tx2) at (6.4,3.65) {$x_2$\\[-1pt]\scriptsize now in slot \textsc{P1}};
\begin{scope}[on background layer]
\node[panel, fit=(x1)(x2)]   (in)  {};
\node[panel, fit=(tx1)(tx2)] (tin) {};
\end{scope}
\draw[->, line width=1.1pt, black!55] (in.east) -- node[pill, above=1pt] {swap $\tau$} (tin.west);
\node[box=pone] (y1)  at (0,0.55)    {$\hat y_1(x)$\\[-1pt]\scriptsize \textsc{P1} description, about $x_1$};
\node[box=ptwo] (y2)  at (0,-0.65)   {$\hat y_2(x)$\\[-1pt]\scriptsize \textsc{P2} description, about $x_2$};
\node[box=pone] (ty2) at (6.4,-0.65) {$\hat y_2(\tau x)$\\[-1pt]\scriptsize \textsc{P2} description, about $x_1$};
\node[box=ptwo] (ty1) at (6.4,0.55)  {$\hat y_1(\tau x)$\\[-1pt]\scriptsize \textsc{P1} description, about $x_2$};
\begin{scope}[on background layer]
\node[panel, fit=(y1)(y2)]   (out)  {};
\node[panel, fit=(ty2)(ty1)] (tout) {};
\end{scope}
\draw[->, line width=1pt, black!55] (in.south)  -- node[pill, right=2pt] {model} (out.north);
\draw[->, line width=1pt, black!55] (tin.south) -- node[pill, right=2pt] {model} (tout.north);
\node[note] at (0,-1.95) {original outputs are relabelled by $\mathcal T_\Pi$\\(Person~1 $\leftrightarrow$ Person~2) before comparison};
\draw[bad] (y1.east) -- node[pill, above=1pt, text=qbadc!90!black] {$S_{\mathrm{wrong}}$} (ty1.west);
\draw[bad] (y2.east) -- node[pill, below=1pt, text=qbadc!90!black] {$S_{\mathrm{wrong}}$} (ty2.west);
\draw[ok] (y1.east) -- (ty2.west);
\draw[ok] (y2.east) -- (ty1.west);
\node[pill, text=prel!90!black] at (3.2,-0.05) {$S_{\mathrm{correct}}$};
\node[box=pone, minimum width=4.2cm] (g1) at (12.8,-0.65)  {$y_1^{*}$\\[-1pt]\scriptsize reference for $x_1$};
\node[box=ptwo, minimum width=4.2cm] (g2) at (12.8,0.55) {$y_2^{*}$\\[-1pt]\scriptsize reference for $x_2$};
\begin{scope}[on background layer]
\node[panel, fit=(g1)(g2)] {};
\end{scope}
\draw[->, line width=1.1pt, black!65] (ty1.east) -- node[pill, above=1pt] {Swap-Tgt} (g2.west);
\draw[->, line width=1.1pt, black!65] (ty2.east) -- node[pill, below=1pt] {Swap-Tgt} (g1.west);
\node[box=pone, minimum width=0.5cm, minimum height=0.32cm] (l1) at (10.7,3.75) {};
\node[anchor=west, font=\scriptsize] at (l1.east) {\ physical participant 1};
\node[box=ptwo, minimum width=0.5cm, minimum height=0.32cm] (l2) at (10.7,3.15) {};
\node[anchor=west, font=\scriptsize] at (l2.east) {\ physical participant 2};
\node[anchor=west, font=\scriptsize, text=black!60] at (10.45,2.55) {colour follows the person, not the slot};
\node[anchor=north, inner sep=0pt] at (6.4,-2.75) {%
\footnotesize
\renewcommand{\arraystretch}{1.28}\setlength{\tabcolsep}{5pt}
\begin{tabular}{@{}l@{\hspace{4pt}}l@{\hspace{10pt}}p{6.3cm}@{\hspace{8pt}}c@{}}
\toprule
\textbf{Score} & \textbf{Definition} & \textbf{A grounded model\ldots} & \textbf{LLaMMo $\to$ InterRole}\\
\midrule
\textcolor{prel!90!black}{$S_{\mathrm{correct}}\uparrow$}
 & $\tfrac12\bigl[s(\mathcal T_\Pi\hat y_1(x),\hat y_2(\tau x))+s(\mathcal T_\Pi\hat y_2(x),\hat y_1(\tau x))\bigr]$
 & keeps the description of a person the same wherever the person is placed (high)
 & 0.822 $\to$ 0.816\\
\textcolor{qbadc!90!black}{$S_{\mathrm{wrong}}\downarrow$}
 & $\tfrac12\bigl[s(\mathcal T_\Pi\hat y_1(x),\hat y_1(\tau x))+s(\mathcal T_\Pi\hat y_2(x),\hat y_2(\tau x))\bigr]$
 & lets the same slot describe the \emph{other} person, so the two outputs differ (low); a high value means outputs stay with the slot
 & 0.793 $\to$ 0.603\\
$G_{\mathrm{swap}}\uparrow$
 & $S_{\mathrm{correct}}-S_{\mathrm{wrong}}$
 & follows people ($G_{\mathrm{swap}}$ large); a slot-bound model has $S_{\mathrm{correct}}\approx S_{\mathrm{wrong}}$ and $G_{\mathrm{swap}}\approx0$
 & 0.029 $\to$ 0.214\\
Swap-Tgt$\uparrow$
 & $\tfrac12\bigl[s(\hat y_1(\tau x),y_2^{*})+s(\hat y_2(\tau x),y_1^{*})\bigr]$
 & still matches the exchanged references (high): quality is kept while the people are swapped
 & 0.544 $\to$ 0.594\\
\bottomrule
\end{tabular}};
\node[note] at (6.4,-7.4) {$s$: BERTScore-F1, averaged over test interactions. Example values: Qwen2.5-1.5B on Inter-X (Tab.~\ref{tab:grounding_diagnosis}).};
\end{tikzpicture}}

%% file: figures/fig_supp_2afc.tex
\pgfplotsset{afcstyle/.style={
  ybar, bar width=6pt, width=0.46\textwidth, height=4.4cm,
  ymin=0, ymax=100, ytick={0,25,50,75,100},
  symbolic x coords={P1,P2,Rel}, xtick=data, enlarge x limits=0.22,
  tick label style={font=\scriptsize}, label style={font=\small},
  ymajorgrids, grid style={black!8}, axis line style={black!45}, tick style={black!45},
  axis x line*=bottom, axis y line*=left,
  legend style={font=\scriptsize, draw=none, fill=none, column sep=4pt,
    at={(0.5,1.04)}, anchor=south, legend columns=4},
  legend image code/.code={\fill[#1] (0cm,-0.08cm) rectangle (0.22cm,0.08cm);},
  error bars/y dir=both, error bars/y explicit,
  error bars/error bar style={black!80, line width=0.45pt},
  error bars/error mark options={rotate=90, mark size=1pt, black!80}}}
\begin{tikzpicture}
\begin{axis}[afcstyle, ylabel={2AFC accuracy (\%)}, name=a]
\addplot[fill=black!18, draw=white] coordinates {(P1,56.8) += (0,2.9) -= (0,2.8) (P2,60.9) += (0,2.9) -= (0,2.9) (Rel,59.7) += (0,2.8) -= (0,2.8)};
\addplot[fill=black!38, draw=white] coordinates {(P1,56.8) += (0,2.9) -= (0,2.9) (P2,56.6) += (0,2.9) -= (0,2.9) (Rel,55.9) += (0,2.8) -= (0,2.9)};
\addplot[fill=black!60, draw=white] coordinates {(P1,59.7) += (0,2.8) -= (0,2.9) (P2,58.3) += (0,2.9) -= (0,2.8) (Rel,60.2) += (0,2.8) -= (0,2.9)};
\addplot[fill=ptwo, draw=white] coordinates {(P1,68.1) += (0,2.8) -= (0,2.7) (P2,63.7) += (0,2.8) -= (0,2.9) (Rel,75.9) += (0,2.4) -= (0,2.6)};
\draw[densely dashed, black!60] (rel axis cs:0,0.5) -- (rel axis cs:1,0.5);
\legend{Instruction, Task-Query, LLaMMo, InterRole}
\end{axis}
\begin{axis}[afcstyle, at={(a.east)}, xshift=1.3cm, anchor=west, ylabel={Role retention (\%)}]
\addplot[fill=black!18, draw=white] coordinates {(P1,54.4) (P2,58.0) (Rel,56.9)};
\addplot[fill=black!38, draw=white] coordinates {(P1,53.0) (P2,54.9) (Rel,52.2)};
\addplot[fill=black!60, draw=white] coordinates {(P1,56.9) (P2,58.3) (Rel,59.4)};
\addplot[fill=ptwo, draw=white] coordinates {(P1,65.0) (P2,59.1) (Rel,70.7)};
\draw[densely dashed, black!60] (rel axis cs:0,0.5) -- (rel axis cs:1,0.5);
\end{axis}
\end{tikzpicture}

%% file: figures/fig_qualitative_supp.tex
\input{figures/sq_macros}

\begin{figure*}[p]
\centering
\fontsize{6.0}{6.7}\selectfont
\sqheader
\sqcasem{\sqcaselabel{Block}{G026T007A038R017}}{{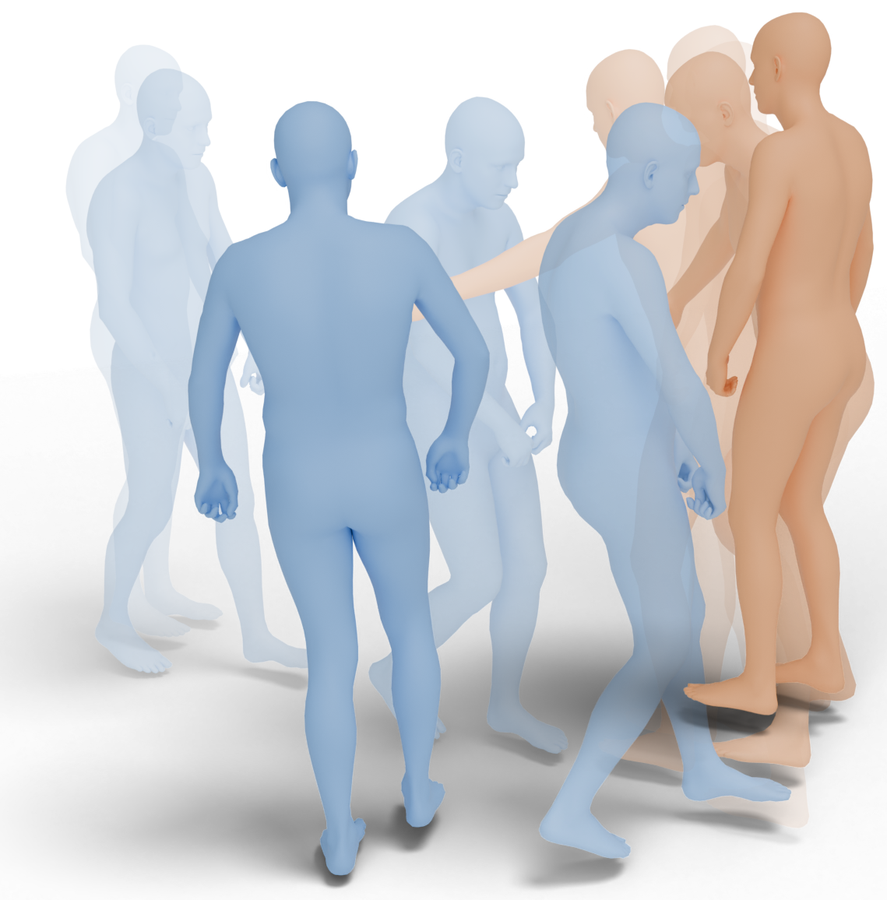}{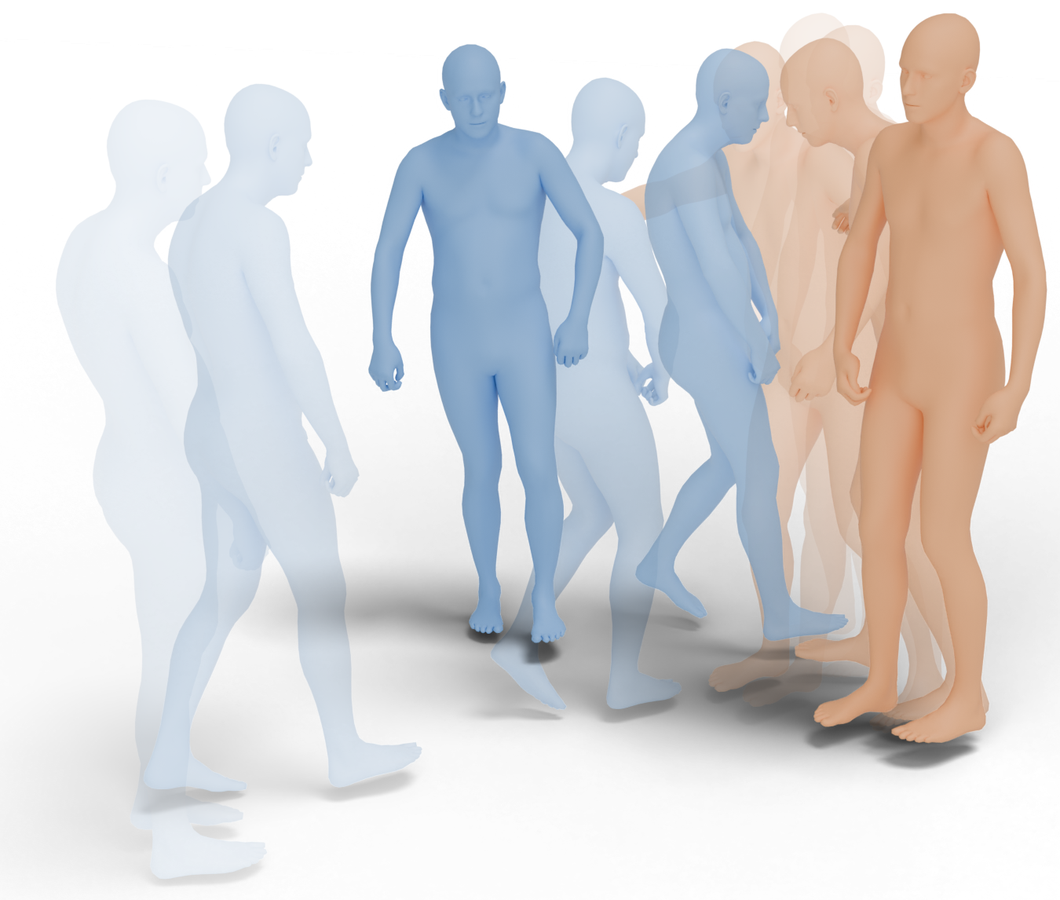}}{{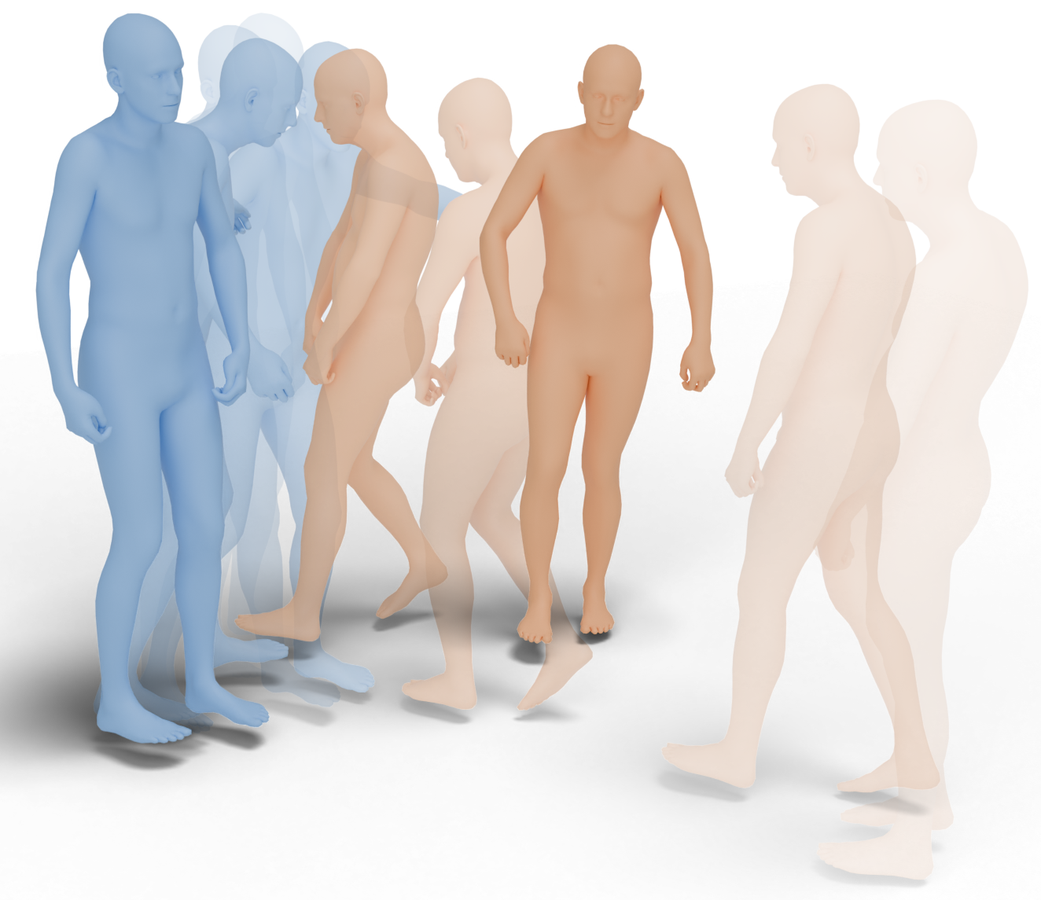}{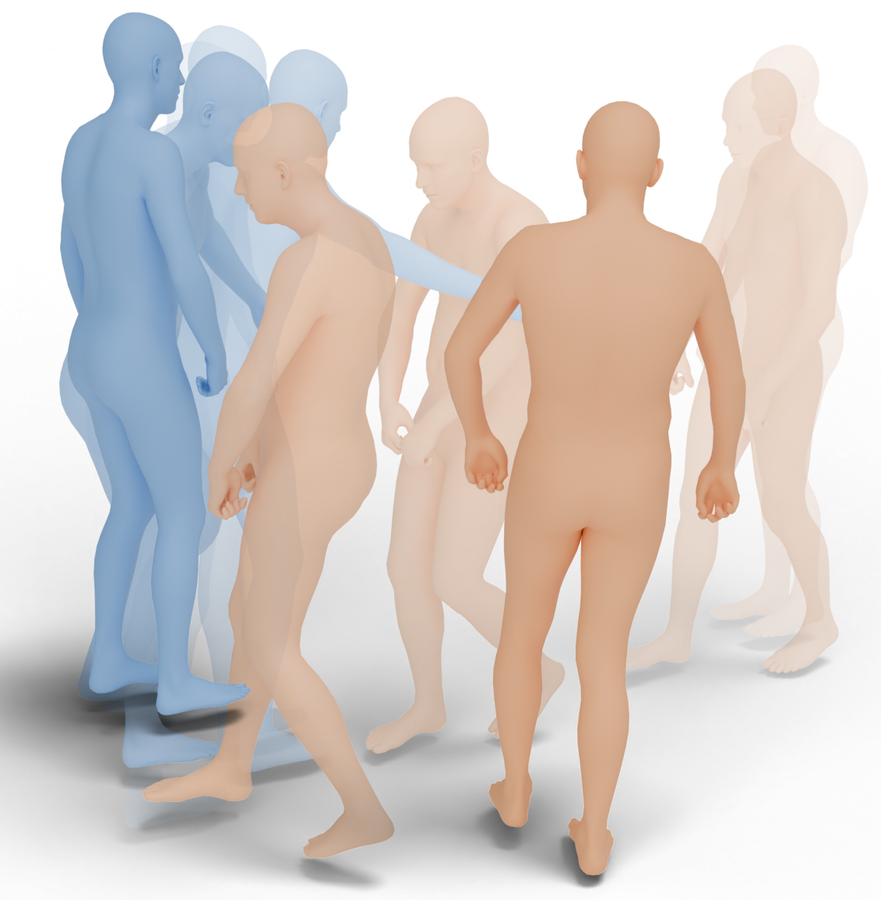}}{%
  \qm{Ref.} & \sqv{Ovr} & \multicolumn{2}{R{3.57in}}{One person wants to move forward, but the other person blocks him by extending both arms.} \\
   & \sqv{P1} & \qA{} wants to move forward but is blocked. & \qA{} extends both arms to block \qB{}. \\
   & \sqv{P2} & \qB{} extends both arms to block \qA{}. & \qB{} wants to move forward but is blocked. \\
   & \sqv{Rel} & \qB{} blocks \qA{} by extending both arms. & \qA{} blocks \qB{} by extending both arms. \\[2pt]
  \qm{LLaMMo} & \sqv{Ovr} & \multicolumn{2}{R{3.57in}}{One person grabs the other person's right hand with both hands and pulls him/her forward.} \\
   & \sqv{P1} & \qA{} \qb{is pulled forward} by \qB{}'s right hand.\qno & \qA{} \qb{is pulled forward} by \qB{}'s right hand.\qno \\
   & \sqv{P2} & \qB{} \qb{is pulled forward} by \qA{}'s hand.\qno & \qB{} \qb{is pulled forward} by \qA{}'s right hand.\qno \\
   & \sqv{Rel} & \qb{\qA{} grabs \qB{}'s right hand} and pulls it towards themselves.\qno & \qb{\qA{} grabs \qB{}'s right arm} with both hands and pulls \qB{} forward.\qno \\[2pt]
  \rowcolor{qours}
  \qm{InterRole} & \sqv{Ovr} & \multicolumn{2}{R{3.57in}}{One person walks forward, and the other person blocks him by stretching out his hands.} \\
  \rowcolor{qours}
   & \sqv{P1} & \qA{} \qg{walks forward}.\qyes & \qA{} walks towards \qB{} and \qg{extends the left hand to block}.\qyes \\
  \rowcolor{qours}
   & \sqv{P2} & \qB{} walks towards \qA{} and \qg{extends the left hand to block}.\qyes & \qB{} walks forward and \qg{is blocked by \qA{}}'s open hands.\qyes \\
  \rowcolor{qours}
   & \sqv{Rel} & \qg{\qB{} blocks \qA{}} by opening his/her arms.\qyes & \qg{\qA{} blocks \qB{}} by opening arms.\qyes \\
}
\sqrule

\sqcasem{\sqcaselabel{Handshake}{G035T000A001R011}}{{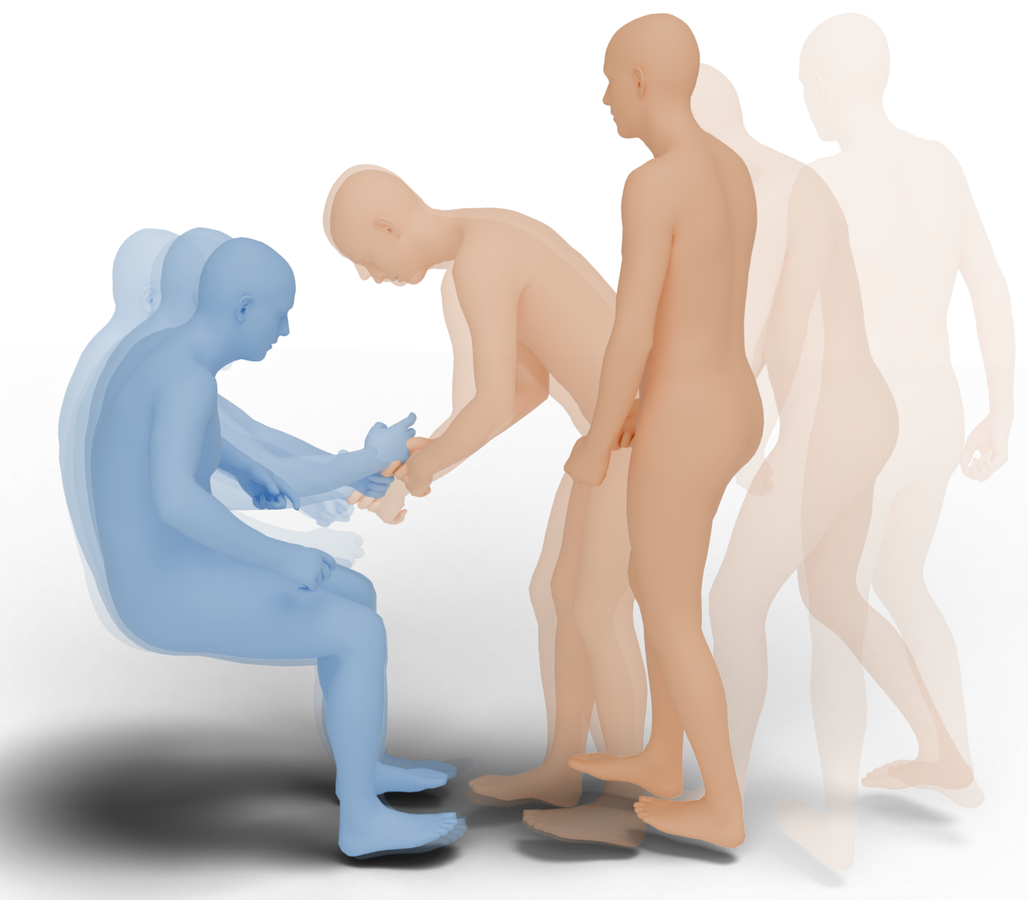}{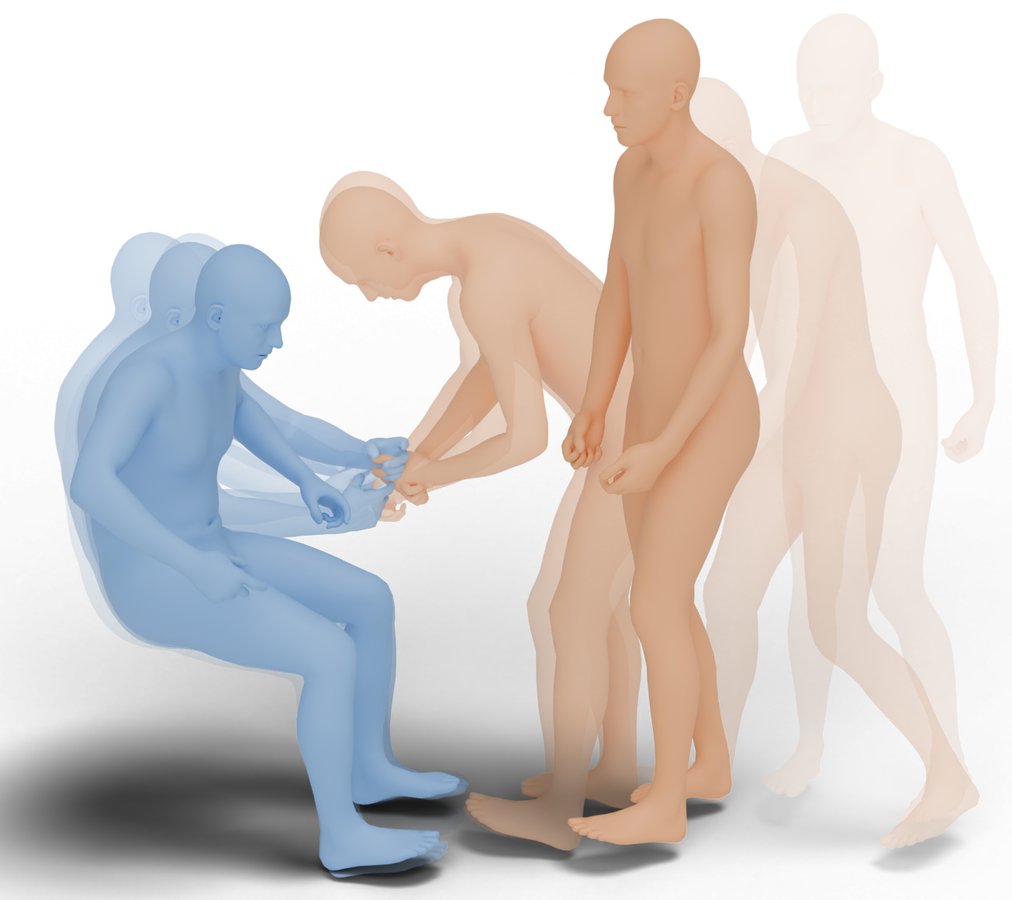}}{{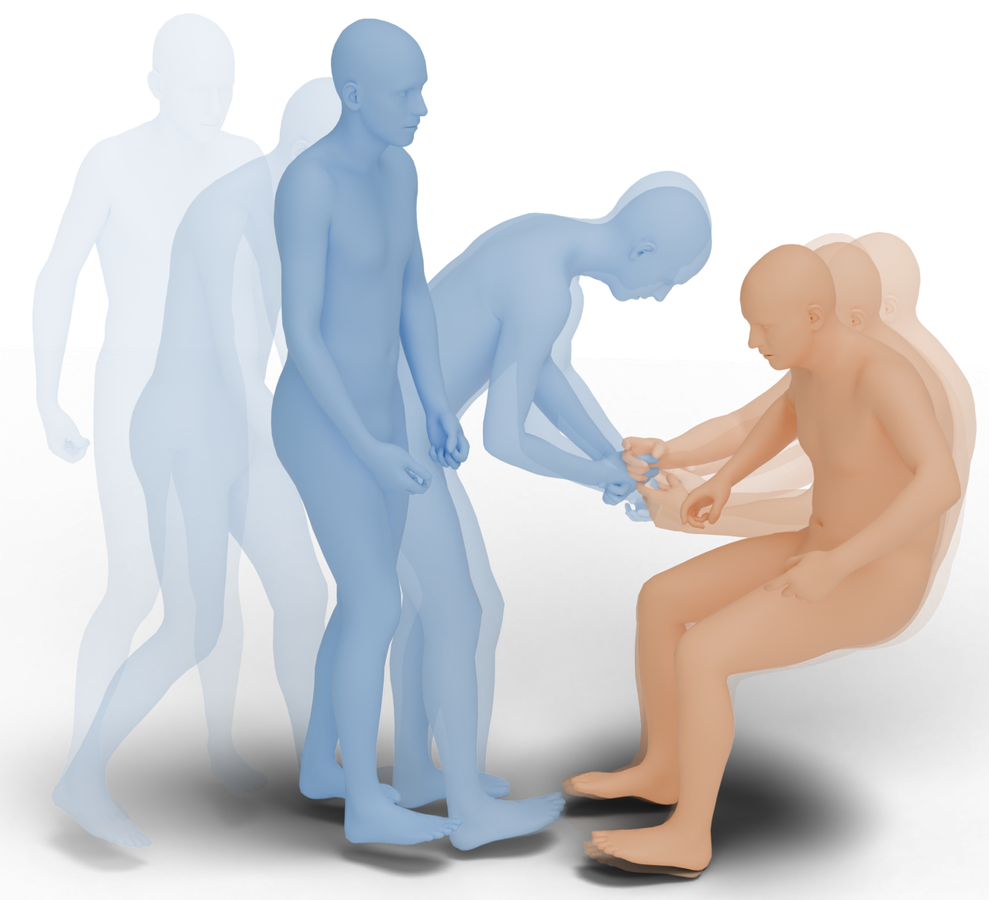}{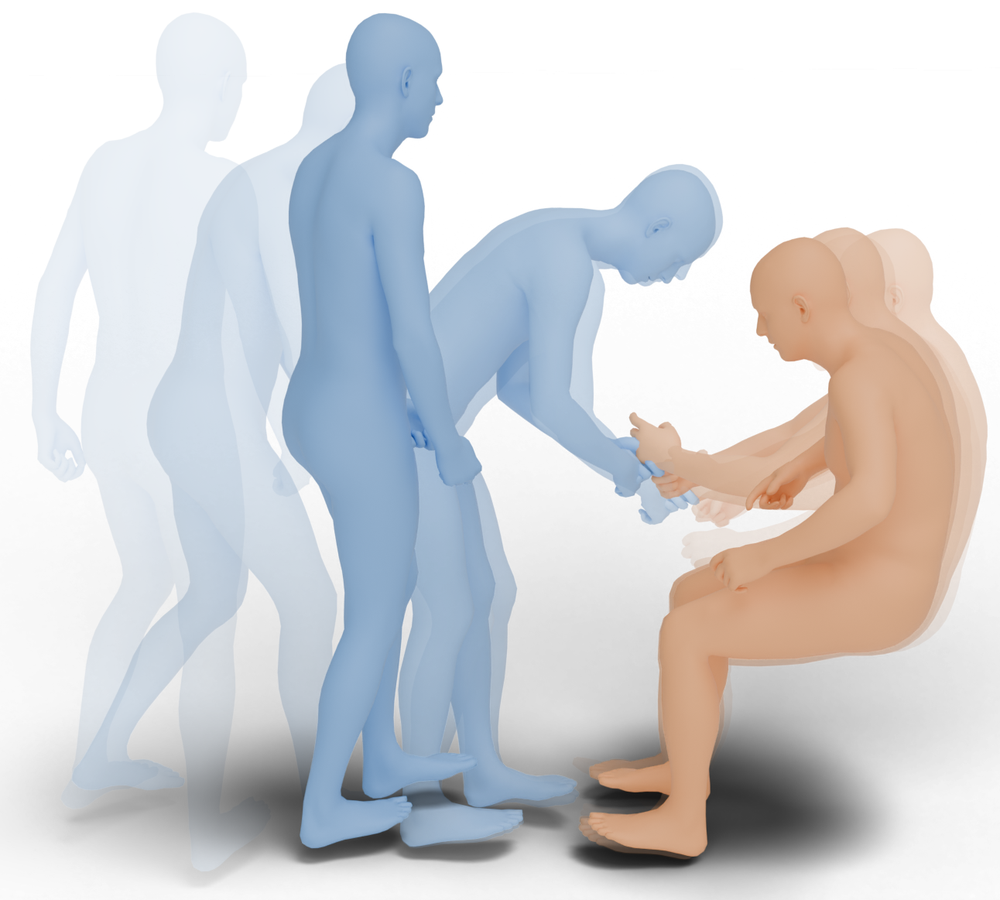}}{%
  \qm{Ref.} & \sqv{Ovr} & \multicolumn{2}{R{3.57in}}{One person walks up to the other person who is sitting, holds his/her hands with both hands, and shakes them up and down.} \\
   & \sqv{P1} & \qA{} is sitting and has their hands held by \qB{} with both hands, and they shake them up and down. & \qA{} walks up to \qB{} who is sitting, holds \qB{}'s hands with both hands, and shakes them up and down. \\
   & \sqv{P2} & \qB{} walks up to \qA{} who is sitting, holds \qA{}'s hands with both hands, and shakes them up and down. & \qB{} is sitting and has their hands held by \qA{} with both hands, and they shake them up and down. \\
   & \sqv{Rel} & \qB{} walks up to \qA{} who is sitting, holds \qA{}'s hands with both hands, and shakes them up and down. & \qA{} walks up to \qB{} who is sitting, holds \qB{}'s hands with both hands, and shakes them up and down. \\[2pt]
  \qm{LLaMMo} & \sqv{Ovr} & \multicolumn{2}{R{3.57in}}{The first person sits on the ground, and the second person helps him up by grabbing his left arm with both hands.} \\
   & \sqv{P1} & \qA{} sits on the ground and \qb{is pulled up by \qB{}}.\qno & \qA{} \qb{sits on the ground} and is pulled up by \qB{}.\qno \\
   & \sqv{P2} & \qB{} \qb{sits on the ground} and is pulled up by \qA{}.\qno & \qB{} sits on the ground and \qb{is pulled up by \qA{}}.\qno \\
   & \sqv{Rel} & \qb{\qA{} grabs \qB{}'s right hand} and pulls them up.\qno & \qA{} walks up to \qB{}, who is sitting on the ground. \qA{} extends both hands and grabs \qB{}'s right arm, \qb{helping \qB{} stand up} from the ground.\qno \\[2pt]
  \rowcolor{qours}
  \qm{InterRole} & \sqv{Ovr} & \multicolumn{2}{R{3.57in}}{One person sits while the other person stands and extends both hands to shake his/her hand.} \\
  \rowcolor{qours}
   & \sqv{P1} & \qA{} \qg{sits on a chair} and extends both hands.\qyes & \qA{} \qg{walks up to \qB{}}, extends both hands, and shakes hands with \qB{}.\qyes \\
  \rowcolor{qours}
   & \sqv{P2} & \qB{} \qg{walks up to \qA{}}, extends both hands, and shakes hands with \qA{}.\qyes & \qB{} \qg{sits on a chair} and extends the hands.\qyes \\
  \rowcolor{qours}
   & \sqv{Rel} & \qg{\qB{} walks up to \qA{}}, extends both hands, and shakes hands with \qA{}.\qyes & \qg{\qA{} walks up to \qB{}}, extends both hands, and shakes hands with \qB{}.\qyes \\
}
\sqrule

\sqcasem{\sqcaselabel{Pull}{G048T001A008R019}}{{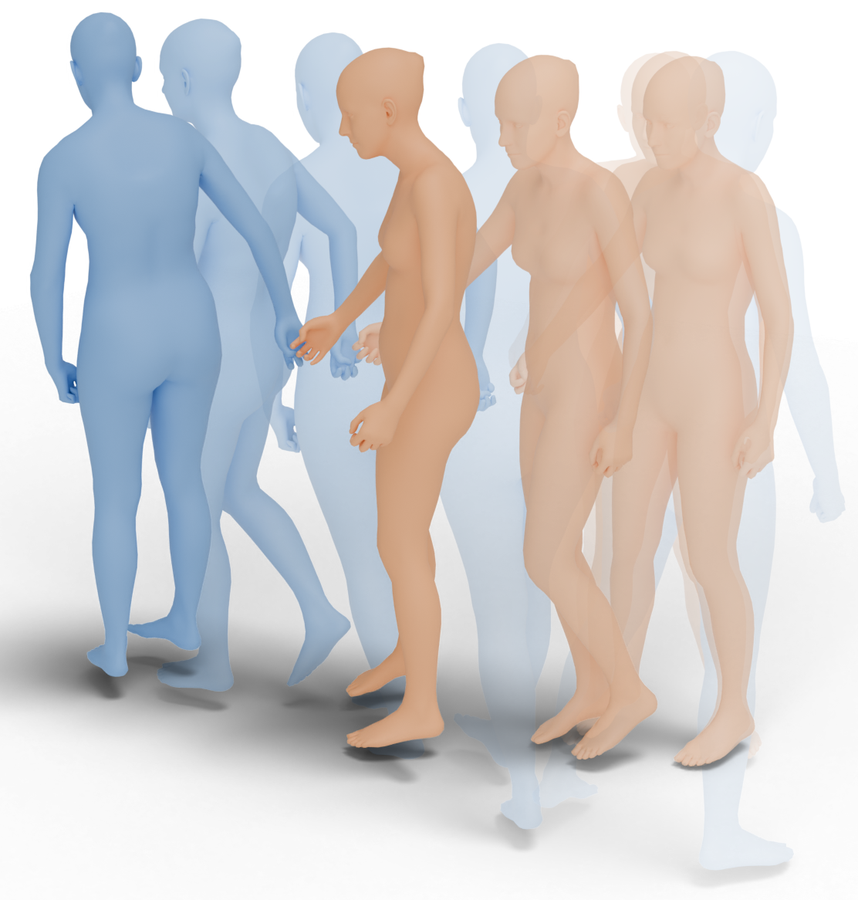}{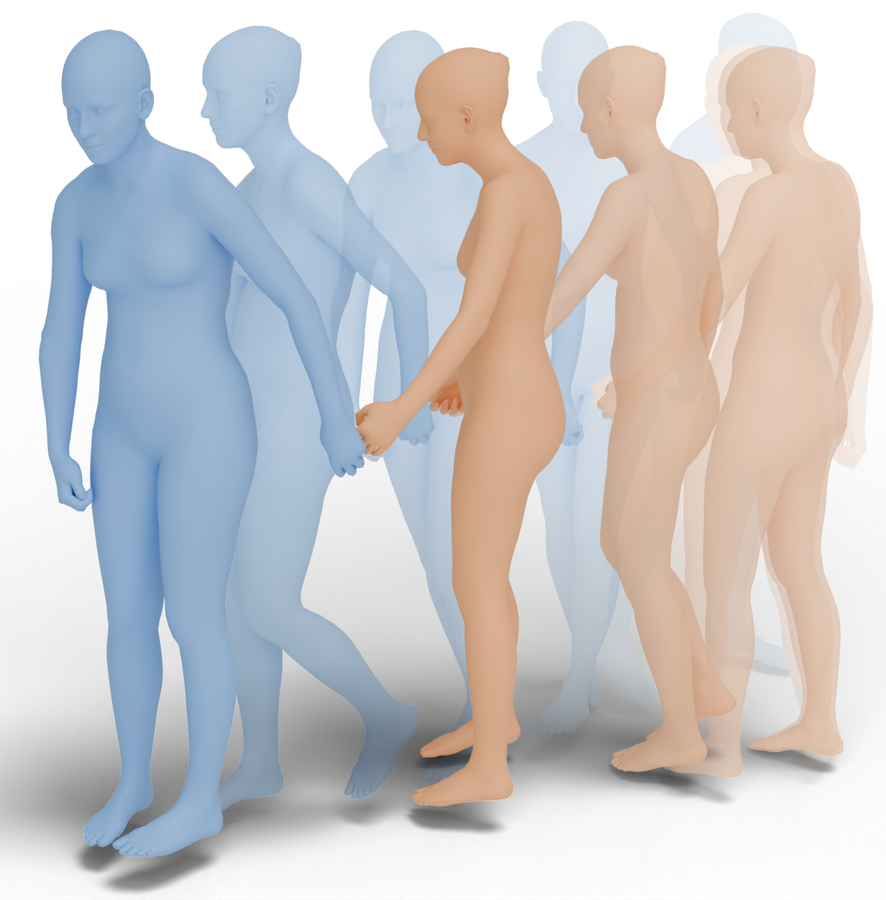}}{{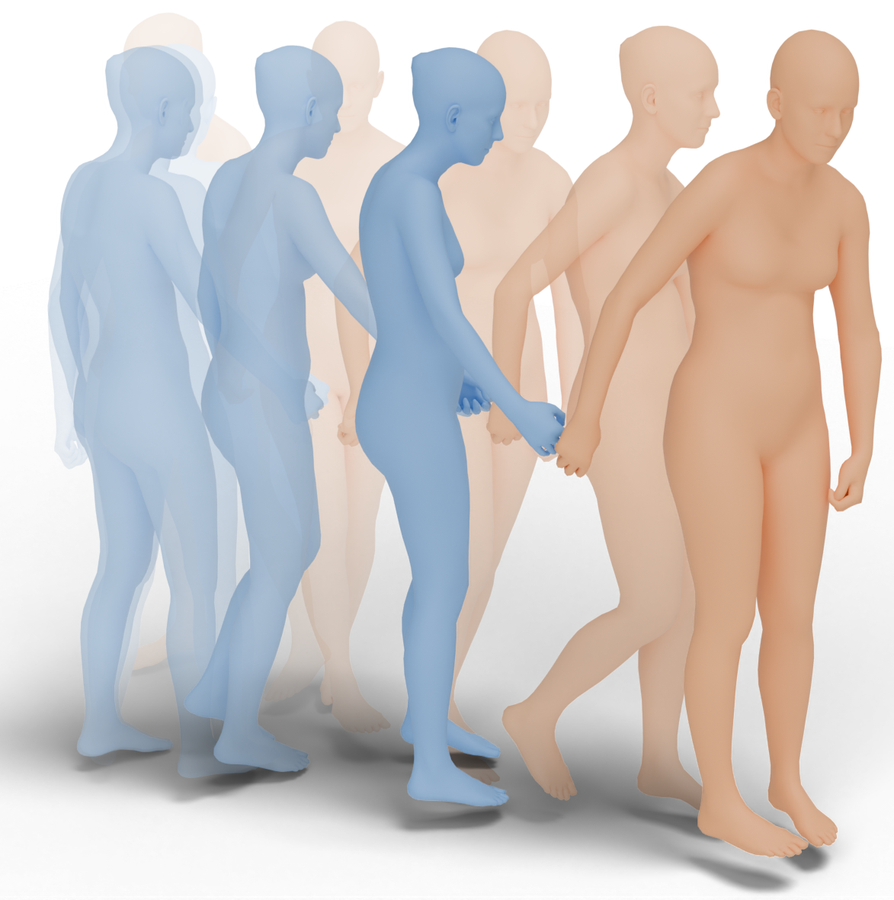}{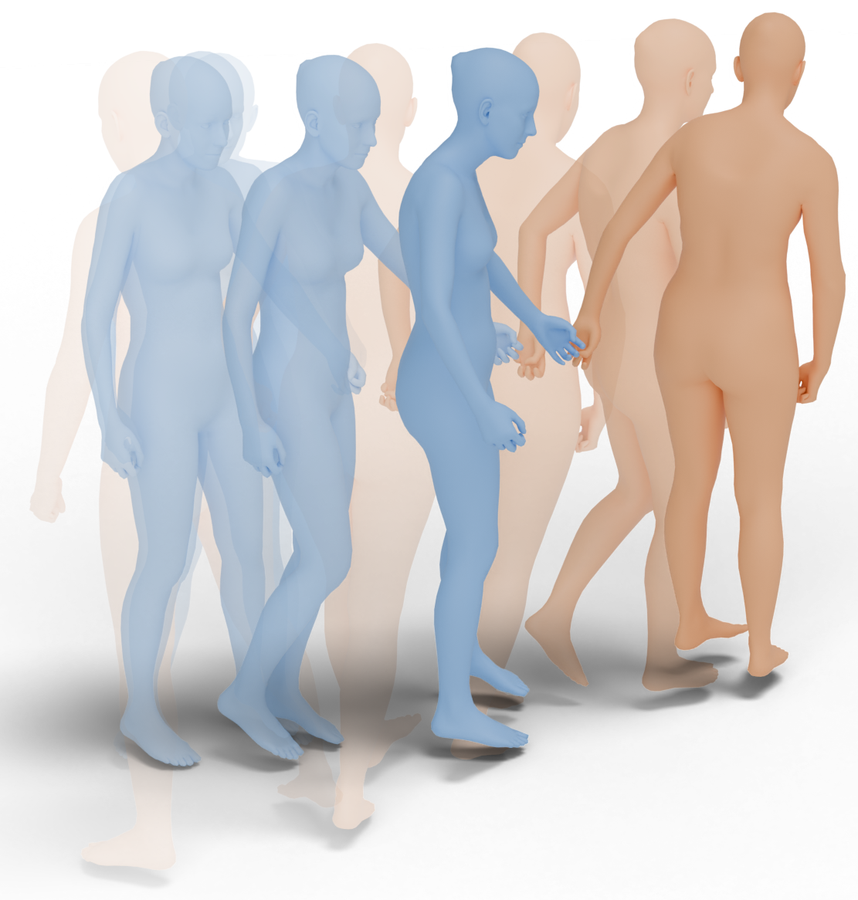}}{%
  \qm{Ref.} & \sqv{Ovr} & \multicolumn{2}{R{3.57in}}{A person walks up to another person, reaches out a hand to hold the other person's hand, and then pulls them forward.} \\
   & \sqv{P1} & \qA{} walks up to \qB{}, reaches out a hand to hold \qB{}'s hand, and pulls \qB{} forward. & \qA{} is pulled forward by \qB{} after \qB{} holds their hand. \\
   & \sqv{P2} & \qB{} is pulled forward by \qA{} after \qA{} holds their hand. & \qB{} walks up to \qA{}, reaches out a hand to hold \qA{}'s hand, and pulls \qA{} forward. \\
   & \sqv{Rel} & \qA{} walks up to \qB{}, holds their hand, and pulls them forward. & \qB{} walks up to \qA{}, holds their hand, and pulls them forward. \\[2pt]
  \qm{LLaMMo} & \sqv{Ovr} & \multicolumn{2}{R{3.57in}}{The first person walks towards the second person, and the second person raises his/her right hand to block him/her.} \\
   & \sqv{P1} & \qA{} stands facing \qB{}, takes a step forward with the left leg, and extends the right hand to \qb{pat \qB{}'s back}.\qno & \qA{} stands facing \qB{}, takes a step forward with the left leg, and extends the right hand to \qb{push \qB{}'s left shoulder}.\qno \\
   & \sqv{P2} & \qB{} stands facing \qA{} and \qb{is pushed on the right shoulder} by \qA{}.\qno & \qB{} stands facing \qA{} and \qb{is pushed on the right shoulder} by \qA{}.\qno \\
   & \sqv{Rel} & \qb{\qB{} pushes \qA{}'s right shoulder} with the left hand, causing \qA{} to take a few steps back.\qno & \qb{\qB{} pushes \qA{}'s right shoulder} with the left hand, causing \qA{} to take a few steps back.\qno \\[2pt]
  \rowcolor{qours}
  \qm{InterRole} & \sqv{Ovr} & \multicolumn{2}{R{3.57in}}{One person walks up to another and grabs his/her right arm with both hands, pulling him/her forward.} \\
  \rowcolor{qours}
   & \sqv{P1} & \qA{} walks up to \qB{} and \qg{extends the right hand}.\qyes & \qA{} \qg{is pulled forward by \qB{}}'s right hand.\qyes \\
  \rowcolor{qours}
   & \sqv{P2} & \qB{} \qg{is pulled forward by \qA{}}'s right hand.\qyes & \qB{} stands facing \qA{}, extends the right hand to grasp \qA{}'s arm, and \qg{pulls \qA{} forward} a few steps.\qyes \\
  \rowcolor{qours}
   & \sqv{Rel} & \qg{\qA{} grabs \qB{}'s right hand} and pulls it towards themselves.\qyes & \qg{\qB{} grabs \qA{}'s right hand} and pulls it towards themselves.\qyes \\
}
\caption{%
Additional qualitative results on Inter-X (Qwen2.5-1.5B; part 1/2): further cases in the format of Fig.~\ref{fig:qualitative}.
\emph{Left of each pair:} SMPL-X renderings (\textcolor{pone}{blue}: slot~1, \textcolor{ptwo}{orange}: slot~2; fainter silhouettes are earlier frames; the swapped input is viewed from the opposite side); below each, the same frames seen from the opposite side and mirrored (left and right exchanged); left/right in the texts refer to the upper render. \emph{Right:} reference, LLaMMo and InterRole descriptions for the original input \textbf{(a)} and the input with exchanged streams \textbf{(b)}, verbatim; $\Ovr$ does not depend on the slot order and is shown once. \textcolor{qokc}{\ding{51}}/\textcolor{qbadc}{\ding{55}}: the described action (for $\Rel$, its direction) is attributed to the correct person.
}
\label{fig:qualitative_supp}
\end{figure*}

\begin{figure*}[p]
\centering
\fontsize{6.0}{6.7}\selectfont
\sqheader
\sqcasem{\sqcaselabel{Touch head}{G052T006A032R013}}{{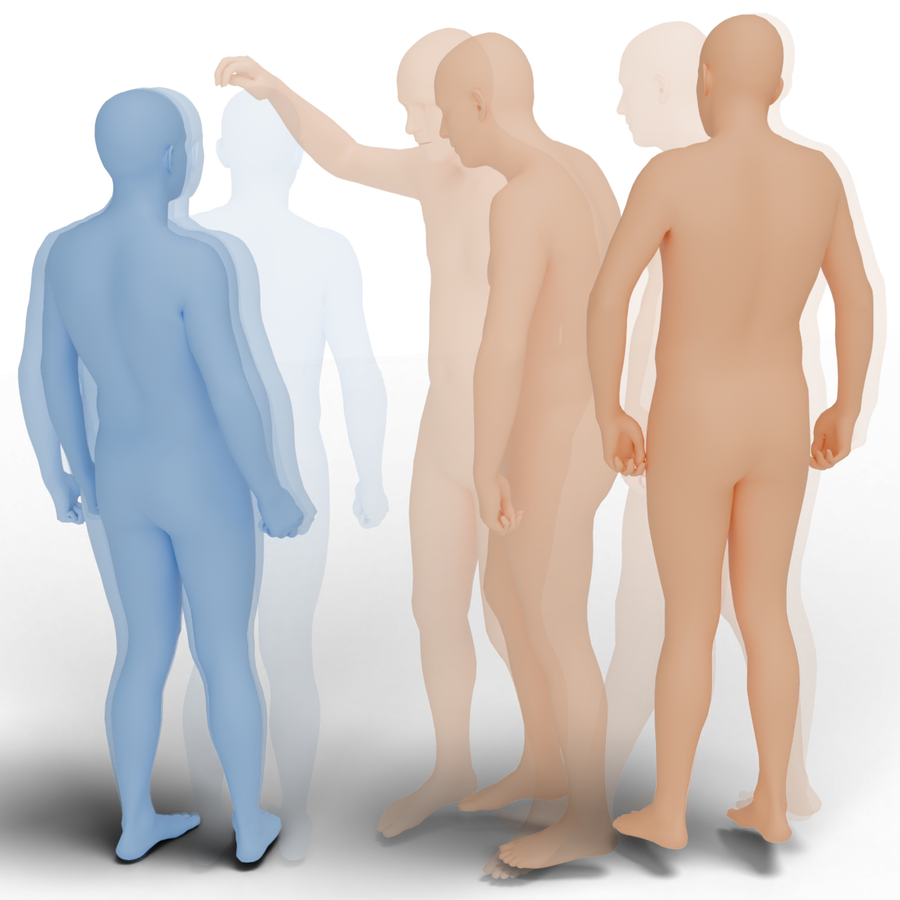}{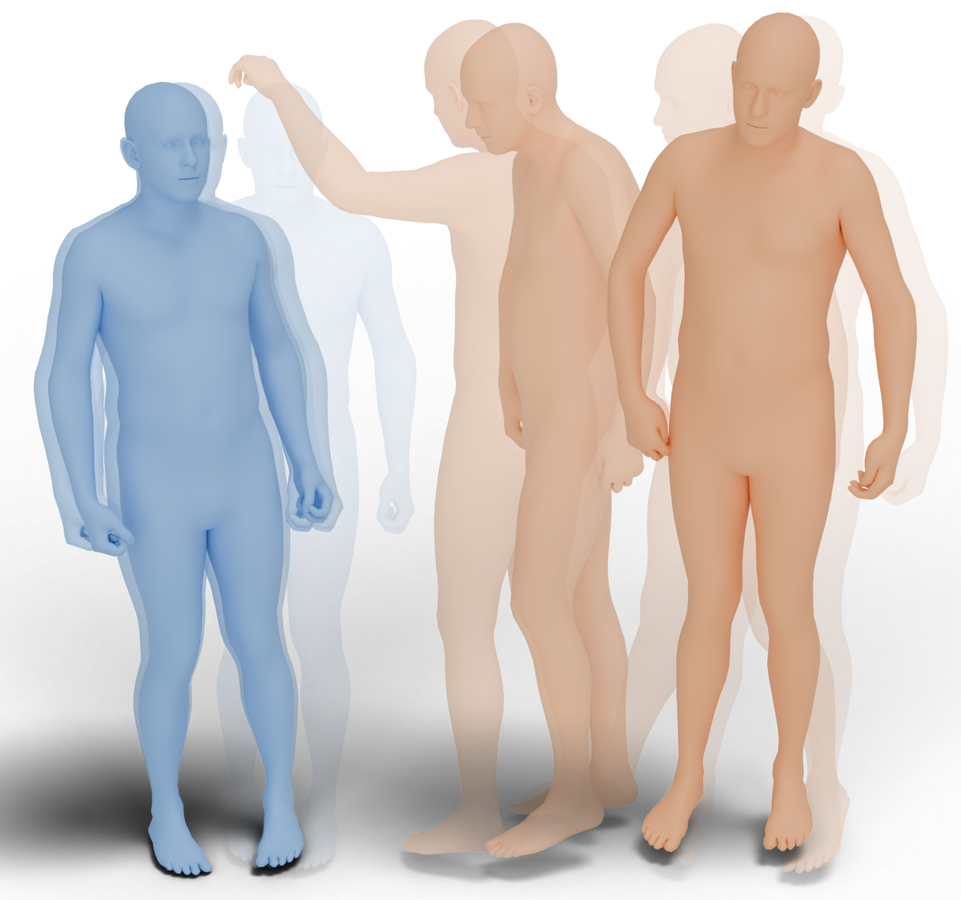}}{{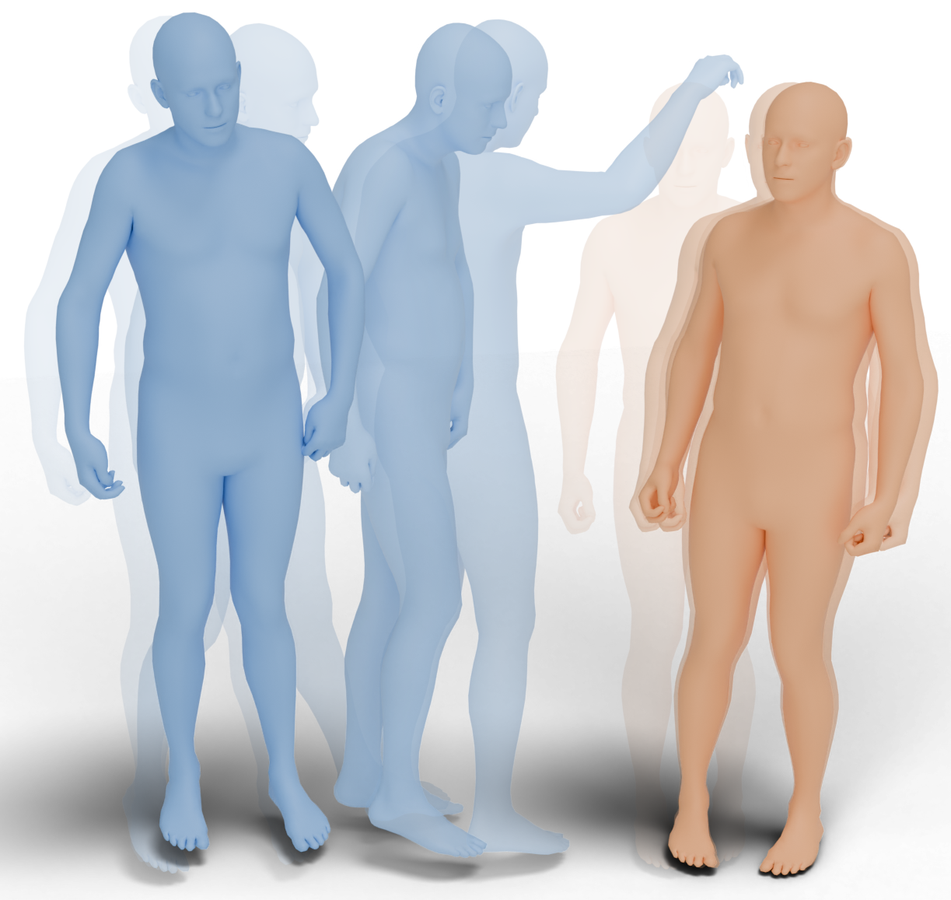}{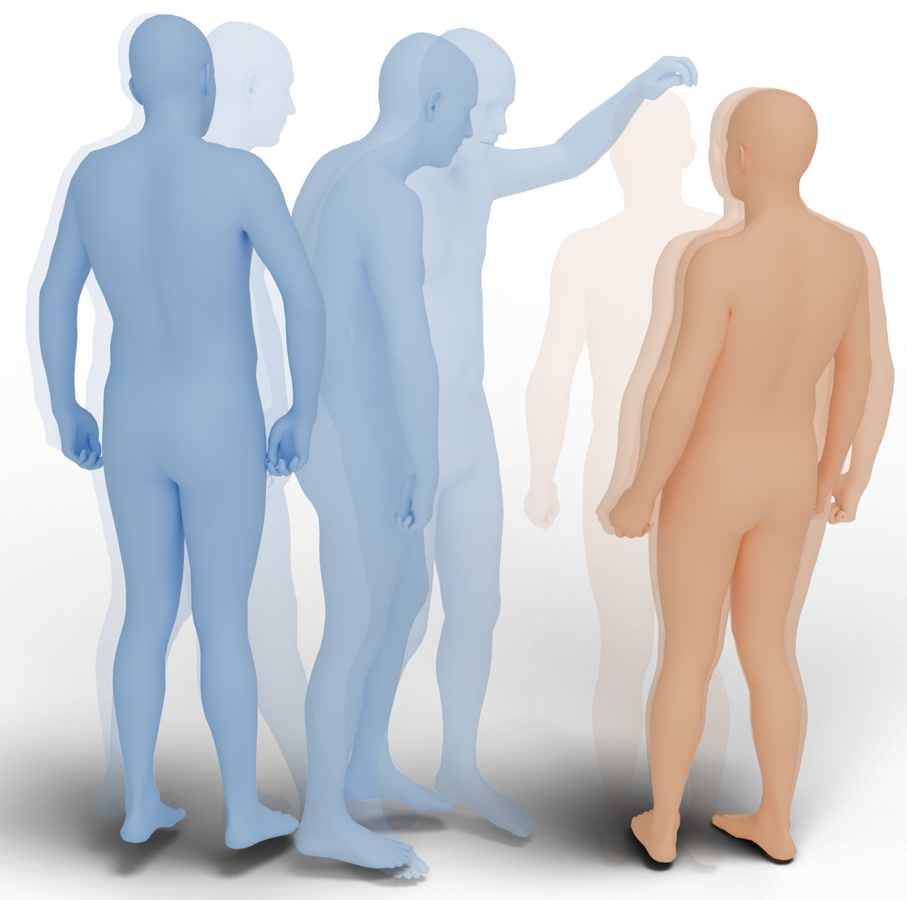}}{%
  \qm{Ref.} & \sqv{Ovr} & \multicolumn{2}{R{3.57in}}{Both people stand. One person walks towards the other, extends his/her right hand, and touches the other person's head. The other person nods slightly.} \\
   & \sqv{P1} & \qA{} nods slightly after being touched on the head. & \qA{} walks towards \qB{}, extends the right hand, and touches \qB{}'s head. \\
   & \sqv{P2} & \qB{} walks towards \qA{}, extends the right hand, and touches \qA{}'s head. & \qB{} nods slightly after being touched on the head. \\
   & \sqv{Rel} & \qB{} walks towards \qA{}, extends the right hand, and touches \qA{}'s head, causing \qA{} to nod slightly. & \qA{} walks towards \qB{}, extends the right hand, and touches \qB{}'s head, causing \qB{} to nod slightly. \\[2pt]
  \qm{LLaMMo} & \sqv{Ovr} & \multicolumn{2}{R{3.57in}}{One person walks up to the other and pats him/her on the back with his/her right hand.} \\
   & \sqv{P1} & \qA{} stands facing \qB{}, raises the right hand to \qb{point at \qB{}}.\qno & \qA{} stands facing \qB{}, raises the right hand to \qb{point at \qB{}}.\qno \\
   & \sqv{P2} & \qB{} stands facing \qA{}, raises the right hand to \qb{point at \qA{}}.\qno & \qB{} stands facing \qA{}, raises the right hand to \qb{point at \qA{}}.\qno \\
   & \sqv{Rel} & \qb{\qA{} pats \qB{}'s back}, and \qB{} turns around to look at \qA{}.\qno & \qb{\qA{} pats \qB{}'s back}, and \qB{} turns around.\qno \\[2pt]
  \rowcolor{qours}
  \qm{InterRole} & \sqv{Ovr} & \multicolumn{2}{R{3.57in}}{One person walks up to the other and touches his/her head with one hand.} \\
  \rowcolor{qours}
   & \sqv{P1} & \qA{} stands facing \qB{} and \qg{is patted on the head by \qB{}}.\qyes & \qA{} stands facing \qB{} and raises the right hand to \qg{touch \qB{}'s head}.\qyes \\
  \rowcolor{qours}
   & \sqv{P2} & \qB{} stands facing \qA{} and raises the right hand to \qg{touch \qA{}'s head}.\qyes & \qB{} \qg{stands facing \qA{}}.\qyes \\
  \rowcolor{qours}
   & \sqv{Rel} & \qB{} walks up to \qA{} and \qg{touches \qA{}'s head} with the right hand.\qyes & \qA{} walks towards \qB{}, raises the right hand to \qg{touch \qB{}'s head}.\qyes \\
}
\sqrule

\sqcasem{\sqcaselabel{Kick}{G003T001A005R000}}{{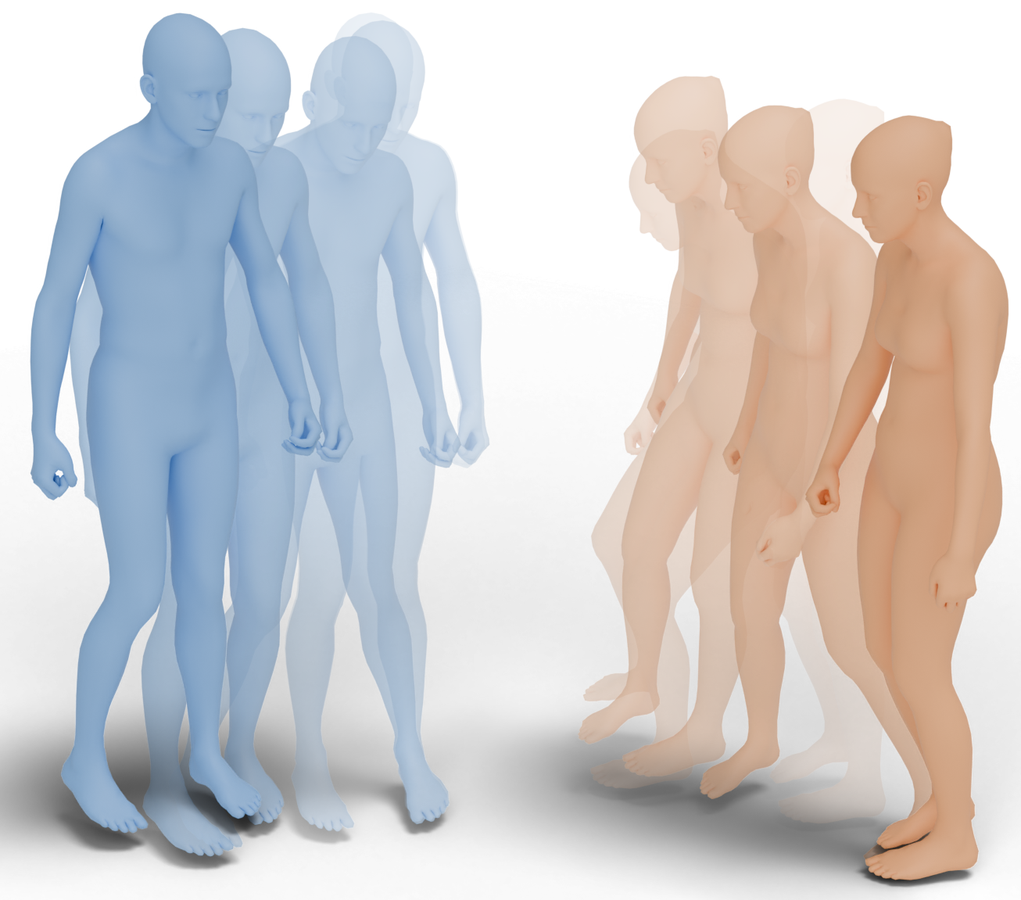}{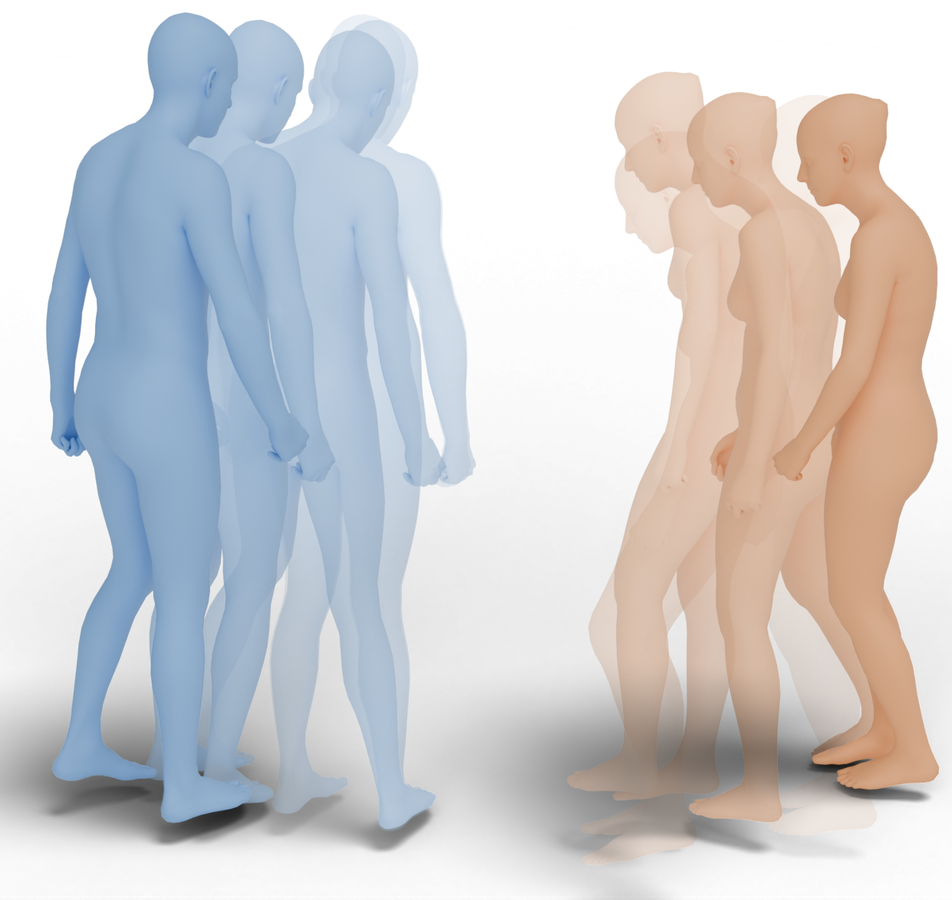}}{{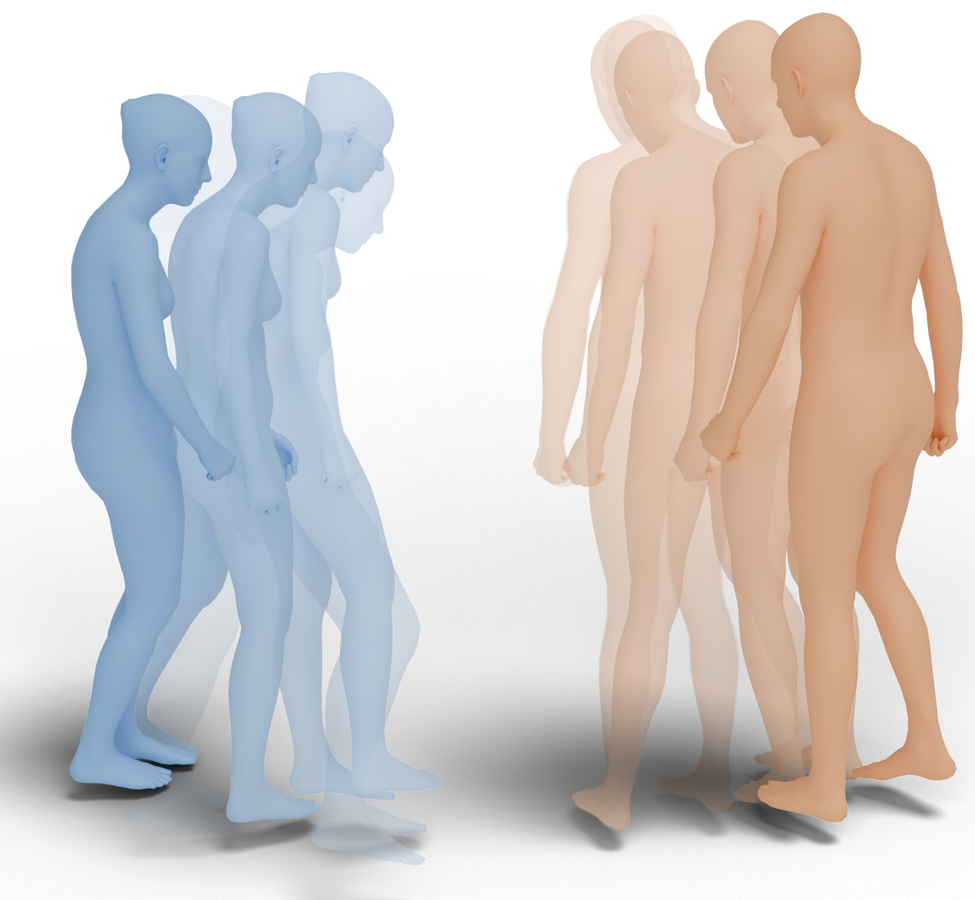}{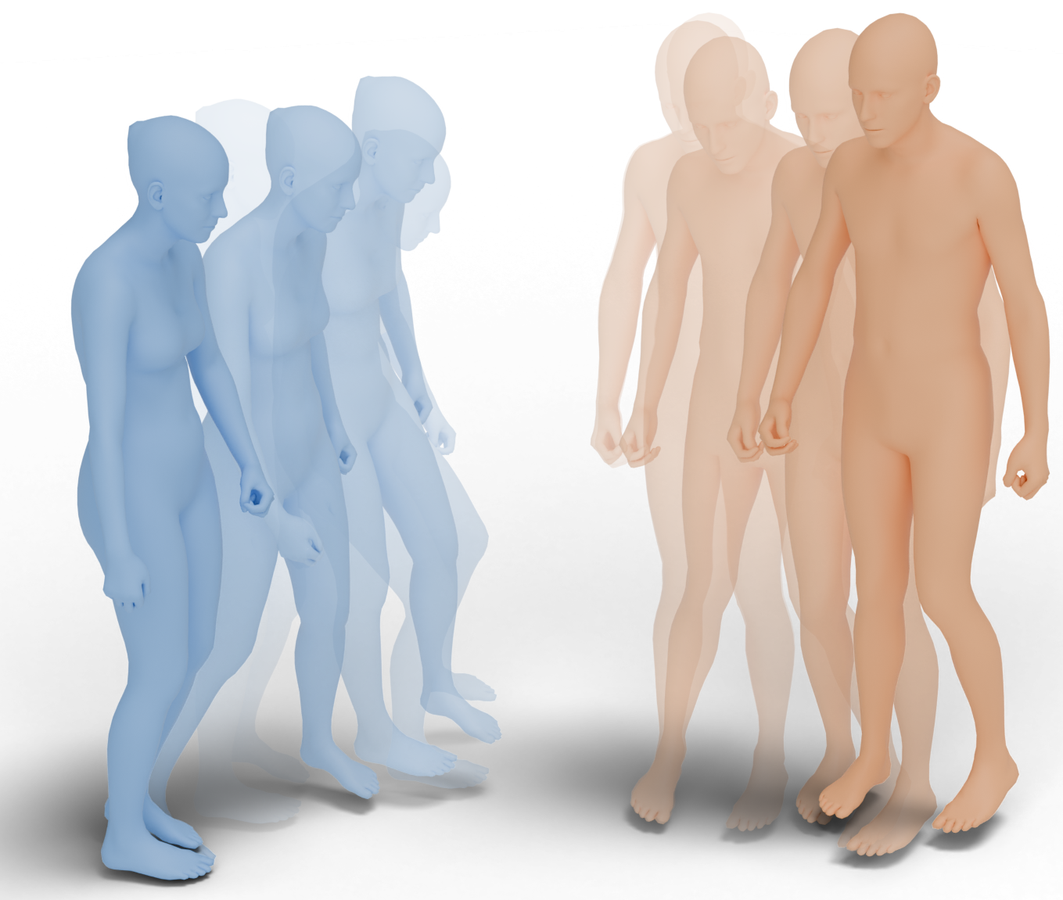}}{%
  \qm{Ref.} & \sqv{Ovr} & \multicolumn{2}{R{3.57in}}{The first person stands in front of the second person, positioned to the left, and faces him. He gently kicks the other person's left calf with his right foot.} \\
   & \sqv{P1} & \qA{} is kicked on the left calf by \qB{}'s right foot. & \qA{} stands in front of \qB{}, positioned to the left, and faces \qB{}. \qA{} gently kicks \qB{}'s left calf with his right foot. \\
   & \sqv{P2} & \qB{} stands in front of \qA{}, positioned to the left, and faces \qA{}. \qB{} gently kicks \qA{}'s left calf with his right foot. & \qB{} is kicked on the left calf by \qA{}'s right foot. \\
   & \sqv{Rel} & \qB{} stands in front of \qA{}, positioned to the left, and faces \qA{}. \qB{} gently kicks \qA{}'s left calf with his right foot. & \qA{} stands in front of \qB{}, positioned to the left, and faces \qB{}. \qA{} gently kicks \qB{}'s left calf with his right foot. \\[2pt]
  \qm{LLaMMo} & \sqv{Ovr} & \multicolumn{2}{R{3.57in}}{Two people stand facing each other. One person steps forward and kicks the other person's left leg with his/her right foot, causing him/her to take a few steps back.} \\
   & \sqv{P1} & \qA{} stands facing \qB{}, takes a step forward with the right foot, and \qb{kicks \qB{}'s left leg}.\qno & \qA{} stands facing \qB{}, takes a step forward with the right foot, and \qb{steps on \qB{}'s left foot}.\qno \\
   & \sqv{P2} & \qB{} stands facing \qA{} and steps forward, lifting the right foot to \qb{step on \qA{}'s left foot}.\qno & \qB{} stands facing \qA{} and steps forward, lifting the right foot to \qb{step on \qA{}'s left foot}.\qno \\
   & \sqv{Rel} & \qg{\qB{} kicks \qA{}'s left leg}, causing \qA{} to step back.\qyes & \qb{\qB{} kicks \qA{}'s left leg}, causing \qA{} to step back.\qno \\[2pt]
  \rowcolor{qours}
  \qm{InterRole} & \sqv{Ovr} & \multicolumn{2}{R{3.57in}}{One person kicks the other in the left leg with his/her right foot, causing him/her to take a few steps back.} \\
  \rowcolor{qours}
   & \sqv{P1} & \qA{} stands facing \qB{} and \qg{is kicked in the left calf by \qB{}}'s right foot.\qyes & \qA{} stands facing \qB{} and \qg{kicks \qB{}'s left calf} with the right foot.\qyes \\
  \rowcolor{qours}
   & \sqv{P2} & \qB{} stands facing \qA{} and \qg{kicks \qA{}'s left leg} with the right foot.\qyes & \qB{} takes a few steps back \qg{after being kicked}.\qyes \\
  \rowcolor{qours}
   & \sqv{Rel} & \qg{\qB{} kicks \qA{}'s left leg} with the right foot, causing \qA{} to take a few steps back.\qyes & \qg{\qA{} kicks \qB{}'s left leg} with the right foot, causing \qB{} to take a few steps back.\qyes \\
}
\sqrule

\sqcasem{\sqcaselabel{Massaging leg}{G002T005A027R014}}{{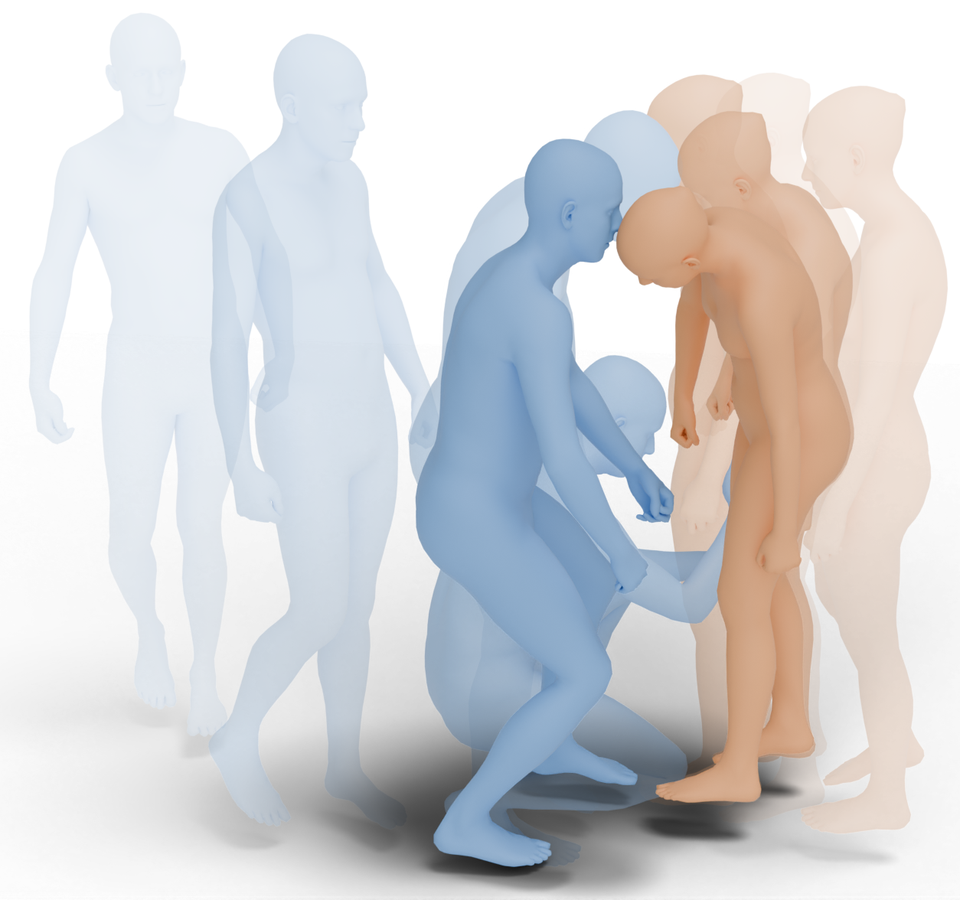}{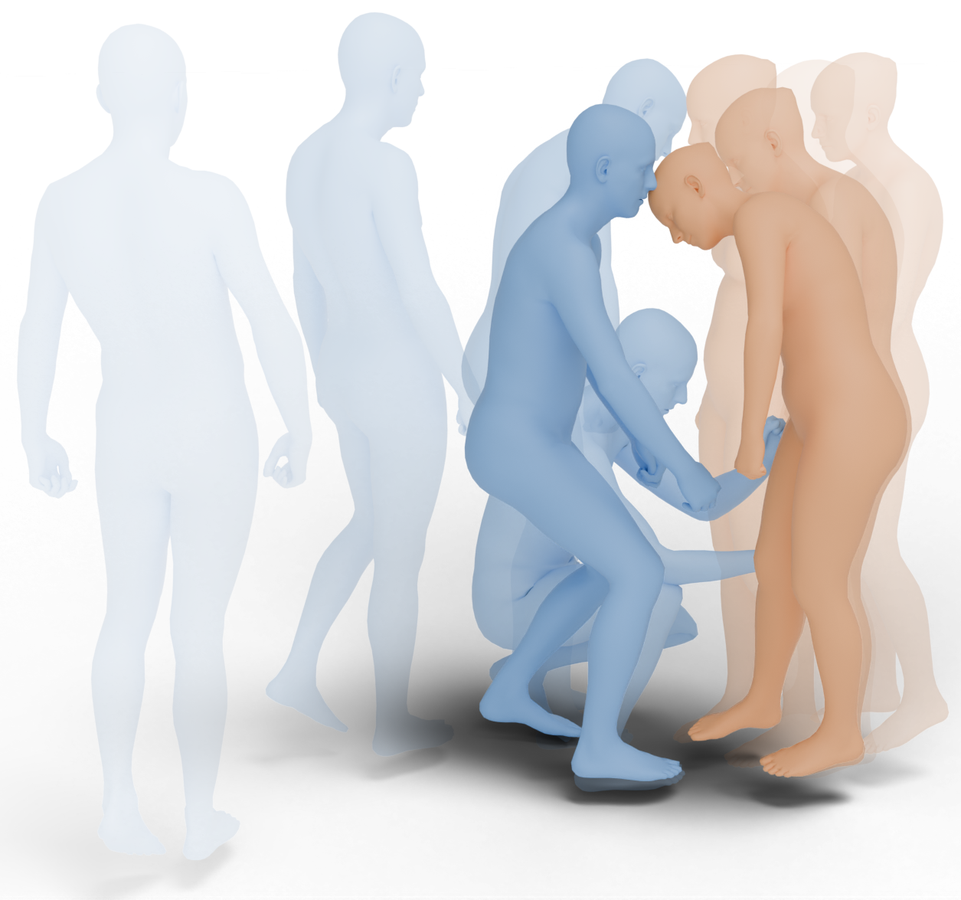}}{{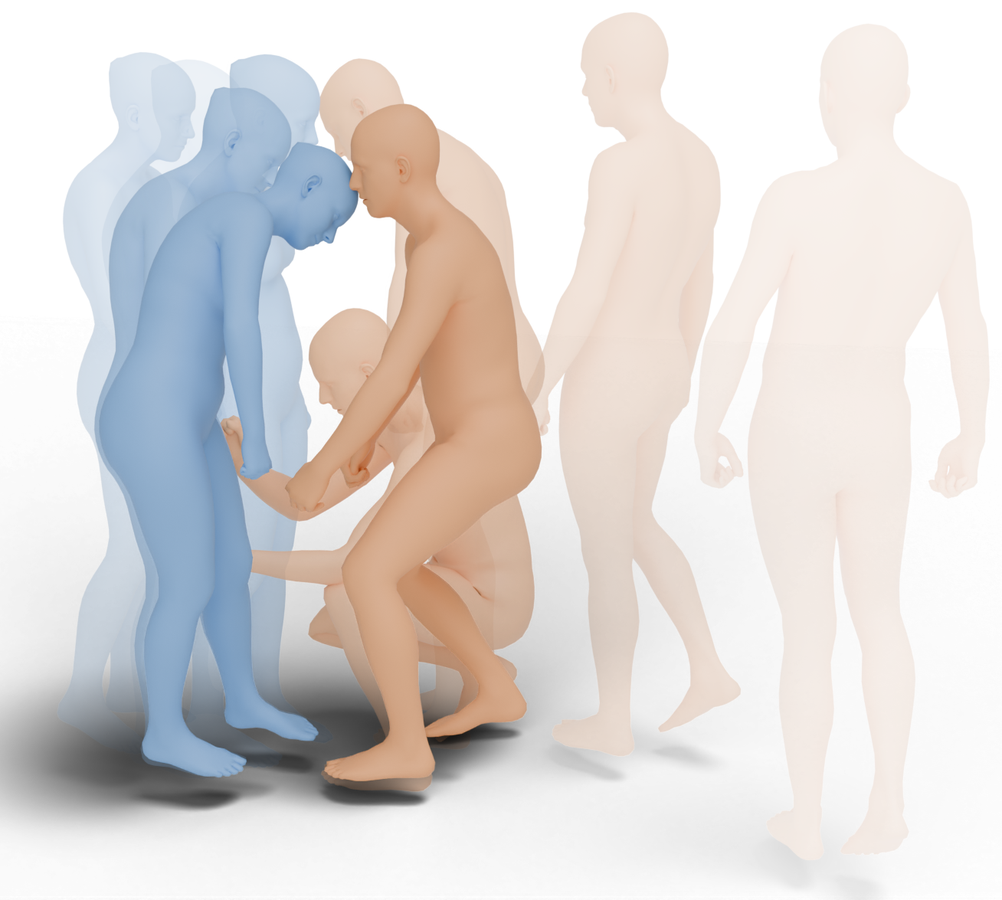}{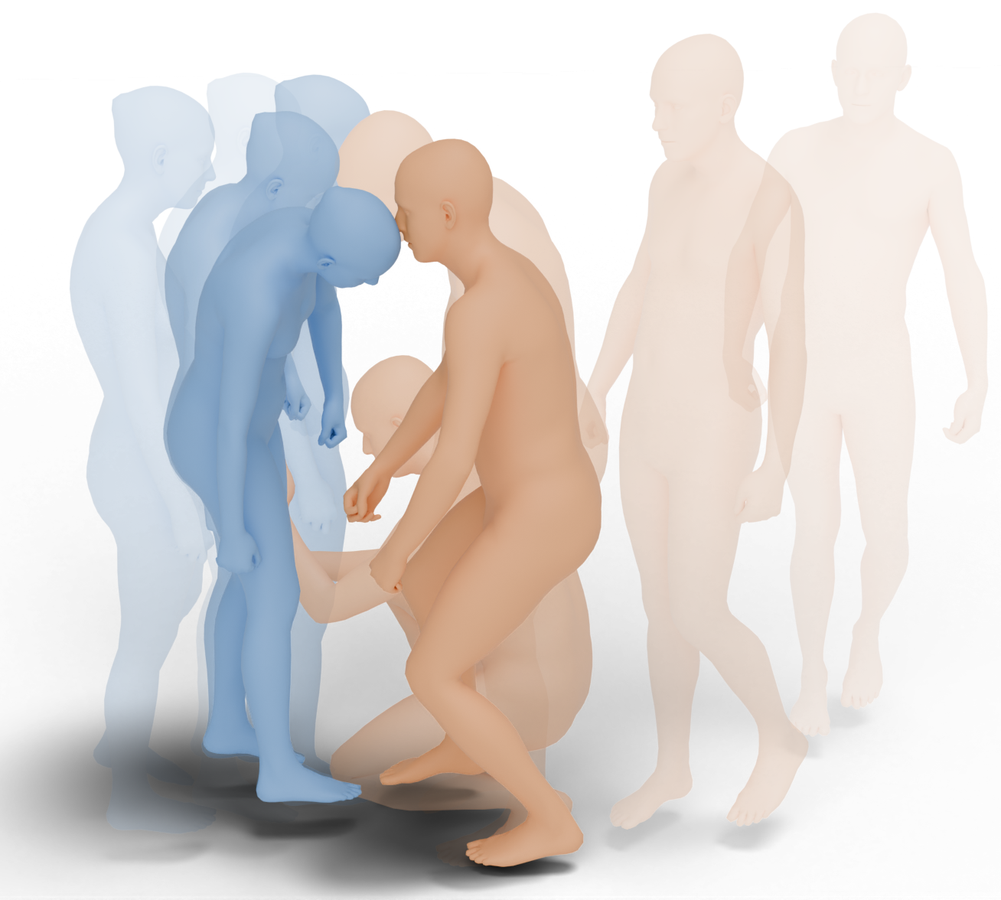}}{%
  \qm{Ref.} & \sqv{Ovr} & \multicolumn{2}{R{3.57in}}{The first person stands in front of the second person, while the second person crouches down and gently massages his/her left leg by softly punching it with both hands.} \\
   & \sqv{P1} & \qA{} crouches down and softly punches \qB{}'s left leg with both hands. & \qA{} stands in front of \qB{}. \\
   & \sqv{P2} & \qB{} stands in front of \qA{}. & \qB{} crouches down and softly punches \qA{}'s left leg with both hands. \\
   & \sqv{Rel} & \qA{} crouches down and softly punches \qB{}'s left leg with both hands. & \qB{} crouches down and softly punches \qA{}'s left leg with both hands. \\[2pt]
  \qm{LLaMMo} & \sqv{Ovr} & \multicolumn{2}{R{3.57in}}{One person sits while the other person stands. The standing individual extends his/her right hand to lightly touch the left leg of the sitting person, who then raises both hands and places them on his/her own legs.} \\
   & \sqv{P1} & \qA{} \qb{sits down}.\qno & \qA{} \qb{sits down}.\qno \\
   & \sqv{P2} & \qB{} \qb{sits down}.\qno & \qB{} \qb{sits down}.\qno \\
   & \sqv{Rel} & \qb{\qB{} massages \qA{}'s legs} while standing behind them.\qno & \qg{\qB{} massages \qA{}'s legs} while standing behind them.\qyes \\[2pt]
  \rowcolor{qours}
  \qm{InterRole} & \sqv{Ovr} & \multicolumn{2}{R{3.57in}}{One person squats down and massages the other person's legs with both hands.} \\
  \rowcolor{qours}
   & \sqv{P1} & \qA{} crouches down and \qg{massages \qB{}'s legs} with both hands.\qyes & \qA{} \qg{is massaged on the right leg by \qB{}}.\qyes \\
  \rowcolor{qours}
   & \sqv{P2} & \qB{} \qg{is massaged on the right leg by \qA{}}.\qyes & \qB{} crouches down and \qg{massages \qA{}'s legs} with both hands.\qyes \\
  \rowcolor{qours}
   & \sqv{Rel} & \qg{\qA{} massages \qB{}'s right leg} while squatting down.\qyes & \qB{} squats down and \qg{massages \qA{}'s legs} by hammering them with both hands.\qyes \\
}
\caption{%
Additional qualitative results on Inter-X (part 2/2, continued). Layout and marks as in Fig.~\ref{fig:qualitative_supp}.
}
\end{figure*}

%% file: figures/sq_macros.tex
\input{figures/qual_base_macros}
\ifdefined\sqcase\else
\newcommand{\sqv}[1]{{\color{qlgray}\scriptsize #1}}
\newcommand{\sqimg}[1]{\makebox[1.10in]{\includegraphics[width=1.10in,height=1.08in,keepaspectratio]{#1}}}
\newcommand{\sqcase}[4]{%
  \noindent
  \begin{minipage}[c]{0.20in}\centering
    \rotatebox[origin=c]{90}{#1}
  \end{minipage}\hfill
  \begin{minipage}[c]{2.26in}\centering
    \sqimg{#2}\hspace{4pt}\sqimg{#3}
  \end{minipage}\hfill
  \begin{minipage}[c]{4.24in}
    \renewcommand{\arraystretch}{0.92}\setlength{\tabcolsep}{2pt}
    \begin{tabular}{@{}p{0.40in}p{0.15in}R{1.76in}R{1.76in}@{}}
      #4
    \end{tabular}
  \end{minipage}\par
}
\newcommand{\sqimgm}[2]{\makebox[1.17in]{\parbox{1.17in}{\centering
  \includegraphics[width=1.17in,height=0.98in,keepaspectratio]{#1}\\[1pt]
  \includegraphics[width=1.17in,height=0.98in,keepaspectratio]{#2}}}}
\newcommand{\sqcasem}[4]{%
  \noindent
  \begin{minipage}[c]{0.20in}\centering
    \rotatebox[origin=c]{90}{#1}
  \end{minipage}\hfill
  \begin{minipage}[c]{2.42in}\centering
    \sqimgm#2\hspace{4pt}\sqimgm#3
  \end{minipage}\hfill
  \begin{minipage}[c]{4.24in}
    \renewcommand{\arraystretch}{0.92}\setlength{\tabcolsep}{2pt}
    \begin{tabular}{@{}p{0.40in}p{0.15in}R{1.76in}R{1.76in}@{}}
      #4
    \end{tabular}
  \end{minipage}\par
}
\newcommand{\sqcaselabel}[2]{\shortstack{\textbf{\small #1}\\[1pt]{\color{qlgray}\tiny #2}}}
\newcommand{\sqrule}{\par\vspace{2pt}{\color{black!22}\hrule height 0.4pt}\vspace{3pt}}
\newcommand{\sqheader}{%
  \noindent
  \begin{minipage}[c]{0.20in}\ \end{minipage}\hfill
  \begin{minipage}[c]{2.26in}\centering
    \makebox[1.10in]{\textbf{(a)}~Original}\hspace{4pt}\makebox[1.10in]{\textbf{(b)}~Swapped}
  \end{minipage}\hfill
  \begin{minipage}[c]{4.24in}
    \setlength{\tabcolsep}{2pt}
    \begin{tabular}{@{}p{0.40in}p{0.15in}R{1.76in}R{1.76in}@{}}
      & & \textbf{(a)}~Original input & \textbf{(b)}~Participants swapped \\
    \end{tabular}
  \end{minipage}\par
  \sqrule}
\fi

%% file: figures/fig_crossdataset_supp.tex

\input{figures/sq_macros}
\begin{figure*}[p]
\centering
\fontsize{6.6}{7.7}\selectfont
\sqheader
\sqcase{\sqcaselabel{Leg strike}{InterHuman 631}}{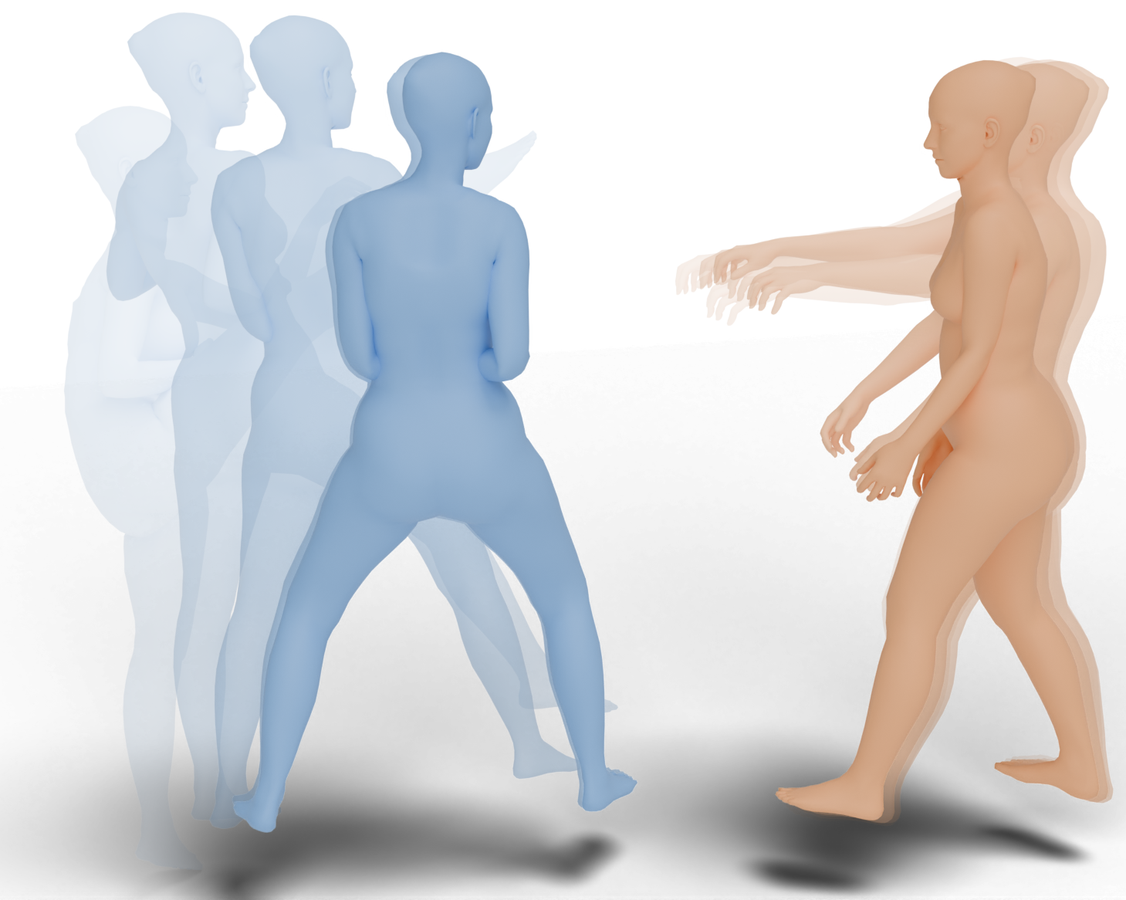}{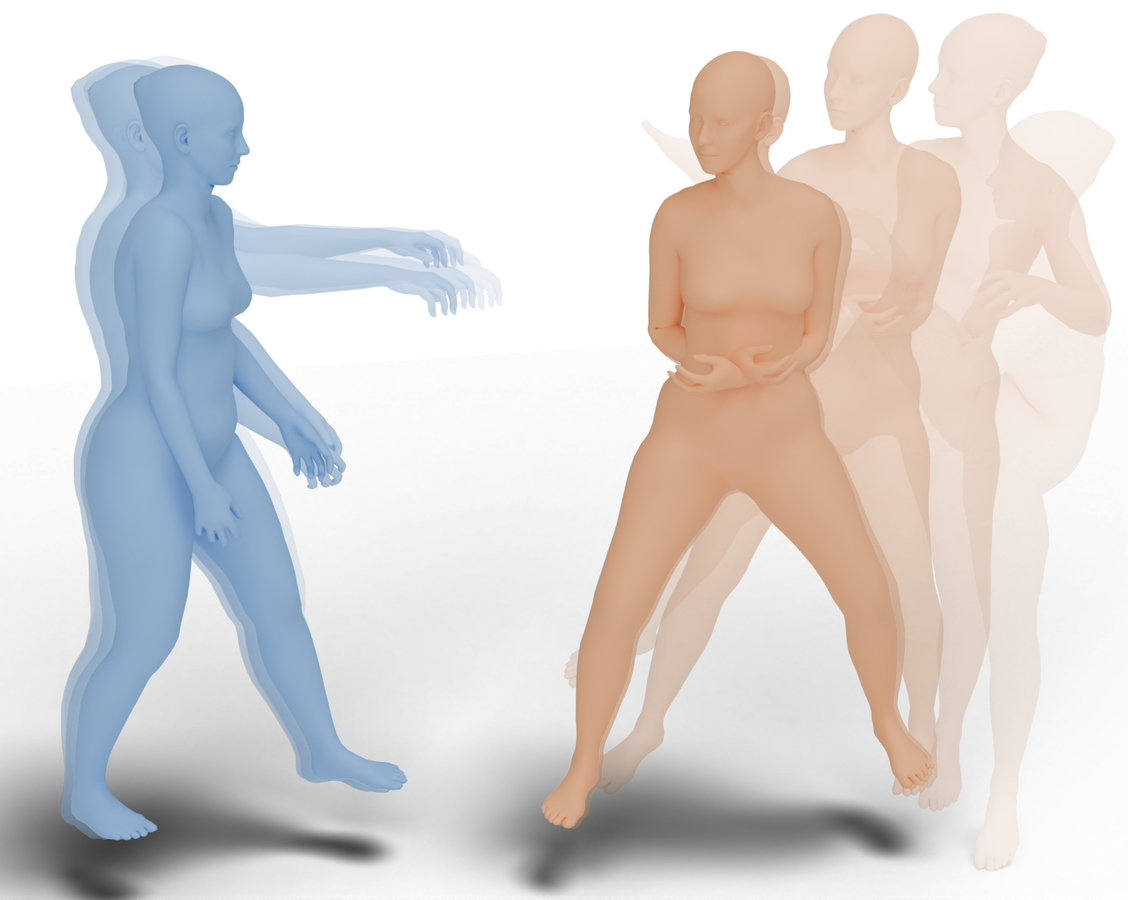}{%
  \qm{Ref.} & \sqv{GT} & \multicolumn{2}{R{3.44in}}{One performs a right-leg strike, while the other holds out a magazine in front of her with both arms.} \\[2pt]
  \qm{LLaMMo} & \sqv{Ov.} & One person stands up and raises his/her right leg to the left, while the other person imitates him/her by raising his/her left leg to the left. & \sqv{(not generated)} \\
   & \sqv{P1} & \qA{} stands facing \qB{}, \qg{raises the right leg to kick}, and then raises both legs.\qyes & \qA{} \qb{turns around and waves}.\qno \\
   & \sqv{P2} & \qB{} stands facing \qA{}, \qb{raises the right leg to kick}, and then raises both legs.\qno & \qB{} \qb{turns around and waves}.\qno \\
   & \sqv{Rel} & \qA{} and \qB{} face each other, with \qb{\qA{} performing a gesture of approval} and \qB{} reciprocating by raising the right hand.\qno & \qb{\qB{} waves to \qA{}}, who turns around and raises his/her right hand.\qno \\[2pt]
  \rowcolor{qours}
  \qm{InterRole} & \sqv{Ov.} & The first person raises his/her right leg and kicks the second person, while the second person extends both hands in front of his/her chest. & \sqv{(not generated)} \\
  \rowcolor{qours}
   & \sqv{P1} & \qA{} \qg{raises the right leg and kicks} \qB{}'s left thigh.\qyes & \qA{} \qg{raises both hands and waves}.\qyes \\
  \rowcolor{qours}
   & \sqv{P2} & \qB{} \qg{raises both hands} and waves them in front of his/her chest.\qyes & \qB{} \qg{raises the right leg to kick} \qA{}.\qyes \\
  \rowcolor{qours}
   & \sqv{Rel} & \qg{\qA{} kicks} \qB{}'s left leg, and \qB{} raises both hands to block.\qyes & \qg{\qB{} kicks} \qA{}'s left leg, causing \qA{} to turn around.\qyes \\
}
\sqrule

\sqcase{\sqcaselabel{Kick}{InterHuman 4653}}{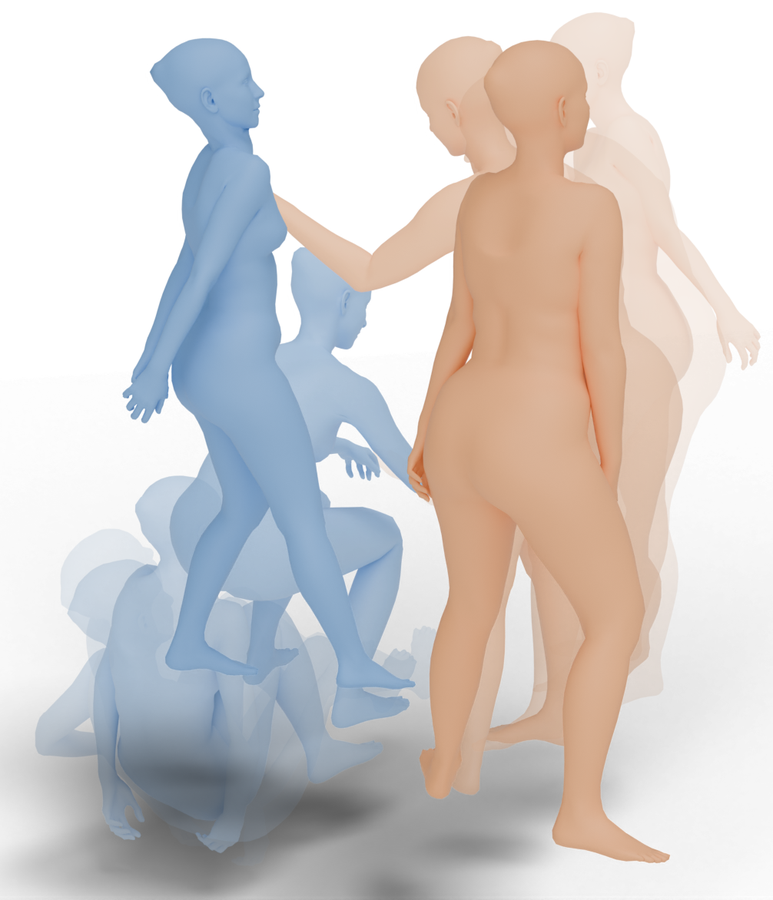}{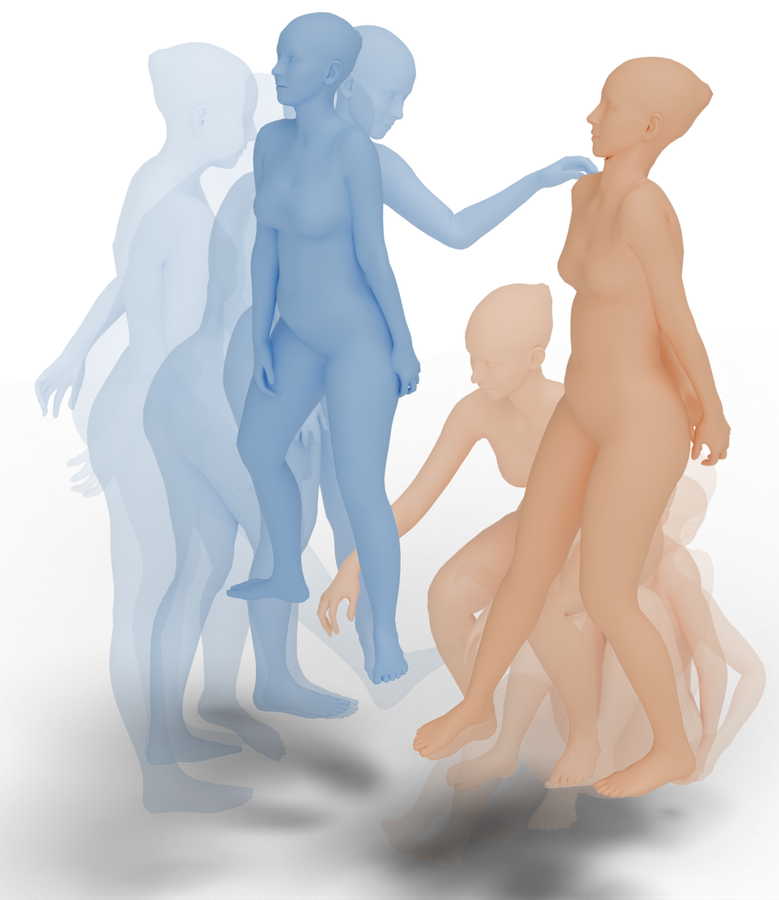}{%
  \qm{Ref.} & \sqv{GT} & \multicolumn{2}{R{3.44in}}{One kicks the other with their right foot, then strikes them again with their left foot. the other person tries to resist but then stands up.} \\[2pt]
  \qm{LLaMMo} & \sqv{Ov.} & One person sits on the ground, and another person walks over to him/her. The first person extends his/her right hand, and the second person grabs it with both hands to assist him/her in standing up. & \sqv{(not generated)} \\
   & \sqv{P1} & \qA{} sits on the ground and \qb{is assisted in standing up by \qB{}}.\qno & \qA{} sits on the ground and \qb{is assisted in standing up by \qB{}}.\qno \\
   & \sqv{P2} & \qB{} sits on the ground and \qb{is helped to stand up by \qA{}}.\qno & \qB{} sits on the ground and \qb{is assisted in standing up by \qA{}}.\qno \\
   & \sqv{Rel} & \qB{} walks towards \qA{}, who is sitting on the ground. \qB{} extends both hands to support \qA{}'s left arm and \qb{helps lift \qA{} up} from the ground.\qno & \qB{} grabs \qA{}'s right arm and \qb{helps \qA{} stand up}.\qno \\[2pt]
  \rowcolor{qours}
  \qm{InterRole} & \sqv{Ov.} & One person walks towards the other, and when he/she is halfway, the second person lifts his/her right foot to step on the first person's left foot. The first person then stands up and raises both hands in front of him/her. & \sqv{(not generated)} \\
  \rowcolor{qours}
   & \sqv{P1} & \qA{} sits on the ground and \qg{is kicked by \qB{}}.\qyes & \qA{} stands to the right of \qB{}, lifts his/her left foot, and \qg{kicks \qB{}'s right leg}.\qyes \\
  \rowcolor{qours}
   & \sqv{P2} & \qB{} \qg{walks towards \qA{}} and steps on \qA{}'s right foot with his/her left foot.\qyes & \qB{} sits on the ground and \qg{is kicked by \qA{}}.\qyes \\
  \rowcolor{qours}
   & \sqv{Rel} & \qg{\qB{} steps on \qA{}'s right foot}, causing \qA{} to stand up and touch the injured foot.\qyes & \qg{\qA{} kicks} \qB{}'s left leg, causing \qB{} to fall down.\qyes \\
}
\caption{%
Zero-shot transfer from Inter-X to InterHuman: both models are trained on Inter-X only (20-joint input) and never see InterHuman data, statistics or captions during training. The renders show the SMPL-X meshes of the motion that the models read as joints: \textcolor{pone}{blue} = input slot~1, \textcolor{ptwo}{orange} = input slot~2; fainter silhouettes are earlier frames; (b) is viewed from the opposite side so that slot~1 stays on the left. Descriptions are shown verbatim for the original input (a) and after exchanging the two motion streams (b); InterHuman provides only an Overall reference (Ref., the first of three); {Ov.} is the Overall description, generated for (a) only and not marked. \textcolor{qokc}{\ding{51}}/\textcolor{qbadc}{\ding{55}} mark whether the described action is attributed to the participant who performs it, judged by us from the renders. Over all 1176 InterHuman test sequences, LLaMMo writes the same description for both participants (after exchanging role names) in 74.6\% of sequences versus 27.9\% for InterRole, and its swap-consistency gain $G_{\mathrm{swap}}$ is 0.034 versus 0.081; the Overall BERTScore is similar (0.291 vs.\ 0.301).
}
\label{fig:crossdataset_supp}
\end{figure*}

%% file: figures/fig_retrieval_top5.tex
\providecolor{pone}{RGB}{42,120,214}
\providecolor{ptwo}{RGB}{214,120,60}
\providecolor{qokc}{RGB}{26,127,75}
\providecolor{qbadc}{RGB}{203,36,49}
\providecolor{qours}{RGB}{232,240,252}
\providecolor{qlgray}{RGB}{110,110,110}
\providecommand{\qA}{\textcolor{pone}{\mbox{Person\,1}}}
\providecommand{\qB}{\textcolor{ptwo}{\mbox{Person\,2}}}
\providecommand{\ryes}[1]{\textcolor{qokc}{\ding{51}~#1}}
\providecommand{\rno}[1]{\textcolor{qbadc}{\ding{55}~#1}}
\providecommand{\rimg}[1]{\includegraphics[width=1.42in,height=1.00in,keepaspectratio]{#1}}
\providecommand{\rcase}[6]{%
  \noindent
  \begin{minipage}[c]{0.16in}\centering\rotatebox[origin=c]{90}{\textbf{\small #1}}\end{minipage}\hfill
  \begin{minipage}[c]{1.22in}\raggedright #2\end{minipage}\hfill
  \colorbox{qours}{\begin{minipage}[c]{2.52in}
    \begin{minipage}[c]{1.44in}\centering\rimg{#3}\end{minipage}\hfill\begin{minipage}[c]{1.04in}\raggedright #4\end{minipage}\end{minipage}}\hfill
  \begin{minipage}[c]{2.52in}
    \begin{minipage}[c]{1.44in}\centering\rimg{#5}\end{minipage}\hfill\begin{minipage}[c]{1.04in}\raggedright #6\end{minipage}\end{minipage}\par
}
\providecommand{\rrule}{\par\vspace{2pt}{\color{black!22}\hrule height 0.4pt}\vspace{3pt}}
\providecommand{\rtile}[3]{%
  \begin{minipage}[t]{1.20in}\centering
    \includegraphics[width=1.18in,height=0.80in,keepaspectratio]{#1}\\[1pt]
    {\color{qlgray}\scriptsize #2}\ #3
  \end{minipage}}
\providecommand{\rrow}[2]{%
  \noindent\begin{minipage}[c]{0.62in}\raggedright\textbf{#1}\end{minipage}\hfill
  \begin{minipage}[c]{6.18in}#2\end{minipage}\par}
\begin{figure*}[p]
\centering
\fontsize{6.8}{8.0}\selectfont
\noindent\textbf{Push.} Query: \qA{} pushes \qB{}'s left arm, causing \qB{} to take a few steps to the right.\par\vspace{3pt}
\rrow{InterRole (ours)}{\colorbox{qours}{\begin{minipage}{6.10in}\rtile{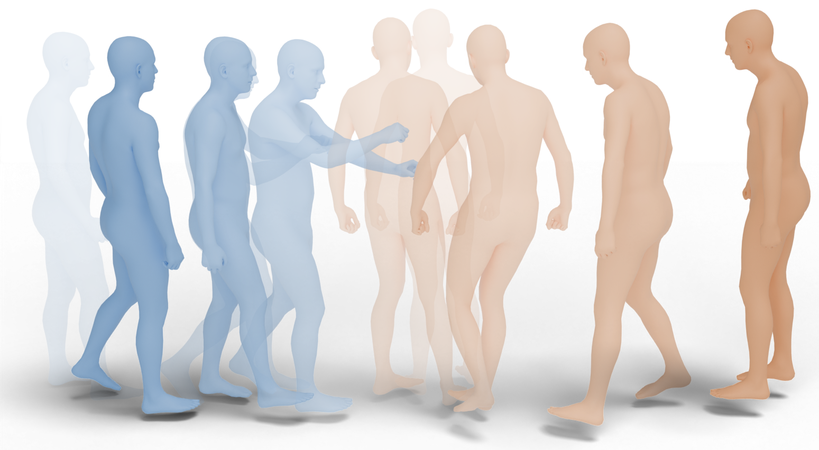}{1}{\textcolor{qokc}{\ding{51}}}\hfill\rtile{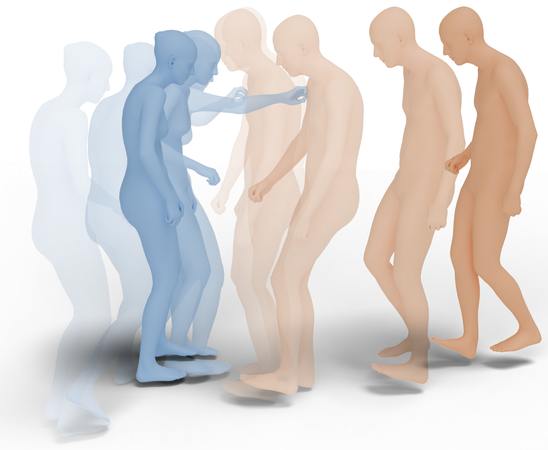}{2}{\textcolor{qokc}{\ding{51}}}\hfill\rtile{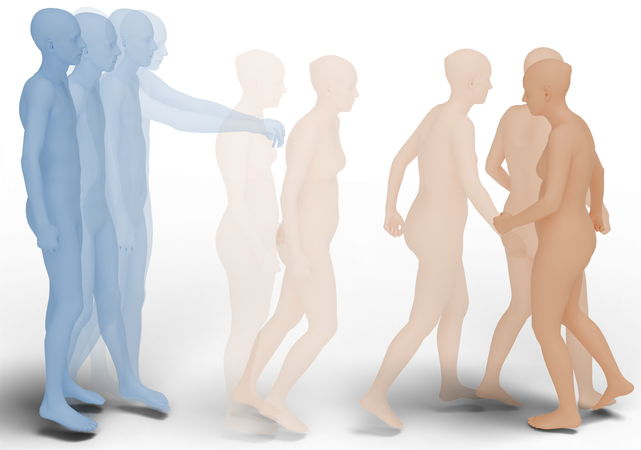}{3}{\textcolor{qokc}{\ding{51}}}\hfill\rtile{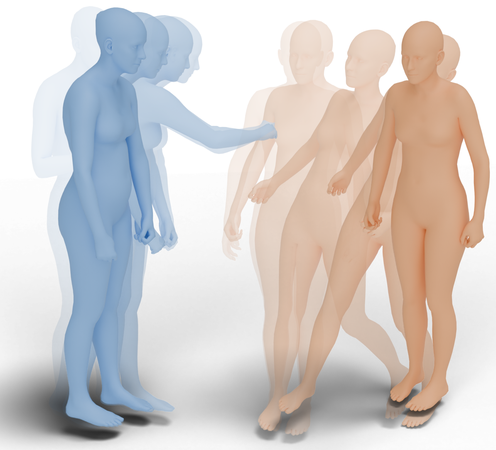}{4}{\textcolor{qbadc}{\ding{55}\,Hit}}\hfill\rtile{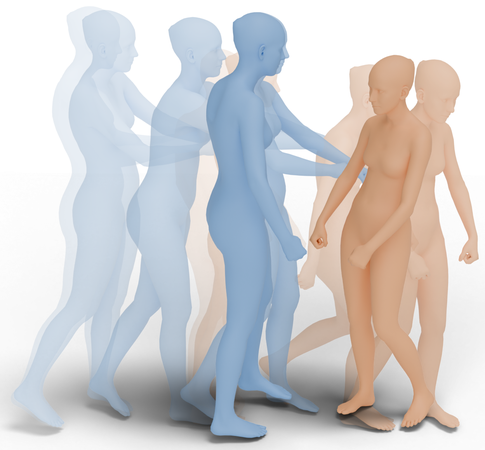}{5}{\textcolor{qokc}{\ding{51}}}\end{minipage}}}
\vspace{2pt}
\rrow{LLaMMo}{\rtile{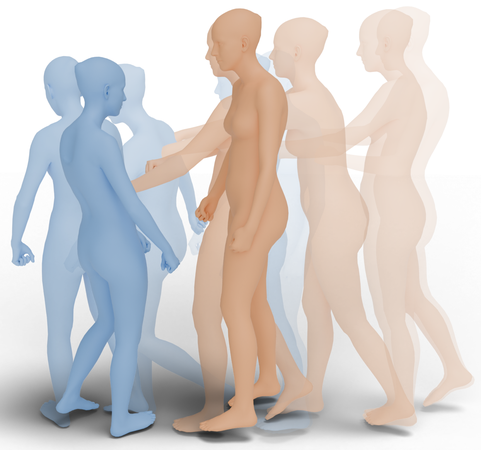}{1}{\textcolor{qbadc}{\ding{55}\,roles reversed}}\hfill\rtile{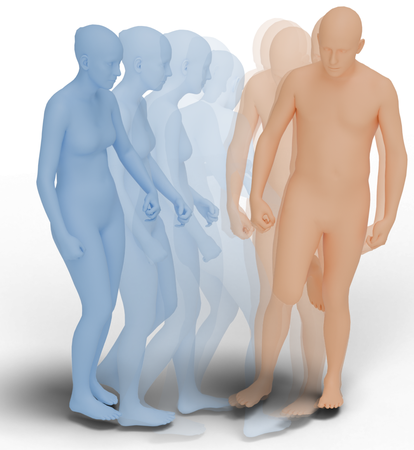}{2}{\textcolor{qbadc}{\ding{55}\,Step on foot}}\hfill\rtile{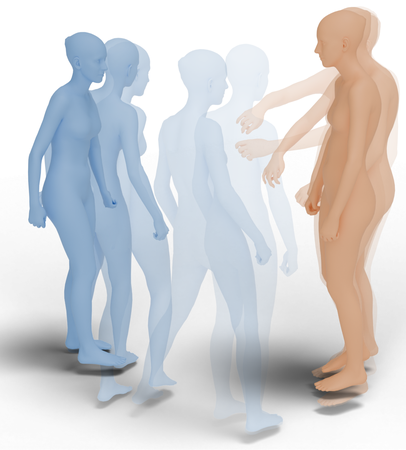}{3}{\textcolor{qbadc}{\ding{55}\,Hit}}\hfill\rtile{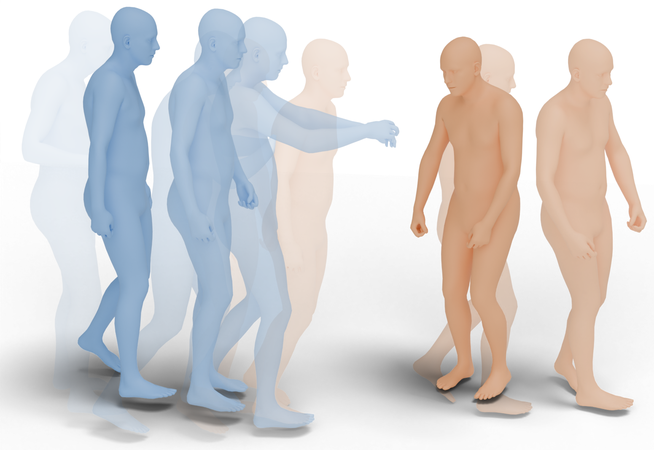}{4}{\textcolor{qokc}{\ding{51}}}\hfill\rtile{img_figures_retrieval_top5_G046T001A007R014__swapped.png}{5}{\textcolor{qokc}{\ding{51}}}}
\rrule\vspace{5pt}
\noindent\textbf{Hug.} Query: \qB{} and \qA{} stand side by side, sway, turn to face each other, and briefly hug.\par\vspace{3pt}
\rrow{InterRole (ours)}{\colorbox{qours}{\begin{minipage}{6.10in}\rtile{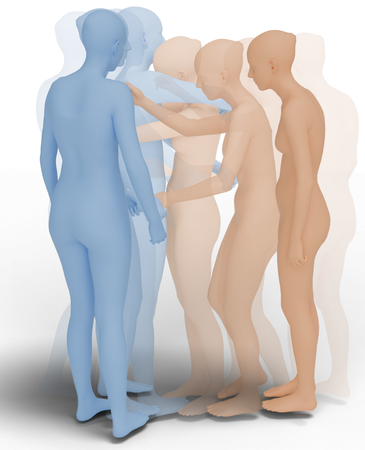}{1}{\textcolor{qokc}{\ding{51}}}\hfill\rtile{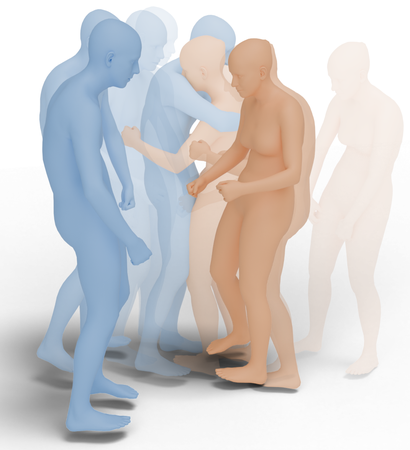}{2}{\textcolor{qokc}{\ding{51}}}\hfill\rtile{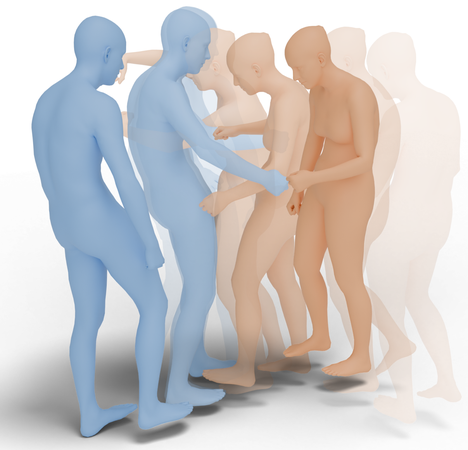}{3}{\textcolor{qokc}{\ding{51}}}\hfill\rtile{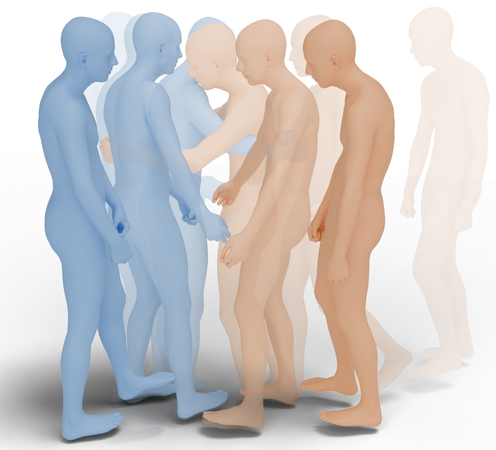}{4}{\textcolor{qokc}{\ding{51}}}\hfill\rtile{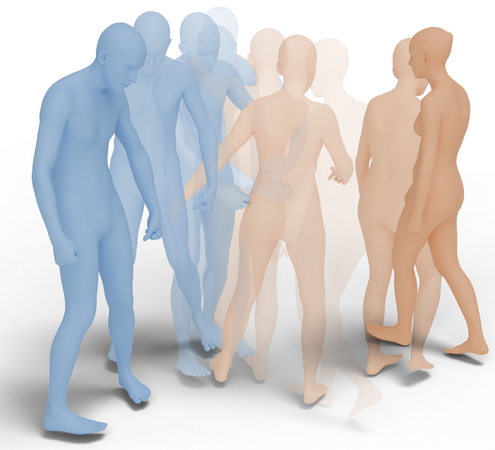}{5}{\textcolor{qokc}{\ding{51}}}\end{minipage}}}
\vspace{2pt}
\rrow{LLaMMo}{\rtile{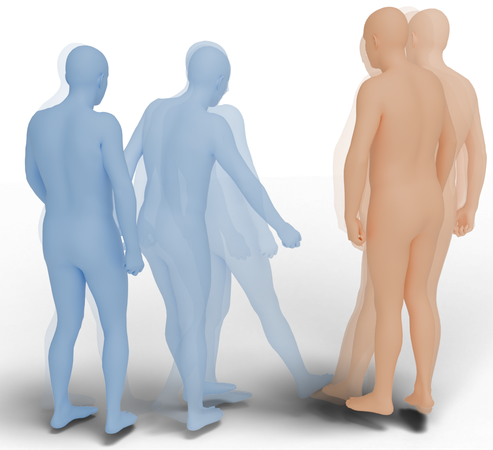}{1}{\textcolor{qbadc}{\ding{55}\,Step on foot}}\hfill\rtile{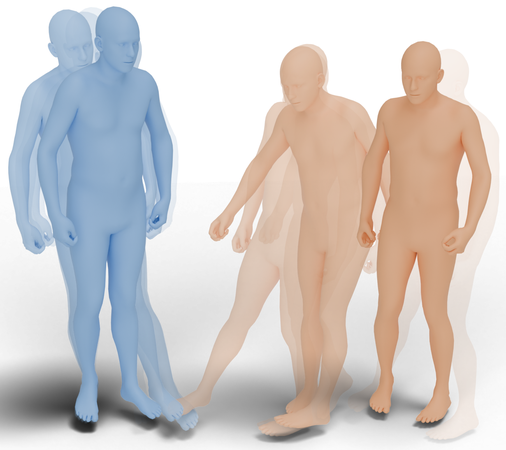}{2}{\textcolor{qbadc}{\ding{55}\,Step on foot}}\hfill\rtile{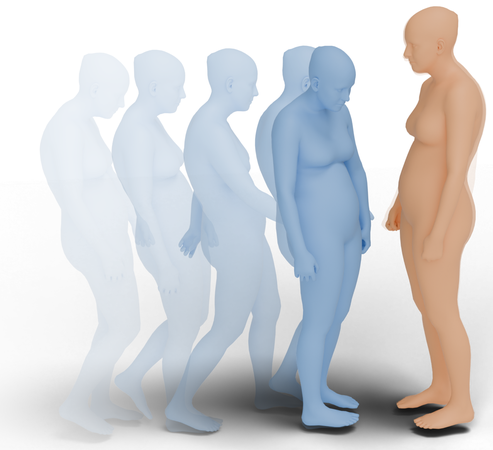}{3}{\textcolor{qbadc}{\ding{55}\,Walk towards}}\hfill\rtile{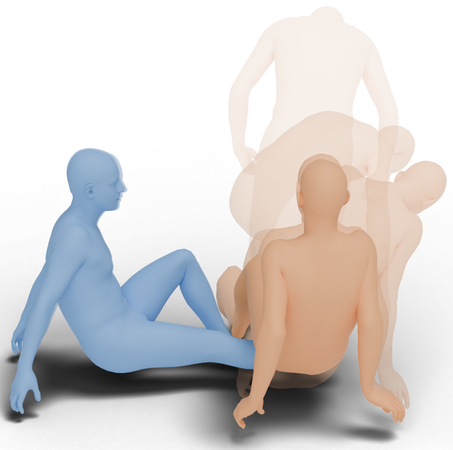}{4}{\textcolor{qbadc}{\ding{55}\,Sit on leg}}\hfill\rtile{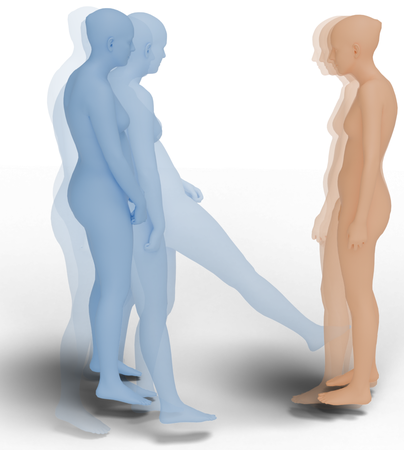}{5}{\textcolor{qbadc}{\ding{55}\,Kick}}}
\rrule\vspace{5pt}
\noindent\textbf{Pull.} Query: \qB{} steps forward and holds \qA{}'s right elbow joint with both hands, pulling \qA{} towards themselves.\par\vspace{3pt}
\rrow{InterRole (ours)}{\colorbox{qours}{\begin{minipage}{6.10in}\rtile{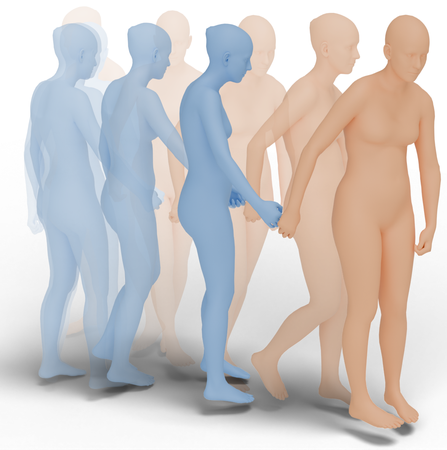}{1}{\textcolor{qokc}{\ding{51}}}\hfill\rtile{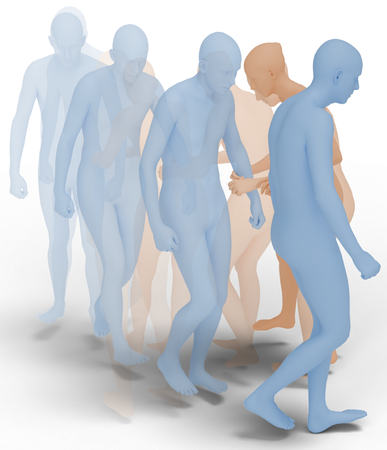}{2}{\textcolor{qokc}{\ding{51}}}\hfill\rtile{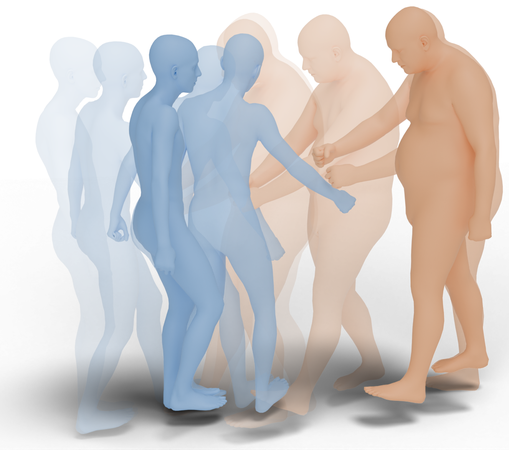}{3}{\textcolor{qokc}{\ding{51}}}\hfill\rtile{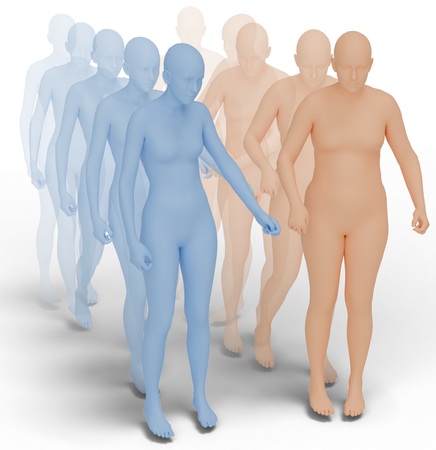}{4}{\textcolor{qokc}{\ding{51}}}\hfill\rtile{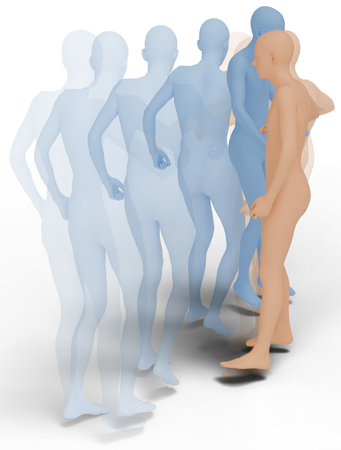}{5}{\textcolor{qokc}{\ding{51}}}\end{minipage}}}
\vspace{2pt}
\rrow{LLaMMo}{\rtile{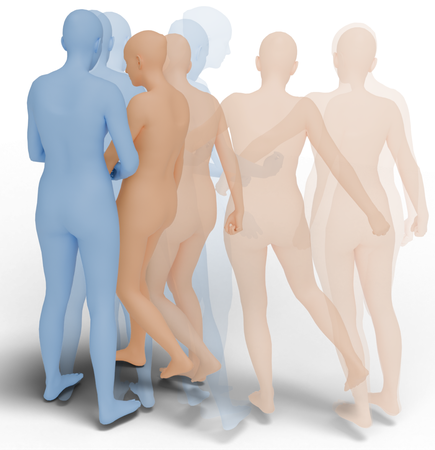}{1}{\textcolor{qbadc}{\ding{55}\,roles reversed}}\hfill\rtile{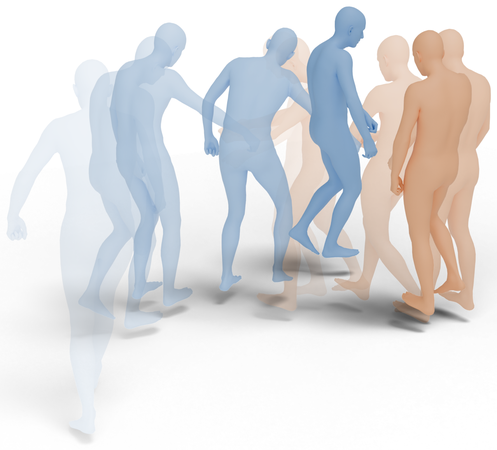}{2}{\textcolor{qokc}{\ding{51}}}\hfill\rtile{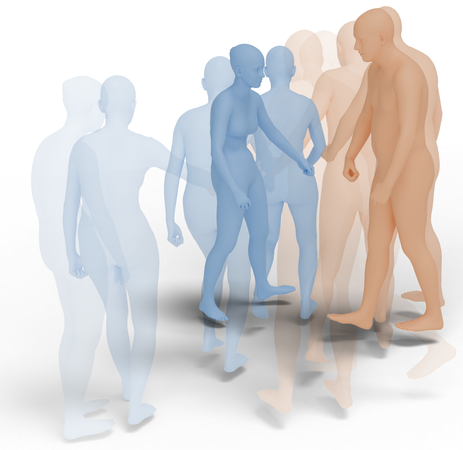}{3}{\textcolor{qbadc}{\ding{55}\,Grab}}\hfill\rtile{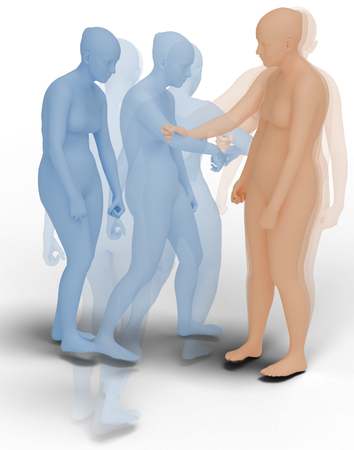}{4}{\textcolor{qbadc}{\ding{55}\,Grab}}\hfill\rtile{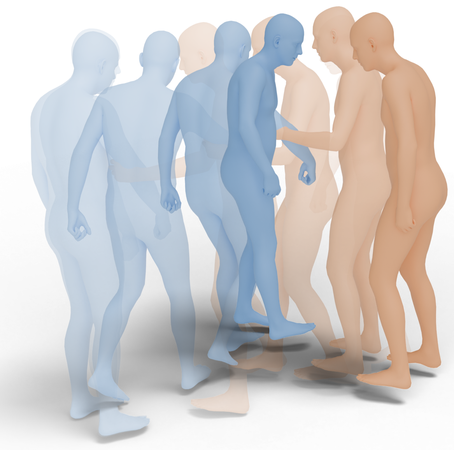}{5}{\textcolor{qbadc}{\ding{55}\,Grab}}}
\caption{Top-5 text-to-motion retrieval for three queries (full test gallery of originals and participant-swapped copies). \textcolor{pone}{Blue} = \qA{}, \textcolor{ptwo}{orange} = \qB{}; fainter silhouettes are earlier frames. Under each motion: rank and verdict; \textcolor{qokc}{\ding{51}} = queried action with the queried participant as the actor (judged from the retrieved motion's first stored annotation that names the participants), whichever sequence it comes from; for symmetric interactions (handshake, hug) only the action is checked; \textcolor{qbadc}{\ding{55}} = roles reversed, or a different action (named).}
\label{fig:retrieval_top5}
\end{figure*}

%% file: figures/fig_robot_supp.tex
\input{figures/retrieval_robot_macros}
{\renewcommand{\rrobotdir}{retrieval/robot_add}%
\begin{figure*}[t]
\centering
\fontsize{6.8}{8.0}\selectfont
\noindent\begin{minipage}[c]{0.16in}\ \end{minipage}\hfill\begin{minipage}[c]{1.22in}\textbf{Text query}\end{minipage}\hfill\makebox[2.64in][l]{\textbf{InterRole (ours): Top-1 motion}}\hfill\makebox[2.52in][l]{\textbf{LLaMMo: Top-1 motion}}\par\rrule
\rcaseR{Hug}{\qB{} and \qA{} stand side by side, sway, turn to face each other, and briefly hug.}{{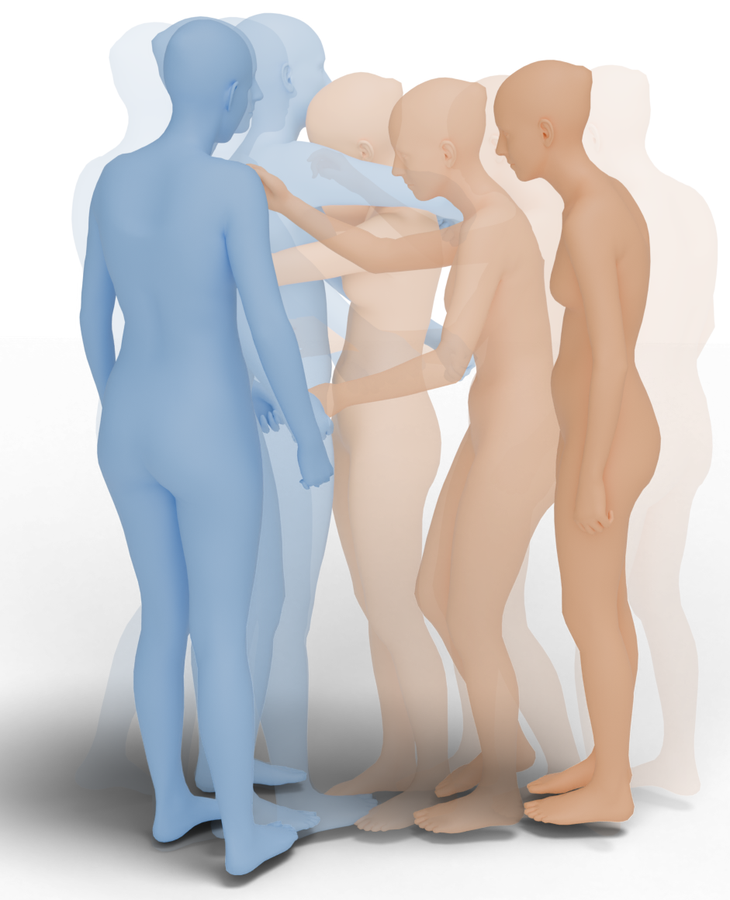}{0.279}{0.062}{0.755}{0.107}}{%
{\color{qlgray}\scriptsize Hug}\\[1pt]``\qB{} approaches and hugs \qA{}'s back, while \qA{} reciprocates the hug.''\\[2pt]\ryes{same action}}{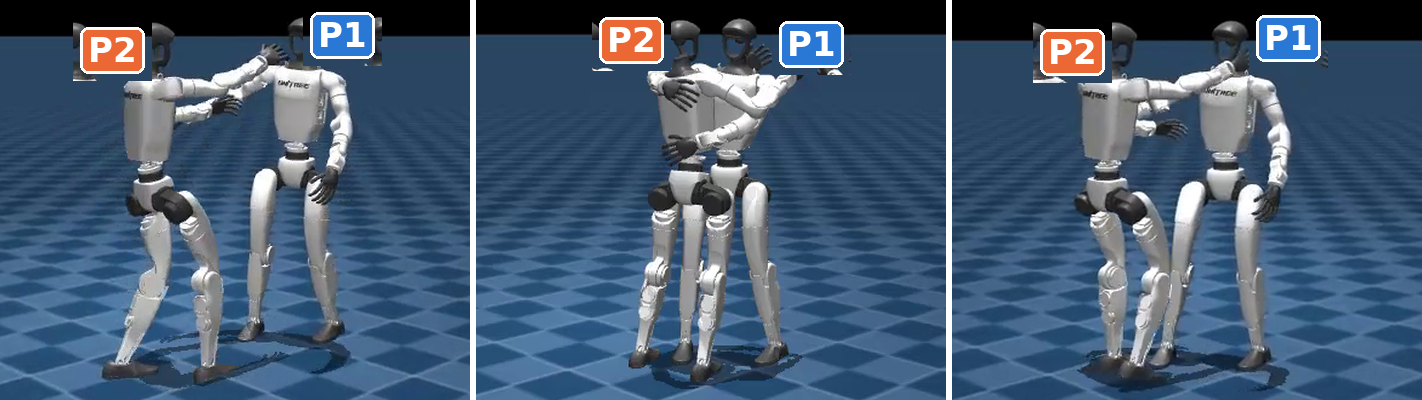}{{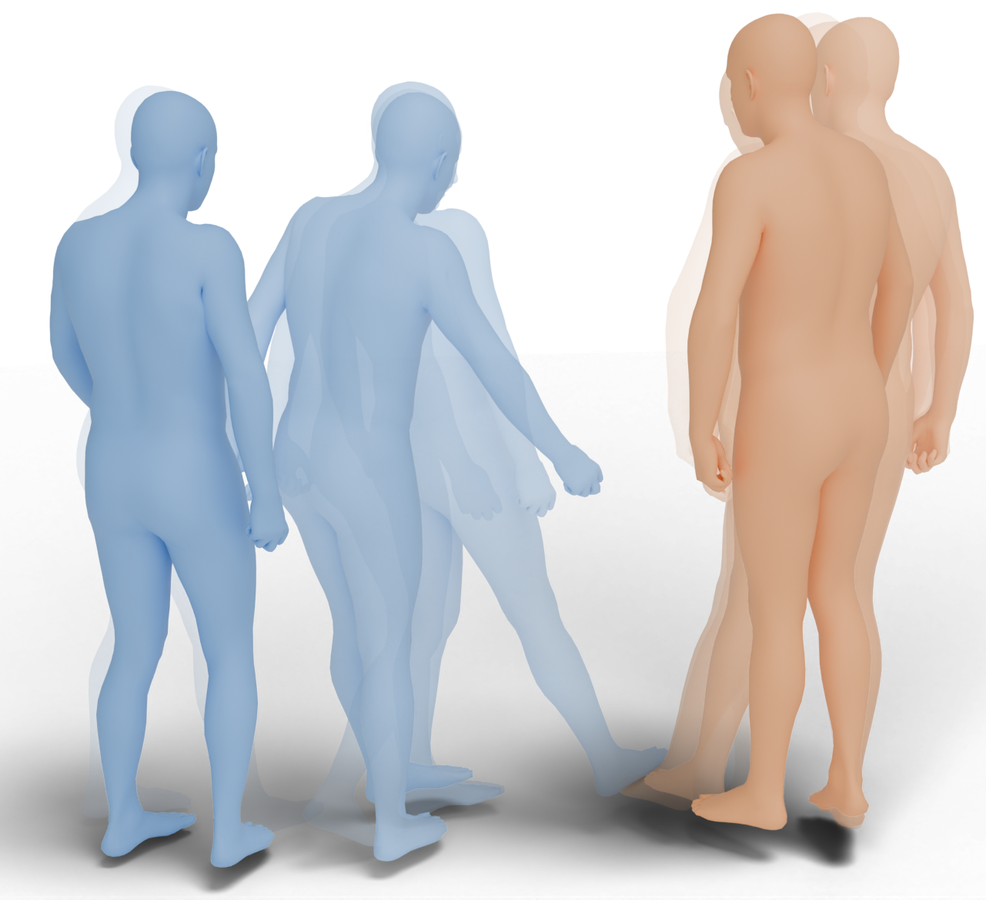}{0.172}{0.141}{0.789}{0.055}}{%
{\color{qlgray}\scriptsize Step on foot}\\[1pt]``\qA{} walks towards \qB{}, stretches out the right foot, and steps on \qB{}'s left foot.''\\[2pt]\rno{other action (step on foot)}}{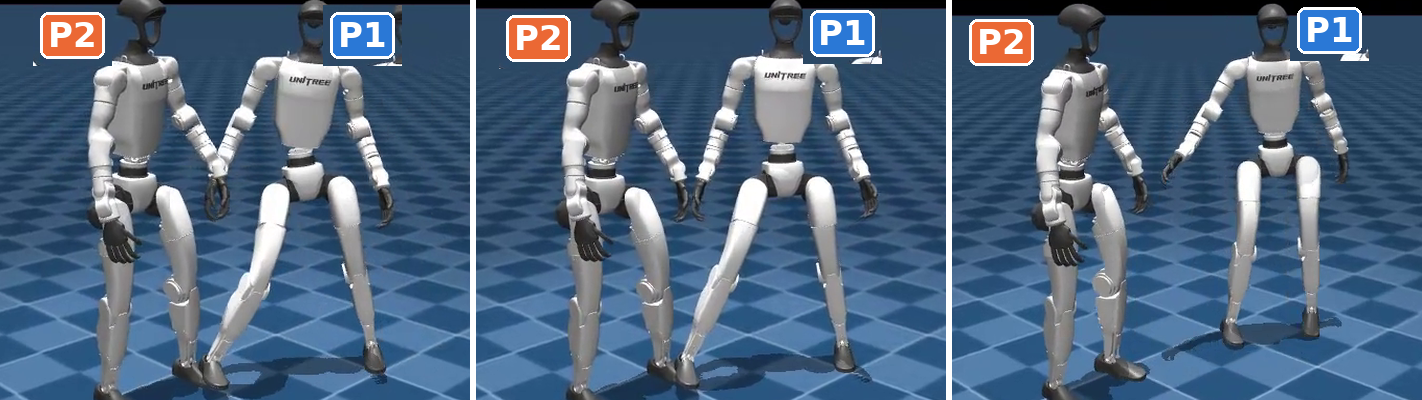}
\rrule
\rcaseR{Pull}{\qB{} steps forward and holds \qA{}'s right elbow joint with both hands, pulling \qA{} towards themselves.}{{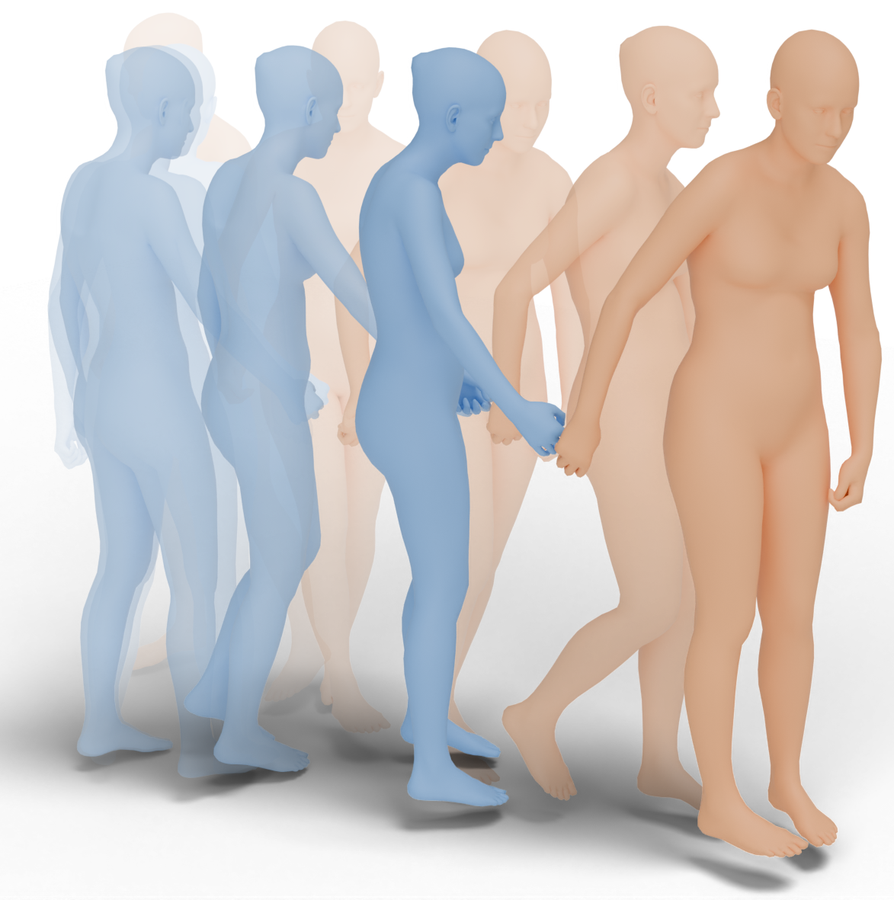}{0.503}{0.092}{0.908}{0.073}}{%
{\color{qlgray}\scriptsize Pull}\\[1pt]``\qB{} walks up to \qA{}, holds their hand, and pulls them forward.''\\[2pt]\ryes{correct roles}}{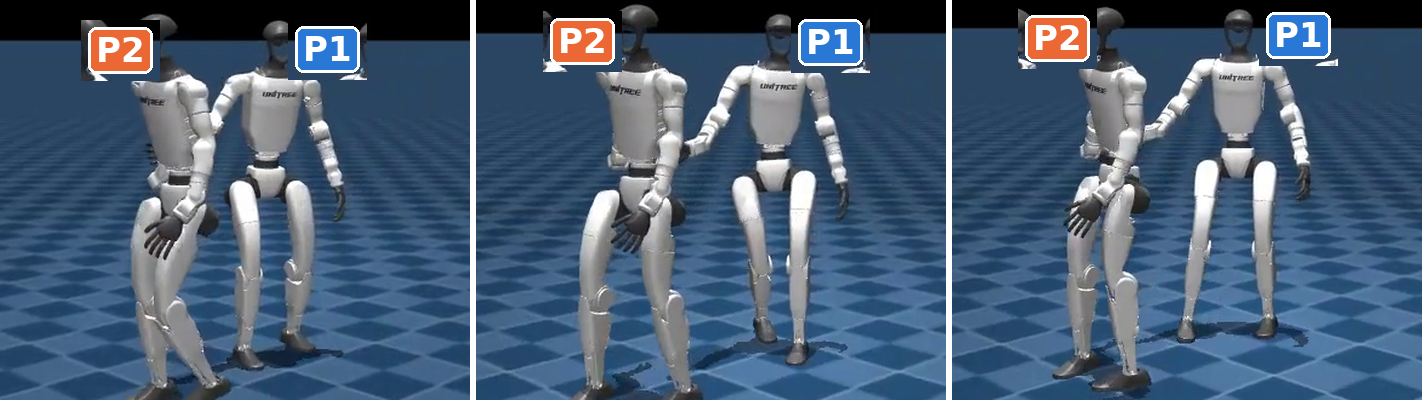}{{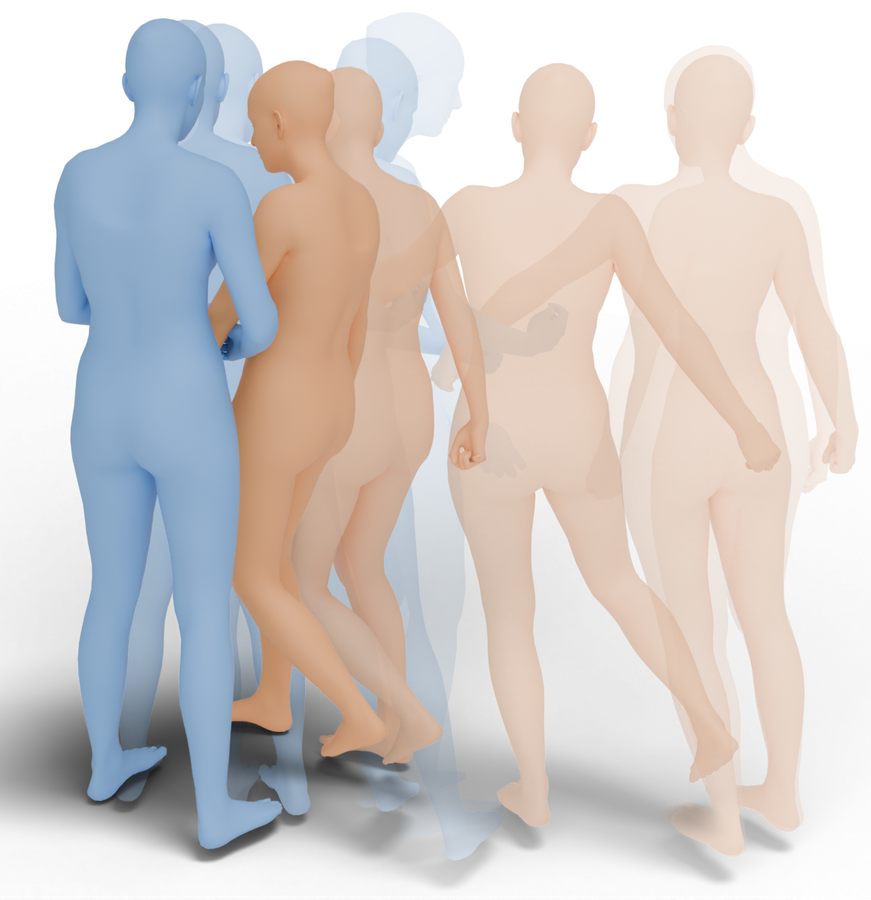}{0.187}{0.060}{0.340}{0.107}}{%
{\color{qlgray}\scriptsize Pull}\\[1pt]``\qA{} pulls \qB{}, and \qB{} raises his hands and moves forward.''\\[2pt]\rno{roles reversed}}{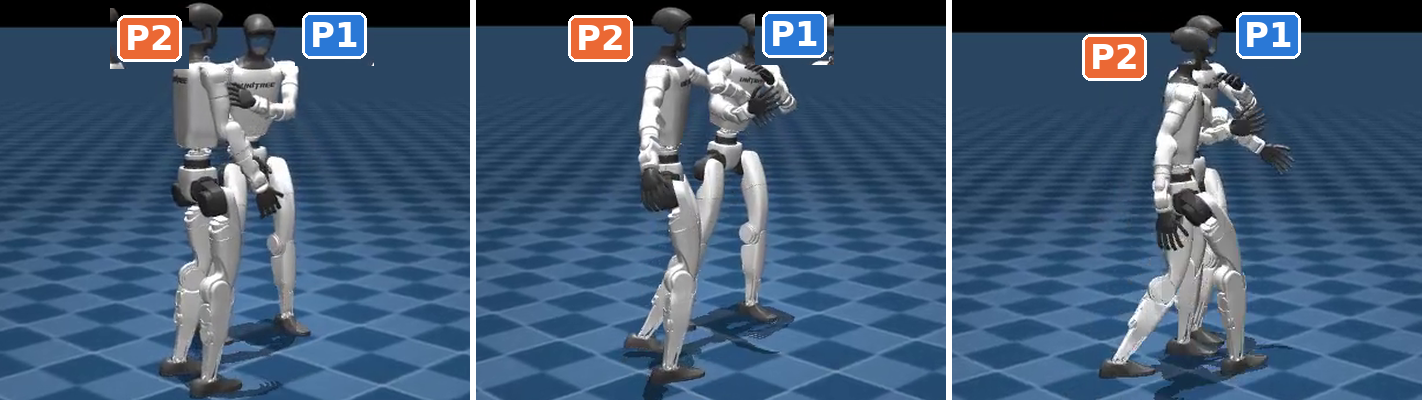}
\caption{Top-1 retrieval and its execution by two simulated humanoid robots for a Hug and a Pull query (robots labelled in the frames; the simulator camera differs from the render camera). For Hug, InterRole retrieves a hug and the robots hug, whereas LLaMMo retrieves a different action (stepping on a foot) and the robots do not hug. For Pull, InterRole's retrieved motion has Person~2 pull Person~1 as the query asks, whereas LLaMMo's has the roles reversed. The Push example is in Fig.~6 of the main paper.}
\label{fig:robot_supp}
\end{figure*}}

%% file: figures/fig_retrieval_binary.tex
\providecolor{pone}{RGB}{42,120,214}
\providecolor{ptwo}{RGB}{214,120,60}
\providecolor{qokc}{RGB}{26,127,75}
\providecolor{qbadc}{RGB}{203,36,49}
\providecolor{qours}{RGB}{232,240,252}
\providecolor{qlgray}{RGB}{110,110,110}
\providecommand{\qA}{\textcolor{pone}{\mbox{Person\,1}}}
\providecommand{\qB}{\textcolor{ptwo}{\mbox{Person\,2}}}
\providecommand{\ryes}[1]{\textcolor{qokc}{\ding{51}~#1}}
\providecommand{\rno}[1]{\textcolor{qbadc}{\ding{55}~#1}}
\providecommand{\rimg}[1]{\includegraphics[width=1.42in,height=1.00in,keepaspectratio]{#1}}
\providecommand{\rcase}[6]{%
  \noindent
  \begin{minipage}[c]{0.16in}\centering\rotatebox[origin=c]{90}{\textbf{\small #1}}\end{minipage}\hfill
  \begin{minipage}[c]{1.22in}\raggedright #2\end{minipage}\hfill
  \colorbox{qours}{\begin{minipage}[c]{2.52in}
    \begin{minipage}[c]{1.44in}\centering\rimg{#3}\end{minipage}\hfill\begin{minipage}[c]{1.04in}\raggedright #4\end{minipage}\end{minipage}}\hfill
  \begin{minipage}[c]{2.52in}
    \begin{minipage}[c]{1.44in}\centering\rimg{#5}\end{minipage}\hfill\begin{minipage}[c]{1.04in}\raggedright #6\end{minipage}\end{minipage}\par
}
\providecommand{\rrule}{\par\vspace{2pt}{\color{black!22}\hrule height 0.4pt}\vspace{3pt}}
\providecommand{\bimg}[1]{\includegraphics[width=1.46in,height=1.04in,keepaspectratio]{#1}}
\providecommand{\btag}[2]{\par\vspace{1pt}{\scriptsize\textcolor{#1}{#2}}}
\providecommand{\bbar}[3]{\makebox[1.20in][l]{\rlap{\hspace*{0.60in}\textcolor{black!70}{\rule[-3pt]{0.6pt}{11pt}}}%
  \hspace*{#1in}\textcolor{#3}{\rule{#2in}{5pt}}}}
\providecommand{\bscore}[6]{
  \begin{minipage}[c]{0.60in}\raggedright\textbf{#1}\end{minipage}%
  \begin{minipage}[c]{1.22in}#2\\[-1pt]{\color{qlgray}\tiny #5\,/\,#6}\end{minipage}%
  \begin{minipage}[c]{0.62in}\raggedleft #3~#4\end{minipage}}
\providecommand{\bcase}[6]{%
  \noindent
  \begin{minipage}[c]{0.16in}\centering\rotatebox[origin=c]{90}{\textbf{\small #1}}\end{minipage}\hfill
  \begin{minipage}[c]{1.12in}\raggedright #2\end{minipage}\hfill
  \begin{minipage}[c]{1.50in}\centering #3\btag{qokc}{\ding{51}\ matches query}\end{minipage}\hfill
  \begin{minipage}[c]{1.50in}\centering #4\btag{qbadc}{\ding{55}\ roles reversed}\end{minipage}\hfill
  \begin{minipage}[c]{2.44in}
    {\setlength{\fboxsep}{0pt}\colorbox{qours}{\begin{minipage}{2.44in}\vspace{2pt}#5\vspace{2pt}\end{minipage}}}\\[4pt]
    #6
  \end{minipage}\par}
\providecommand{\brule}{\par\vspace{2pt}{\color{black!22}\hrule height 0.4pt}\vspace{4pt}}
\begin{figure*}[!t]
\centering
\fontsize{6.8}{8.0}\selectfont
\noindent\begin{minipage}[c]{0.16in}\ \end{minipage}\hfill\begin{minipage}[c]{1.12in}\textbf{Text query}\end{minipage}\hfill\makebox[1.50in]{\textbf{(a) Original}}\hfill\makebox[1.50in]{\textbf{(b) Participants swapped}}\hfill\begin{minipage}[c]{2.44in}\textbf{Which version is preferred?}\\[1pt]\hspace*{0.60in}\makebox[1.20in]{\scriptsize\textcolor{qbadc}{$\leftarrow$ (b)}\hfill\textcolor{qokc}{(a) $\rightarrow$}}\end{minipage}\par\brule
\bcase{Massaging leg}{\qB{} rubs \qA{}'s legs while \qA{} bends down and looks at \qB{}.}{\bimg{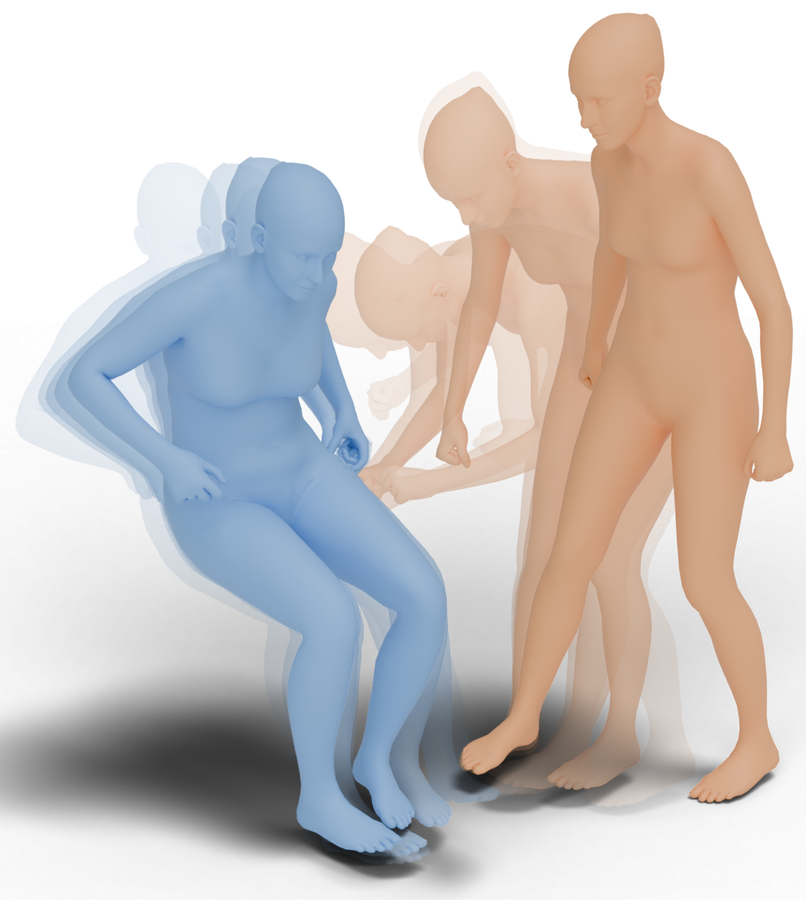}}{\bimg{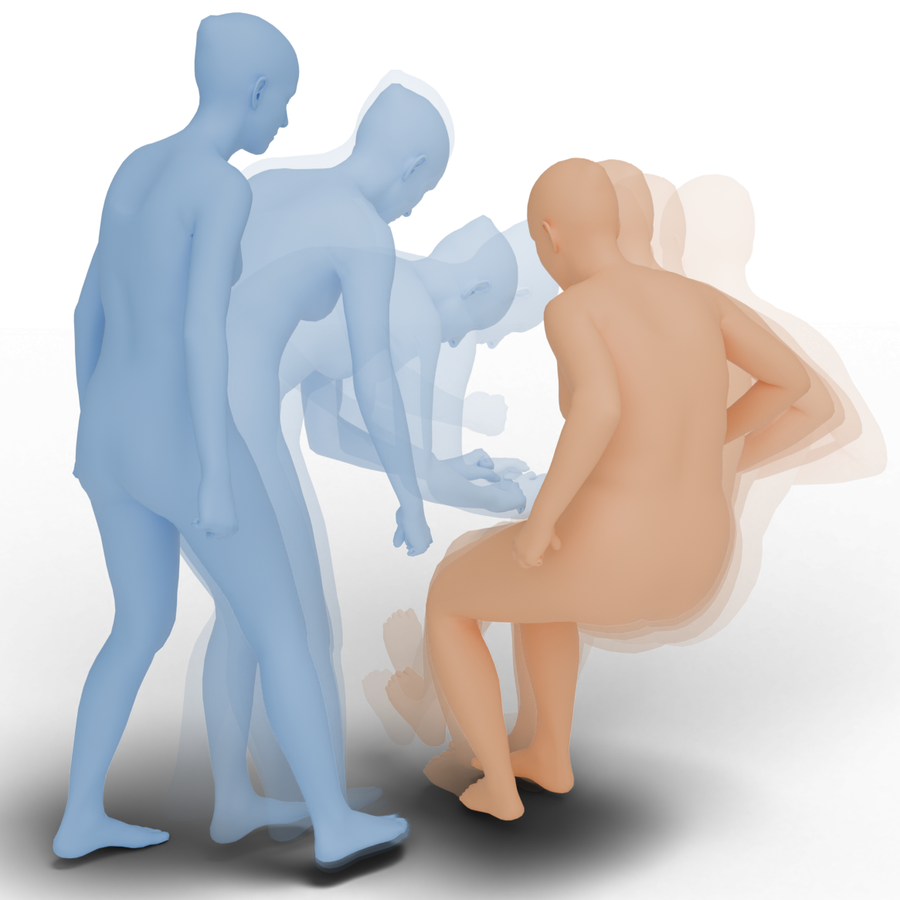}}{%
\bscore{InterRole (ours)}{\bbar{0.600}{0.446}{qokc}}{+0.149}{\textcolor{qokc}{\ding{51}}}{0.619}{0.471}}{%
\bscore{LLaMMo}{\bbar{0.575}{0.025}{qbadc}}{-0.002}{\textcolor{qbadc}{\ding{55}}}{0.404}{0.406}}
\brule
\bcase{Shoulder to shoulder}{\qA{} places the right arm on \qB{}'s left shoulder, and they walk forward together.}{\bimg{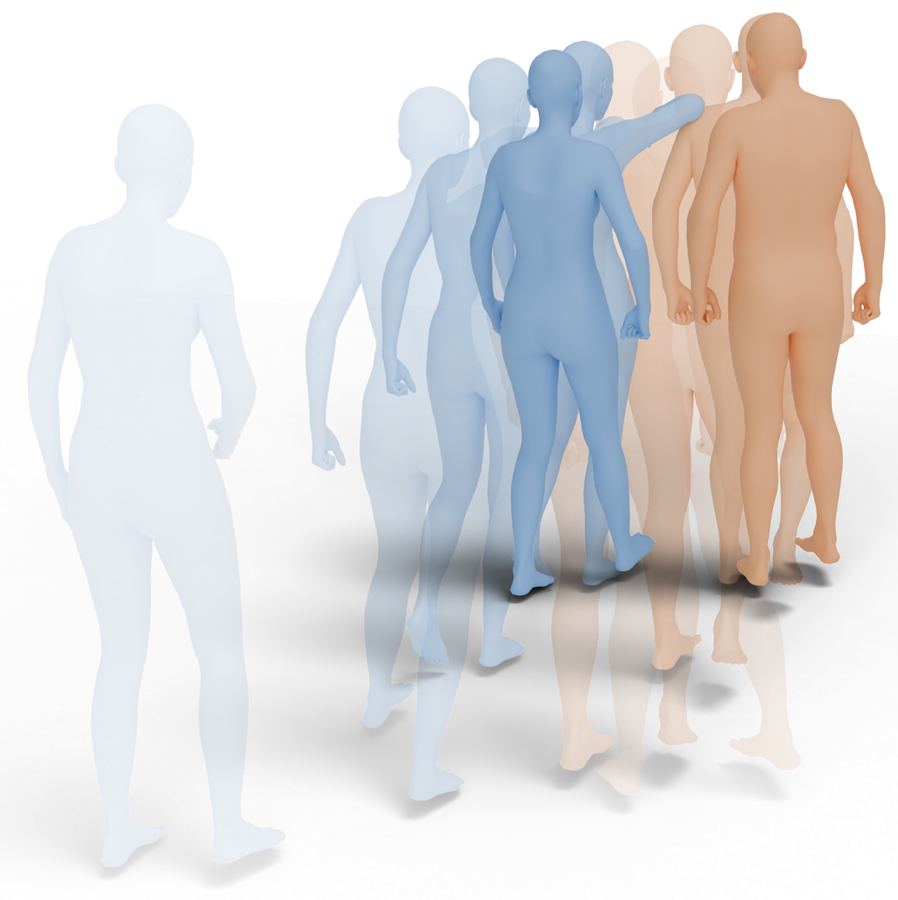}}{\bimg{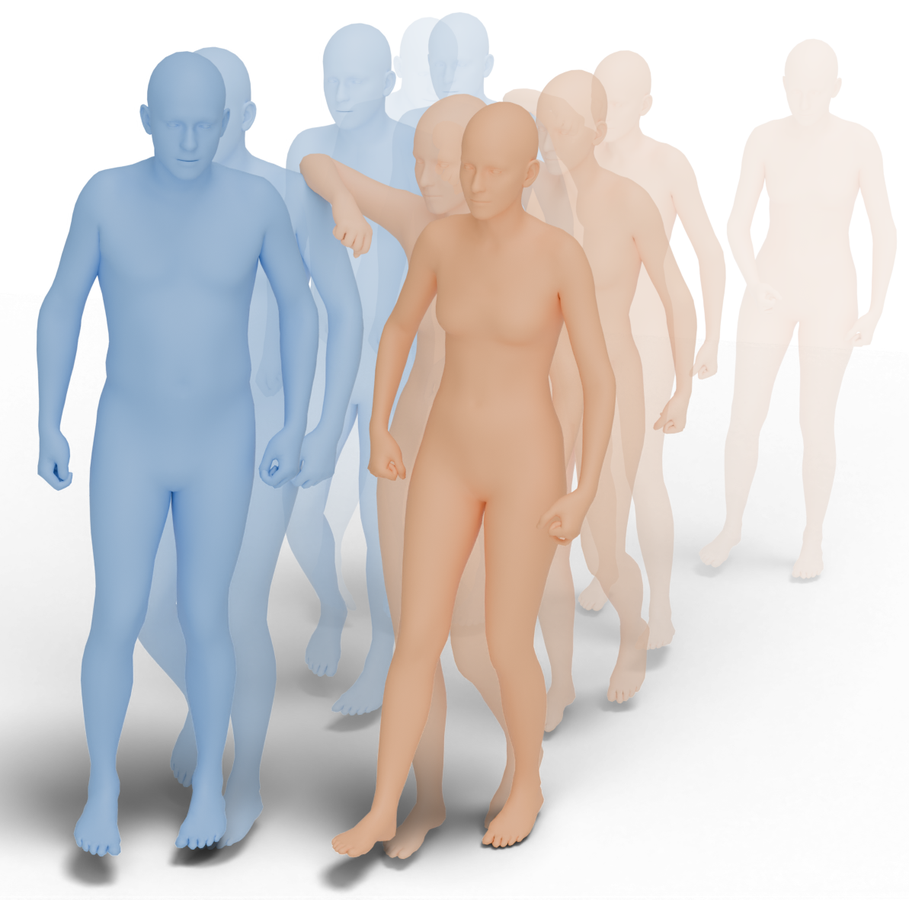}}{%
\bscore{InterRole (ours)}{\bbar{0.600}{0.152}{qokc}}{+0.051}{\textcolor{qokc}{\ding{51}}}{0.726}{0.675}}{%
\bscore{LLaMMo}{\bbar{0.575}{0.025}{qbadc}}{-0.004}{\textcolor{qbadc}{\ding{55}}}{0.498}{0.502}}
\caption{Role-sensitivity test on Inter-X. For a Relation query we compare the motion it describes (a) with the same motion after exchanging the two participants' streams (b); only (a) matches the query. Each method scores both versions and the bar shows $s(a)-s(b)$ (axis clipped at $\pm0.2$; small numbers give $s(a)\,/\,s(b)$): \textcolor{qokc}{right} = the version that matches the query is preferred (\ding{51}), \textcolor{qbadc}{left} = the role-reversed one (\ding{55}). Scores are compared within a method only. Renders: \textcolor{pone}{blue} = \qA{}, \textcolor{ptwo}{orange} = \qB{}; (b) is viewed from the opposite side so that \qA{} stays on the left; fainter silhouettes are earlier frames. Over the asymmetric-interaction test subset (1108 sequences, 3289 Relation queries; symmetric actions such as handshakes are excluded because exchanging roles changes nothing) the matching version wins for 72.2\% of queries with InterRole vs.\ 59.5\% with LLaMMo (chance $50\%$).}
\label{fig:retrieval_binary}
\end{figure*}